\documentclass{article}

\usepackage{amsfonts,amsmath,amssymb,amsthm,mathrsfs}
\usepackage{chngcntr}
\usepackage{fancyvrb,fancyhdr}
\usepackage{etoolbox}
\usepackage{enumerate,enumitem}
\usepackage{tabularx,longtable,makecell}
\usepackage{multirow}
\usepackage{booktabs}
\usepackage{dcolumn}
\usepackage[font=small]{caption}
\usepackage{subcaption}
\usepackage{graphicx,float}
\usepackage[version=4]{mhchem}
\usepackage{tikz}
\usetikzlibrary{matrix,arrows}
\usepackage{indentfirst}
\usepackage[noend]{algpseudocode}
\usepackage{algorithm}
\usepackage[unicode,pdfencoding=auto,psdextra]{hyperref}
\hypersetup{
    pdfborder={0 0 0},
    colorlinks=true,
    linkcolor=blue,     
    urlcolor=blue,
}

\usepackage[sort&compress,numbers]{natbib}

\usepackage{mymath,mycommands}

\setlist[enumerate]{label=(\arabic*)}

\newlength{\midwidth}
\newlength{\smallwidth}
\newlength{\tinywidth}
\newtheorem{definition}{Definition}[section]
\newtheorem{theorem}{Theorem}[section]
\newtheorem{lemma}{Lemma}[section]
\newtheorem{fact}{Proposition}[section]
\newtheorem{remark}{Remark}[section]
\newtheorem{history}[theorem]{Remark}
\newtheorem{example}{Example}[section]
\newtheorem{experiment}{Experiment}[section]

\counterwithin{algorithm}{section}

\newenvironment{keywords}{
  \kwbf{Keywords:}
  \begin{list}{}{
    \setlength{\leftmargin}{0em}
    \setlength{\labelwidth}{0em}
    \setlength{\labelsep}{0em}
  }
  \item
}{
  \end{list}
}

\newcommand{\fref}[1]{Eq.\eqref{#1}}
\newcommand{\asref}[1]{as shown in \autoref{#1}}


\floatname{algorithm}{Algo.}

\makeatletter
\newcommand\figcaption{\def\@captype{figure}\caption}
\newcommand\tabcaption{\def\@captype{table}\caption}
\makeatother

\title{The General Theory of Localization Methods}

\author{Congwei Song \\
  \small{Beijing Institute of Mathematical Sciences and Applications} \\
  \small{\texttt{\href{mailto:william@bimsa.cn}{william@bimsa.cn}}}
  }
\date{\today}

\begin{document}

\maketitle

\begin{abstract}
This paper proposes a general machine learning framework called the localization method, which is fundamentally built on two core concepts: localization kernels and local means --- key components that underpin the self-attention mechanism. To establish a rigorous theoretical foundation, the framework is formally defined through two essential pillars: the formulation of the local(-ized) model and the localization trick.

We systematically investigate the connections between the localization method and a wide range of existing machine learning models/methods, including (but not limited to) kernel methods, lazy learning, the MeanShift algorithm, relaxation labeling, Hopfield networks, local linear embedding (LLE), fuzzy inference, and denoising autoencoders (DAEs). By dissecting these relationships, we clarify the broader theoretical significance of the localization method and demonstrate its practical applicability across diverse machine learning tasks. Furthermore, we explore advanced extensions of the framework, such as adaptive kernels, hierarchical local models, and non-local models. Notably, we show that the Transformer --- a cornerstone of modern sequence modeling --- can be constructed using hierarchical local models, revealing the ability of the localization method to unify and generalize state-of-the-art architectures.

This work not only provides a unified theoretical lens to reinterpret existing models but also offers new methodological tools for designing flexible, data-adaptive learning systems.
\end{abstract}


\section{Introduction}

This paper explores a general nonlinear machine learning approach --- the \term{localization method}. Similar to \term{kernel methods} \cite{micchelli2006,hofmann2008,nusken2023}, localization methods (used in the plural form to highlight their diverse applications) also incorporate the concept of kernel functions, though with a distinct interpretation. While the core concepts and historical development of these two methods differ significantly, they nonetheless share many commonalities.

As a nonlinear approach, the idea of localization methods is more intuitive than that of kernel methods. Consider a machine learning model $y= f(\vect{x})$ with input $\vect{x}\in\mathcal{X}\subset \R^p$ and output $y\in \mathcal{Y}$, where $f:\mathcal{X}\to\mathcal{Y}$ is nonlinear. In this case, the methods of the linear model seem to be ineffective. However, we reasonably assume that $f$ is linear on a small local region $U\subset \mathcal{X}$, such as $f(\vect{x})\approx \vect{x}\cdot \vect{\beta},\vect{x}\in U$. Therefore, the subsample $S(\vect{x}_i\in U)$ in this local region can be used for linear regression to obtain the parameter estimate $\hat{\vect{\beta}}(U)$ depending on $U$. When predicting the value of $f(\vect{x}_*)$ at the target data point $\vect{x}_*$, we also make the prediction according to the neighborhood where $\vect{x}_*$ is located.

Like what is done in the decision tree, the input space $\mathcal{X}$ can be partitioned in advance to obtain a family of local regions $\{U_i\}$, and then perform machine learning independently on each region. Another more flexible method is to construct a neighborhood for the data point $\vect{x}_*$ to be predicted and execute the learning algorithm on this neighborhood. This means that the learning and prediction of the model are carried out simultaneously, or in other words, the learning is not completely independent of the prediction, so it is \term{lazy}. Further, the input space $\mathcal{X}$ can be a very arbitrary set, as long as the concept of \term{neighborhood} (or distance/metric/similarity and generally referred to as topology) can be reasonably defined on it.

The contributions of this paper are mainly divided into four parts:
\begin{enumerate}
\item Introduce the basic concepts of the localization method (localized kernels, smoothed spaces), attempt to establish a theoretical system, and strictly define the local model.
\item Explain how to apply the localization method to the field of machine learning, especially discuss important concepts such as (self-) local mean (also known as Nadaraya-Watson weighted average, as the mathematical expression of the self-attention mechanism).
\item Construct complex models by applying localization techniques to simple models.
\item Point out the connections between the local model and existing models (denoising autoencoders, relaxation labeling, Hopfield networks, fuzzy inference, etc.).
\end{enumerate}

This article employs the following notations.
\begin{enumerate}
  \item $\sum_i,\prod_i$: It represents summation and product. (When the scope of the summation/product operation is clear or does not need to be specifically indicated, the starting and ending numbers of the index will not be marked.)
  \item $\{x_i\}$: A family/set of variables.
\item $X\propto Y$: The two quantities $X, Y$ are in a proportional relationship, that is, $X = CY, C>0$.
\item $X\approx Y$: $X$ is approximately $Y$.
\item $X\sim p$: The \rv $X$ obeys distribution $p$ (may not normalized).
\item $p\sim f$: Equivalent to $p\propto f$, where $p$ is a distribution and $f$ is a non negative function.
\item $f(x;y)$: The main and secondary variables of the function are separated by a semicolon.
\item $\chr_A(x)$: The indicator function (also called the characteristic function) on the set $A$.
\item $\chr_\mathbf{P}:=\begin{cases}1, & \text{if $\mathbf{P}$ holds} \\ 0, & \text{otherwise} \end{cases}$: Another form of the indicator function, where $\mathbf{P}$ is a relational expression.
\item $\delta_{ij}:=\begin{cases}1, & i = j \\ 0, & i\neq j \end{cases}$: The Kronecker symbol, regarded as a binary function.
\item $\delta_{x}$: Dirac function/distribution.
\item $(x)_+$: The rectified linear function.
\item $\sign$: The sign function.
\item $\softmax$: The normalized exponential function.
\item $\matr{A}=\{a_{ij}\}$: $a_{ij}$ is the element in the $i$-th row and $j$-th column of the matrix $\matr{A}$.
\item $\|\matr{A}\|_F := (\sum_{ij}|a_{ij}|^2)^{1/2}$: The Frobenius norm of the matrix.
\item $\matr{A}\circ \matr{B},\matr{A}\oslash\matr{B}$: The pointwise (Hadamard) multiplication and division operations of matrices.
\end{enumerate}

The remainder of this paper is organized as follows. Section 1 introduces the basic concepts and principles of the localization method, including localization kernels, their operations, and smoothing spaces. Section 2 presents the principles of the localization model, covering kernel matrices, Laplacians, and the formal definition of local decision models. Supervised local models, such as local mean/regression and local mode/classification, are discussed in Section 3. Section 4 explores unsupervised local models, including kernel density estimation and self-local mean, with applications to clustering via the MeanShift algorithm. Local time series models and the connection to self-attention mechanisms are examined in Section 5. Section 6 provides examples of local models, including local linear regression, local classifiers, and local PCA. Adaptive localization methods, such as parametric kernels, multi-kernels, and discrete kernels, are detailed in Section 7. Section 8 covers advanced topics, including denoising autoencoders, score models, local mean diffusion models, hierarchical local models, non-local models, and graph theory-inspired kernels. Finally, Section 9 summarizes the paper and outlines future research directions. Appendices A - G offer supplementary material, including comparisons with kernel methods, discussions on center-based classifiers, lazy models, autoencoders, and categorical-style notation.

\section{Basic Concepts and Principles}

According to the above idea, the localization method proposes a new loss/risk function: for each data point $x_*$, construct a weighted loss function $\sum_iw_il(x_i,y_i,\theta)$, where $\{x_i\}$ is the sample, $w_i$ is the corresponding sample weight (localization weight), and $l(x,y,\theta)$ is a given loss function in advance.

\subsection{Localization Kernel}\label{sec:local-kernel}

Since the localization weight $w_i$ is related to $x_*$, we denote $w_i = K(x_*,x_i)$ and call the function $K:\mathcal{X}\times \mathcal{X}\to\R$ the localization kernel (function) on the sample space $\mathcal{X}$, which is simply referred to as the kernel (function) in the paper.

\subsubsection{Basic conditions and examples}

The conditions for constructing the kernel $K(x, x')$ are very weak:
\begin{enumerate}
  \item Non-negativity, that is, $K(x,x')\geq 0$, or more weakly, $\int K(x,x')\diff x'\geq 0$.
  \item Decrease with the increase of distance, that is, $K(x,x')$ is a decreasing function of $d(x,x')$, where $d$ is the distance on $\mathcal{X}$.
  \item Symmetry, that is, $K(x,x') = K(x',x)$.
\end{enumerate}

These three conditions do not have to be strictly adhered to and can be adjusted according to the actual situation. Conditional (2) reflects the characteristics of the localization kernel more. It indicates that the data point $x_i$ in the neighborhood of $x_*$ has a greater loss weight $K(x_*,x_i)$, and the machine learning algorithm will give priority to or only consider these points to achieve the purpose of locally fitting the data at $x_*$. Overall, the kernel $K$ characterizes the similarity between sample points.

The localization kernel and the kernel (positive definite kernel) used in kernel methods are two different concepts. An obvious difference is that the former does not require the kernel function to have symmetric positive definiteness. The two also have a large intersection. For example, the Gaussian function can be used as a positive definite kernel and also as a localization kernel. In fact, many convolution kernels are like this. For the theory of kernel methods, please refer to \cite{micchelli2006,hofmann2008}.

\begin{definition}
  A kernel in the form of $K(x,x')=K_1(x'-x)K_2(x')$ is called a relative kernel, where $K_1$ represents the relative part of the weight and $K_2$ represents the absolute part. Formally, it can be called a weighted convolution kernel.
\end{definition}

\begin{example}
  Compared with the Gaussian kernel, in localization methods, people often use the Epanechnikov kernel \cite{fan1992},
\begin{equation}\label{epanechnikov}
K_h(\vect{x},\vect{x}'):=D(\frac{\|\vect{x}-\vect{x}'\|}{h}), \vect{x},\vect{x}'\in\R^p
\end{equation}
where $D(t) := \frac{3}{4}(1-t^2)_+$, and $h>0$ is the dilation coefficient. A kernel with unknown parameters like the Epanechnikov kernel is called a parameter kernel (see \autoref{sec:param-local-kernel}).
\end{example}

\begin{example}\label{ex:hollow-kernel}
  A kernel is called the hollow kernel, if
$$K(x,x')=0 ~\forall x,x', d(x,x')< \epsilon ~\text{or}~ x=x'.$$
This kernel function is considered beneficial for the training phase of the model.
\end{example}

\begin{example}[Neighborhood Kernel]
  The neighborhood kernel is the simplest kernel:
$$K(x,x'):=\chr_{U(x)}(x')$$
where $U(x')$ is a suitable neighborhood of $x'$. Generally, the neighborhood is determined by a distance $d$, that is, $U(x)=\{x'~:d(x',x)<\epsilon\},\epsilon>0$, also called a spherical neighborhood. At this time, the neighborhood kernel is $K(x,x')=\chr_{d(x,x')<\epsilon}$. More generally, we can let $U(x)=\{x~:d(x,x')<\epsilon(x)\}$, that is, $\epsilon$ changes with the change of $x'$.
\end{example}

\begin{definition}[Probability Kernel]
  The binary joint distribution $p(x,x')$ or the conditional/transition distribution $p(x'|x)$ on $\mathcal{X}$ can be used as a kernel. This is also the probabilistic meaning of the kernel.

  Note that the concept of neighborhood in localization methods is not limited to conventional topology, but rather to the range of likely transitions of each point under the transition probability.
\end{definition}

\begin{example}\label{ex:func-kernel}
  The function-type kernel is defined as
$$K(x,x'):=\chr_{x'=f(x)}$$
where $f$ is a function from $\mathcal{X}\to \mathcal{X}$. The function-type kernel is also a degenerate probability kernel.
\end{example}

\begin{definition}[Discrete Kernel]
  A kernel defined on a discrete space (usually a finite set) $\mathcal{X}$ is a discrete kernel, which can usually be represented by a (non-negative) matrix of size $|\mathcal{X}|\times |\mathcal{X}|$.
\end{definition}

\begin{remark}\label{rm:kernel-topology}
  Do not forget that the concept of discreteness is related to topology. Since there is no conventional distance and neighborhood in the discrete space, the discrete kernel seems unable to reflect the original idea of localization. In fact, the topology of the discrete space should be defined by the kernel in reverse. This also implies that the discrete kernel should be obtained through learning (see \autoref{sec:adaptive-local}). Obviously, there are at most $|\mathcal{X}|^2$ adjustable parameters for the discrete kernel. Although the general kernel function represents a certain topology, it has no necessary connection with the conventional one.
\end{remark}

The closure of $\mathset{x}{f(x)>0}$ (or $\mathset{x}{f(x)>\epsilon}$ where $\epsilon$ is a small number) is called the \term{support}/\term{window} of the function $f(x)$; the sample subset $\mathset{x_i}{f(x_i)>0}$ is called the support sample of the function $f$.

One problem faced by the localization method is that if the support of the function $x\mapsto K(x_*,x)$ is set too small, the size of the corresponding support sample will be too small and overfitting will occur; if the support is set too large, the size will be too large and underfitting will occur \cite{fan1992,kohler2014,sestelo2017}. A simple strategy is to introduce a hyperparameter $h$ to regulate the size of the support of the kernel as the Epanechnikov kernel, which is called the \term{bandwidth} or \term{smoothing parameter}.

\subsubsection{Feature mapping of localization kernels}\label{sec:local-feature}

The concept of a ``feature mapping'' can also be applied to localization kernels.

\begin{definition}[Feature Mapping]\label{df:local-feature}
Let $H$ be a Hilbert space and $\mathcal{X}$ be the sample space. Construct the kernel
\begin{equation}\label{local-feature}
  K(x,y) = \langle\phi(x),\psi(y)\rangle,
\end{equation}
where $\phi,\psi:\mathcal{X}\to H$ are called the feature mappings of $K$, and $H$ is called the feature space. In particular, when $H$ is the Euclidean space $\mathbb{R}^d$, let $K(x,y)=\phi(x)\cdot\psi(y)$. Generally, $\phi$ and $\psi$ are required to be non-negative to ensure the non-negativity of $K$.
\end{definition}

\fref{local-feature} is only a reference for comparison with the feature mapping of kernel methods and is not the only form that the local feature mapping must adopt. An alternative formulation defines the kernel via
\begin{equation}\label{local-feature-exp}
  K(x,x') = \mathrm{e}^{\langle\phi(x),\psi(x')\rangle},
\end{equation}
or, more generally, $K(x,x')=F(\phi(x),\psi(x'))$, where $F$ is a well-defined binary function, called the \term{relation function} \cite{luong2015,oreshkin2018,sung2018}. Meanwhile, the feature space $H$ need not be restricted to a Hilbert space.

\autoref{df:local-feature} reveals another significant difference between localization methods and kernel methods: the localization kernel can correspond to two different feature mappings (which can be respectively called the ``query-feature mapping'' and the ``key-feature mapping'' of the kernel). Obviously, the feature mapping of the localization kernel also gives a real embedding of discrete variables (and general variables). \autoref{rm:kernel-topology} mentions that the kernel can be used to define the topology of the discrete space. Now, by embedding the discrete variables into $\mathbb{R}^d$ using the feature mapping, this can be easily achieved.

\begin{example}[Probability-Feature Mapping]
  Introduce the feature mapping $\phi(x) = p(z|x)$ with the probability distribution as the value. The corresponding kernels are $K(x,x')=\int p(z|x) p(z|x')\diff z$ or $\int p(z|x)p(x'|z)\diff z$.
\end{example}

\begin{example}[Word Embedding \cite{mikolov2013a,mikolov2013b}]\label{ex:wv}
According to the above discussion, we use two feature mappings $v_I,v_O$ (respectively called the input and output vector representations of $w$) to define the conditional word distribution:
 \begin{equation}\label{cbow}
  \ln p(w|c)\sim v_I(c) \cdot v_O(w)=\frac{1}{|c|}\sum_{w':c} v_I(w')\cdot v_O(w).
  \end{equation}
We can replace $\ln p(w|c)$ in \fref{cbow} with $\ln p(c|w)$ or $\ln p(w,c)$ as an altenative way to word embedding.
\end{example}

\subsubsection{Self-localization kernel}

Let $f:\mathcal{X}\to\mathbb{R}$. The localization kernel does not exclude the following form,
\begin{equation}\label{self-local-kernel-xy}
\int K(x,f(x),y,f(y))f(y)\dy,
\end{equation}
where $K$ is a kernel function defined on $\mathcal{X}\times \mathbb{R}$. However, the resulting operator (still denoted as $Kf$) is no longer linear.

To emphasize its difference from the ordinary localization kernel, we call the kernel in \fref{self-local-kernel-xy} the \term{self-localization kernel}. A relatively simple and common special form is
\begin{equation}\label{kernel-sep}
\int K_1(x,y)K_2(f(x),f(y))f(y)\dy.
\end{equation}

\begin{remark}
  \fref{self-local-kernel-xy} might perhaps be called the \term{SUSAN filter} or the \term{Bilateral filter} \cite{smith1997,tomasi1998}. The more original form can be traced back to \cite{yaroslavsky1985}. Moreover, \cite{buades2006} pointed out the connection between these filters and \term{partial differential equations (PDEs)}.
\end{remark}

\begin{remark}
  The kernel is capable of representing perceptual similarity. Consequently, the self-localizing kernel serves to characterize any manifestation of \term{Gestalt laws} \cite{wertheimer1923,zhang2024gestalt} in psychology.
\end{remark}

\subsection{Operations of Kernels}

As a binary function, the localization kernel $K$ can induce a linear operator (still denoted as $K$) on the function space:
\begin{equation}\label{loc-op}
Kf(x) := \int K(x,y)f(y)\dy
\end{equation}
In particular, when $K$ is a convolution kernel, \fref{loc-op} is the convolution operation, that is, $Kf = K*f$.

\subsubsection{Some simple operations}

\begin{definition}[Duality]
 The duality of $K$ is $K^*(x,y):=K(y,x)$. $K^*$ is called the \term{dual kernel} \wrt $K$.
\end{definition}

\begin{definition}[Product]
Corresponding to the product of operators, define $K'K(x,y):=\int K'(x,z)K(z,y)\diff z$, which is called the \term{product kernel}. Then define the power operation of the kernel, denoted as $K^n$, and called the \term{power kernel}.
\end{definition}

\begin{definition}[Regularization]\label{df:regularized-kernel}
Given a kernel $K$, define its the \term{regularized kernel} as $K'(x,y)=\alpha K(x,y)+(1-\alpha)\delta_{xy},0\leq \alpha\leq 1$.
\end{definition}

\subsubsection{Normalization of kernels and identity approximation}\label{sec:conv}

A kernel that satisfies the (asymmetric) normalization condition $\int K(x,y)\diff x = 1$ is called a (relative to $x$) \term{normalized kernel}. Obviously, any kernel $K$ that satisfies $\int K(x,y)\dx>0$ can be normalized to
\begin{equation*}
\tilde{K}(x,y) = K(x,y)/\int K(x,y)\diff x
\end{equation*}
Obviously, the normalization condition of the convolution kernel $K$ is $\int K = 1$.

\begin{remark}
$K(x,y)$ is asymmetric. The normalization should be distinguished as being \wrt~$x$ or $y$, but when there is no ambiguity, only normalization is mentioned.
\end{remark}

\begin{remark}\label{rm:normalized}
An ordinary non-negative kernel corresponds to a joint distribution. A non-negative normalized kernel corresponds to a conditional distribution/transition probability, that is, $p(x|y) = K(x,y)$, and a non-negative normalized convolution kernel corresponds to a probability density. Such a kernel is also called a \term{transition kernel}. Therefore, the power kernel $K^{n}$ is the $n$-th order transition probability/kernel $K^{n}$, and $K^{\infty}(\cdot,x)$ is the stationary distribution corresponding to this transition probability.
\end{remark}

A kernel with a dilation parameter like the Epanechnikov kernel can be called a \term{dilation kernel}. To maintain normalization, let
$$K_h(\vect{x}):=\frac{1}{h^p}K(\frac{\vect{x}}{h}),h>0.
$$

\begin{theorem}[Identity Approximation Principle \cite{zo1976}]\label{th:id-approx}
Let $f(\vect{x})$ be an integrable function on $\R^p$, and $K$ be a reasonable normalized convolution kernel, that is, $\int K = 1$, then
\begin{equation}\label{id-approx}
f(\vect{x})=\lim_{h\to 0}\int K_h(\vect{x}-\vect{y})f(\vect{y})d\vect{x}~~(\text{i.e.}~K_h * f(\vect{x}))
\end{equation}
where $K_h(\cdot)$ is the dilation kernel. (The conclusion holds for non-convolution kernels, but the convolution form is more common.)
\end{theorem}

\begin{remark}
\fref{id-approx} shows that the sequence $K_h$ (the corresponding convolution operator/kernel) converges in a sense to the \term{Dirac kernel} $\delta$ (corresponding to the identity operator) as the convolution identity. This is the origin of the term ``identity approximation''.
\end{remark}

\begin{remark}\label{rm:id-approx}
 The kernel $K$ has certain conditions, but they are very weak, and there is even no non-negativity requirement. Here, it is summarized only with the word ``reasonable''. Even $K_h(\cdot)$ may not be a dilation kernel, and the normalization condition can be weakened to $\int K_h\to 1,h\to 0$.
\end{remark}

The normalization condition $\int K = 1$ in \autoref{th:id-approx} ensures that the convolution has a smoothing effect. A kernel that satisfies this condition and has a smoothing effect is called a \term{smoothing kernel}. If $\int K = 0$, the convolution kernel may have a desmoothing (or edge extraction) effect and can be called a \term{desmoothing kernel}. The most typical examples are the \term{derivative operator/differential operator}, one of whose applications is the \term{image edge detection} \cite{aubert2006}.


\subsubsection{The Laplacian and difference of the kernel}

The following concepts are derived from the \term{Laplacian (matrix)} (see \autoref{df:laplacian-local}).

\begin{definition}[The Laplacian and Difference of the Kernel]
The binary function $L(x,y)=\delta_{xy}-K(x,y)$ is called the \term{Laplacian} of the kernel $K$. Obviously, the Laplacian of a smoothing kernel is a desmoothing kernel. Generally,
$K_1 - K_2$ is a desmoothing kernel if $K_1, K_2$ are smoothing kernels. We call it the difference of the kernel, or the \term{difference kernel}.
\end{definition}

\begin{example}
The Laplacian is a special difference kernel. Another famous example is the \term{difference of Gaussian (DoG)}, which is applied to image \term{saliency detection}\cite{achanta2010}. The DoG can be appropriately generalized: Let $K_h$ be a dilation kernel, then $K_{h_1}-K_{h_2},h_1>h_2$. Similar concepts can be found in \cite{ye2024}.
\end{example}

\begin{definition}[Differential Kernel]
Let $K_h,h\in\R$ be a parametric kernel. Then $\frac{\diff K_h}{\diff h}$ is called the \term{differential kernel}. The differential kernel is the limit form of the difference kernel. When $h\in\R^m$, $m$ partial derivative kernels and directional derivative kernels can be constructed, which are collectively referred to as differential kernels. The most classic example is the \term{Laplacian of the Guassian kernel (LoG)}\cite{marr1980,sotak1989}.
\end{definition}

\begin{definition}[total variation]
Let $L$ be a difference kernel or a differential kernel, then $\|Lf\|$ is called the \term{total variation} of the function $f$ under the action of $L$ (or the corresponding smoothing kernel), where $\|\cdot\|$ is a norm for functions. $|Lf(x)|$ is called the \term{saliency function}.
\end{definition}



\subsection{Smoothing Spaces}\label{sec:smooth-sp}

Positive definite kernels have corresponding \term{Reproducing Kernel Hilbert Spaces (RKHS)}, while localization kernels do not seem to have such spaces. Now we try to give an analogue of RKHS.

\begin{definition}[Smoothing Space]
Let $\mathcal{F}$ be a certain reasonable function space on $\mathcal{X}$, and its norm is denoted as $\|\cdot\|$. Without loss of generality, let $K$ be a normalized kernel and derive the operator $K:\mathcal{F}\to\mathcal{F}$. The set of functions that satisfy the following conditions is called the \term{smoothing space} and is denoted as $\mathcal{F}^1$:
\begin{equation}\label{smooth-func}
  \|f\|_{\mathcal{F}^1}:=\|Lf\|=\|f - Kf\| \leq C (\text{or} <\infty), f\in\mathcal{F}
\end{equation}
where $C\geq 0$ is given, and $Lf:=f - Kf$ is the Laplacian of $K$. The $\|f\|_{\mathcal{F}^1}$ in \fref{smooth-func} is called the \term{smoothing norm} (not a norm in the strict sense), that is, the total variation under the action of the Laplacian. The higher-order smoothing space can be further defined: $f\in\mathcal{F}^m$ ~\iffi~ $\|L^kf\|\leq C, k = 1,\cdots, m$.
\end{definition}

According to the discussion in \autoref{sec:conv}, \fref{smooth-func} indicates that $f\in\mathcal{F}^1$ is sufficiently close to its smoothed result $Kf$.

\begin{remark}
Another way to define the smoothing space is: $\mathcal{F}^m:=K^m(\mathcal{F})$, that is, the image of $\mathcal{F}$ under the action of $K^m$.
\end{remark}

\begin{definition}
The norm $\|f\|_{\mathcal{F}^1}$ induces the distance $d_K(f,g):=\|K(f - g)\|$, which is called the \term{smoothing distance}. If $K$ is not invertible as an operator, then $d_K$ is not a distance in the strict sense, but in practical applications, we do not need it to strictly satisfy the definition of distance. Currently, as an analogue of the \term{Maximum mean Discrepancy (MMD)}\cite{gretton2012}, it can be applied to distributions and also to general functions.
\end{definition}


\begin{example}[Filtering Problem]
Let $K$ be a normalized kernel. The approximate form of the \term{filtering problem} \cite{mumford1989,rudin1992} is
\begin{equation}\label{filter-problem}
  \min_{f\in\mathcal{F}^1}\|f - g\| + \lambda\|Lf\|, ~\lambda>0.
\end{equation}
That is, to find a smooth function $f\in\mathcal{F}^1$ to approximate the non-smooth function $g$. The denoising problem is a special form of the filtering problem, that is, the function to be approximated $g = f_0+\epsilon$, where $f_0\in\mathcal{F}_\epsilon$ is unknown and $\epsilon$ is noise. The filtering problem is often associated with dimensionality reduction, such as $f=\sum_k\alpha_kv_k$, where $\alpha_k\in\R, v_k\in\mathcal{F}^1$ are all unknown.
\end{example}

\section{Principles of the Localization Model}

We have made the theoretical preparations and now begin to introduce the basic principles of the localization tricks/models.

\subsection{Kernel Matrix and Laplacian}\label{sec:laplacian-loc}

Similar to kernel methods, the (local empirical) kernel matrix and the (local empirical) feature matrix are introduced as follows:
\begin{align*}
\matr{K} &:= K(\matr{X},\matr{X})=\{K(x_{i},x_{j})\},\\
\bmPhi &:= \phi(\matr{X})=\{\phi(x_{i})\},\\
\bmPsi &:=\psi(\matr{X})=\{\psi(x_{i})\},
\end{align*}
satisfying $\matr{K}=\bmPhi\bmPsi^{\mathrm{T}}$ (other forms are also allowed).

\begin{remark}\label{rm:kernel-graph}
Just as a positive definite kernel (or its kernel matrix) corresponds to a weighted undirected graph, a localization kernel (or its kernel matrix) corresponds to a weighted directed graph $G=(V,E)$, where $V=\{v_i\}$ is the set of vertices and $E$ is the adjacency matrix. In fact, the weighted directed graph $G=(V,E)$ can be represented by the kernel matrix $\matr{K}=K(\matr{X},\matr{X})$, where $\matr{X}=V$ and $\matr{K}_{ij}=E_{ij}$. Such kernel (matrix), without a specific expression for the kernel function, is understood as a discrete kernel (matrix), or called a \term{concrete kernel}\cite{meng2022}.
\end{remark}

The normalization of the kernel matrix is $\tilde{\matr{K}}:=\matr{D}^{-1}\matr{K}$, where $\matr{D}:=\diag\{\matr{K}\one_N\}$ is the degree matrix of the localization kernel, and related to the \term{(empirical) mean embedding}.

\begin{remark}
  The kernel matrix of a normalized kernel is not necessarily normalized. However, according to the law of large numbers, if the kernel function is normalized under the measure $p(x)\diff x$, its kernel matrix is approximately normalized (up to a constant), where $x_i \sim p(x)$. This fact can be used to avoid normalization operations.
\end{remark}

\begin{definition}[Laplacian]\label{df:laplacian-local}
The localization kernel matrix (or any matrix) $\matr{K}$ can derive the \term{Laplacian},
\begin{equation*}
\matr{L}:=\matr{D}-\matr{K},
\end{equation*}
and $\tilde{\matr{L}}:=\eye-\tilde{\matr{K}}$ is its (asymmetric type) normalized form, where $\tilde{\matr{K}}=\matr{D}^{-1}\matr{K}$ is the normalization of $\matr{K}$.
\end{definition}

\begin{remark}
The smoothing kernel $K(x',x)$ always has a corresponding desmoothing kernel $L:=\delta-K$, that is, the Laplacian; the discrete form of the smoothing kernel is (approximately) the normalized kernel matrix, and the discrete form of the desmoothing kernel is the normalized Laplacian (the normalization here is \wrt~$x$).
\end{remark}

\subsection{Local Decision Model}\label{sec:local-model}

We first give a general model, covering all localization models.
\begin{definition}[Local Decision]\label{df:local-decision}
  Let $l(x,\theta)$ be the loss function of a single sample point on $\mathcal{X}$. The \term{local loss} at the target point $x_*\in\mathcal{X}$ is defined as the following weighted empirical risk,
  \begin{equation}\label{local-loss}
  J(x_*,\theta) := \sum_i K(x_*,x_i) l(x_i,\theta),
  \end{equation}
  where $K(\cdot,\cdot)$ is the kernel, and $\{x_i\}$ is the set of samples. The \term{local decision} refers to the optimization problem related to the local loss function, which is
  \begin{equation}\label{local-dec}
  \argmin_\theta J(x_*,\theta),
  \end{equation}
  and its solution $\hat{\theta}(x_*)$ is called the \term{local parameter estimate}.
\end{definition}

In local decision, each parameter estimate $\hat{\theta}(x_*)$ is only valid for a given target point $x_*$, and the optimal loss value is $J(x_*,\hat{\theta}(x_*))$. The mapping $x \mapsto J(x,\hat{\theta}(x))$ is the loss function for a single sample point. Based on this, we define the total loss of local decisions on the sample $\{x_i\}$ as
\begin{equation}\label{local-kernel-loss}
  J(K) := \sum_i J(x_i,\hat{\theta}(x_i;K)),
\end{equation}
where the expressions $J(K), \hat{\theta}(x_i;K)$ emphasize their dependence on $K$. When we choose an appropriate kernel function, we should use \fref{local-kernel-loss} as the evaluation criterion. Clearly, the computation of $J(K)$ ultimately depends only on the kernel matrix $\{K(x_i,x_j)\}$.

\begin{remark}
  The local loss can also have a more general definition:
  \begin{equation*}
  J(x_*,\theta) := \sum_i l(x_*,x_i,\theta),
  \end{equation*}
  where the loss function $l(x_*,x,\theta)$ depends on the target point $x_*$. This means that different loss functions are defined at different target points.
\end{remark}

\begin{remark}[Global Parameters]\label{rm:global-param}
  A global parameter/hyperparameter $\alpha$ can be introduced for the local decision, and the total loss is defined as
  \begin{equation}\label{loss-global-param}
  J(K,\alpha) := \sum_i J(x_i,\hat{\theta}(x_i;K,\alpha),\alpha) ,
  \end{equation}
  where $\hat{\theta}(x_*;K,\alpha)$ is the solution to the solution to the following optimization problem:
  \begin{equation}\label{min-local-global-param}
  \min_\theta J(x_*,\theta,\alpha) := \sum_i K(x_*,x_i) l(x_i,\theta,\alpha) .
  \end{equation}
  The kernel $K$ itself is also considered as such a hyperparameter.
\end{remark}

When $l(x_i,\theta)$ is the negative log-likelihood $-\ln p(x_i|\theta)$, the \term{local likelihood} and \term{local (maximum) likelihood estimate} are obtained \cite{hjort1996}.

\begin{definition}[Local Likelihood]\label{df:local-likelihood}
  The local likelihood at the target point $x_*\in\mathcal{X}$ is defined as the following weighted likelihood,
  \begin{equation}\label{local-likelihood}
  l_K(x_*,\theta) := \sum_i K(x_*,x_i)\ln p(x_i|\theta) ,
  \end{equation}
  where $K(x_*,x)$ is the kernel, and $\{x_i\}$ is the sample. The local likelihood estimate is the maximization problem for the local likelihood, i.e., $ \max_\theta l_K(x_*,\theta)$.
\end{definition}

\begin{definition}\label{rm:local-emp-distr}
  The weighted empirical distribution $\sum_i K(x_*,x_i) \delta_{x_i}$ (normalized) can be called the \term{local empirical distribution} at $x_*$.
\end{definition}

Clearly, the local likelihood (\fref{local-likelihood}) is the cross-entropy between the local empirical distribution and the model distribution. Since $K(x_*,x_i)$ represents a transition distribution, the local empirical distribution is also referred to as the empirical transition distribution.

\begin{remark}\label{df:additive-stat}
More generally, a statistic of the form $\sum_i \phi(x_i)$ is called the \term{additive statistic}, where $\phi$ is any function applied to each point $x_i$. The corresponding \term{local additive statistic} can be $\sum_i K(x_*, x_i) \phi(x_i)$.
\end{remark}

We now explicitly define the terms \term{localization trick/technique} and \term{local model}.

\begin{definition}[Localization Trick]
We also say that \fref{local-loss} is the localization of the decision model $\min_\theta \sum_i l(x_i,\theta)$, or equivalently, applying the localization trick to this decision model. The localization of an certain model is called a local certain model, collectively referred to as \term{local models}.
\end{definition}

\begin{definition}[Localization Kernel Model]
  If a model is essentially based on localization kernels or kernel matrices (whether or not derived by applying a localization trick to an existing model), it is called a \term{localization kernel model}, also referred to as a local model or kernel model.
\end{definition}

\begin{history}
Localization methods have a long history, dating back at least to the introduction of \term{smoothing methods} \cite{macaulay1931, ezekiel1941}. \cite{cleveland1979, cleveland1988} presents early comprehensive studies on local fitting. The concept of local likelihood was systematically discussed in \cite{tibshirani1987}; the \term{Vicinal Risk Minimization (VRM)} principle was formally introduced in \cite{vapnik1999, chapelle2000}. Most studies were limited to the regression form and did not abstract out a general form like \fref{local-likelihood}, nor did they introduce the concept of local decisions. \cite{loader1996} introduced local density estimation and provided non-parametric and polynomial approximations of \fref{local-likelihood}.
\end{history}

\subsubsection{Neighborhood decision}

Neighborhood decision is the most primitive form of local decision, reflecting the motivation behind the localization method.

\begin{definition}[Neighborhood Loss / Neighborhood Decision]\label{df:neighbor-kernel}
  If a neighborhood kernel $K(x_*,x) = \chr_{U(x_*)}(x)$ is chosen for local decision, the \term{neighborhood loss} is given by
  \begin{equation}
  J(x_*,\theta) := \sum_{x_i \in U(x_*)} l(x_i,\theta) ,
  \end{equation}
  where $U(x_*)$ is a suitable neighborhood of $x_*$. The local decision derived from the neighborhood loss is called the \term{neighborhood decision}.
\end{definition}

In real applications, the neighborhood $U(x_*)$ is typically determined by a distance. The neighborhood decision is also determined by this distance, and hence is a distance-based model.

\subsubsection{Nearest neighbor decision}

The \term{(nearest) neighbor} method is the ancestor of localization methods. On the surface, it appears to be a neighborhood decision, but there are subtle differences between the two.

\begin{definition}[Nearest Neighbor Decision]\label{df:nnd}
  Let the sample be $\samp{X} = \{x_i, i = 1, \cdots, N\}$. The \term{nearest neighbor loss} is generally defined as
  \begin{equation}\label{nnd}
  J(x_*,\theta) := \sum_{x_i \in \mathcal{N}_K(x_*, \samp{X})} l(x_i,\theta) ,
  \end{equation}
  where $\mathcal{N}_K(x_*, \samp{X})$ is the subset of the $K \leq N$ nearest points to $x_*$ (the \term{nearest neighbors}) from the sample $\samp{X}$ under some metric. The corresponding local decision is called the \term{nearest neighbor decision}.
\end{definition}

You will notice that the neighborhood for nearest neighbor loss depends on the entire sample $\samp{X}$, that is an empirical/adaptive neighborhood. The corresponding kernel is called the \term{nearest neighbor kernel}. Based on this, we give the abstract form of the \term{empirical (localization) kernel}.

\begin{definition}[Empirical Kernel]
Building on \fref{nnd}, we introduce
\begin{subequations}
\begin{align}\label{emp-kernel}
  J(x_*,\theta;\samp{X}) &:= \sum_{i} K(x_*,x_i;\samp{X}) l(x_i,\theta) , \\
  \text{or} &:= \sum_{i} K_i(x_*;\samp{X}) l(x_i,\theta) ,
\end{align}
\end{subequations}
where the kernel function $K$ (or the family of functions $K_i$) depends on the sample $\samp{X}$, and thus is an empirical kernel.
\end{definition}

Thus, nearest neighbor decision is a local model based on the empirical kernel. Some of the conclusions in the paper depend on the concept of empirical kernels, but the main focus is on non-empirical kernels.

\subsection{Monte Carlo Method}\label{sec:local-monte-carlo}

By the law of large numbers, we have
\begin{subequations}\label{local-loss-exp}
  \begin{align}
  J(x_*,\theta) & \approx \Exp_{x \sim q} K(x_*, x) l(x, \theta) \\
  & \approx \Exp_{x \sim p(x|x_*)} q(x) l(x, \theta), ~ p(x|x_*) \sim K(x_*, x),
  \end{align}
\end{subequations}
as the expectation form of \fref{local-loss}, where $q$ represents the true distribution of the data. Consequently, we have the corresponding Monte Carlo/resampling estimate:
\begin{subequations}\label{local-mc}
\begin{align}
   J(x_*,\theta) & \approx \sum_{x_i \sim p(x|x_*)} q(x_i) l(x_i, \theta) \label{local-mc-a} \\
   & \approx \sum_{x_i \sim p(x_i|x_*)} l(x_i, \theta), ~ p(x|x_*) \sim K(x_*, x), \label{local-mc-b}
\end{align}
\end{subequations}
where \fref{local-mc-a} samples from the sample space, while \fref{local-mc-b} samples (or resampling) from the sample itself.

The Monte Carlo estimate applied in this manner is referred to as the \term{Monte Carlo localization method}. By controlling the number of samples, \fref{local-mc} can be applied to large data problems. In particular, $J(x_*, \theta) \approx l(x_i, \theta), x_i \sim p(x|x_*)$ is called the \term{stochastic localization method}.

\section{Supervised Local Models}

Compared to general decision models, our primary focus is on supervised learning on $\mathcal{X} \times \mathcal{Y}$, especially machine learning models determined by functions of the form $y \sim f(x)$.
Here, we will express the general localized form of supervised learning models.

\subsection{Common Forms of Supervised Learning}

The general form of the local loss function for supervised models is given by \cite{tibshirani1987, fan1998, eguchi2003}.

\begin{definition}[Local Loss Function for Supervised Learning]\label{df:sl-local}
  Let $l(x_i, y_i, \theta)$ be the loss function for a supervised learning model on $\mathcal{X} \times \mathcal{Y}$. The local loss at point $x_* \in \mathcal{X}$ is defined as
  \begin{equation}\label{local-loss-x}
  J(x_*, \theta) := \sum_i K(x_*, x_i) l(x_i, y_i, \theta) ,
  \end{equation}
  which is the weighted loss function $K(x_*, x) l(x, y, \theta)$ corresponding to the empirical risk, where $K(x_*, x)$ is a kernel on $\mathcal{X}$. The corresponding local decision is called \term{local supervised learning}.
\end{definition}

\begin{remark}\label{rm:sl-kernel}
  For some purposes, we can use
  \begin{equation}\label{local-loss-xy}
  J(x_*, y_*, \theta) := \sum_i K(x_*, y_*, x_i, y_i) l(x_i, y_i, \theta) ,
  \end{equation}
  as a local loss, where $K(x_*, y_*, x, y)$ is a kernel on $\mathcal{X} \times \mathcal{Y}$. Clearly, \fref{local-loss-xy} satisfies the form of local decisions and is the sample form of the self-localization kernel in \fref{self-local-kernel-xy}.
  The technical issue here is that, for a new sample point $x_* \notin \{x_i\}$, $y_*$ is unknown, so the value of $K(x_*, y_*, x_i, y_i)$ cannot be determined. The solution is simple: treat $y_*$ as an unknown parameter and estimate it.
\end{remark}

Once the parameters are known, prediction no longer requires the kernel function, and predictions can be made directly using the pre-defined loss function $l$:
\begin{equation}
  \hat{y}(x_*) := \min_y l(x_*, y, \hat{\theta}(x_*)) ,
\end{equation}
where $\hat{\theta}(x_*)$ is the local parameter estimate.

\begin{definition}[Localization of Machine Learning Models]
  Given a machine learning model $y \sim f(x, \theta): \mathcal{X} \to \mathcal{Y}$, with loss function $l(y, \hat{y})$, its localization is the local decision of the following form,
  \begin{equation}\label{local-machine}
  \hat{\theta}(x_*) := \argmin_{\theta} \sum_i K(x_*, x_i) l(y_i, f(x_i, \theta)) ,
  \end{equation}
  with the prediction formula $\hat{y}(x_*) = f(x_*, \hat{\theta}(x_*))$.
\end{definition}

\begin{example}[Local Linear Regression]\label{ex:local-linear-regression}
  The first local machine learning model considered is naturally \term{local linear regression}. It is the localization of the linear regression model $f(\vect{x}, \vect{\beta}) = \vect{x} \cdot \vect{\beta}$. Using the squared loss function, we have
  \begin{equation}\label{local-linear-est}
  \hat{\vect{\beta}}(\vect{x}_*) = \argmin_{\vect{\beta}} \sum_i K(\vect{x}_*, \vect{x}_i) (y_i - \vect{x}_i \cdot \vect{\beta})^2 ,
  \end{equation}
  and its prediction formula is
  \begin{equation}\label{local-linear-predict}
  \hat{y}(\vect{x}_*) = \vect{x}_* \cdot \hat{\vect{\beta}}(\vect{x}_*)
  \end{equation}
\end{example}

\begin{figure}[H]
  \centering
  \includegraphics[width=\midwidth]{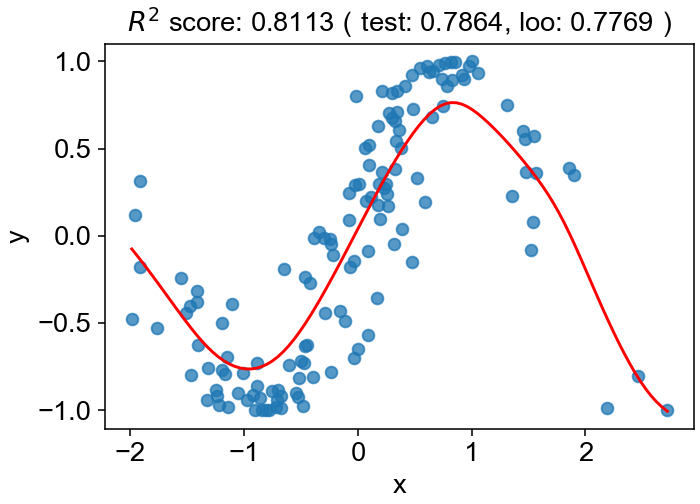}
   \caption{The fitting results of local linear regression (\texttt{test} means the $R^2$ score of test data; \texttt{loo} means the $R^2$ score by leave-one-out)}
  \label{fig:local-linear-regression-1d}
\end{figure}

From the form of \fref{local-linear-predict}, we find that applying the localization technique to a linear model is equivalent to fitting the following non-linear regression,
\begin{equation}
  y \sim \hat{\vect{\beta}}(\vect{x}) \cdot \vect{x} ,
\end{equation}
where $\hat{\vect{\beta}}(\vect{x})$ is an unknown function-like parameter, and \fref{local-linear-est} provides its estimate. \autoref{fig:local-linear-regression-1d} shows the data fitting effect of local linear regression.

\begin{remark}
  Local linear regression was originally called \term{local kernel regression}, which is one of the most classic local models \cite{fan1992}.
\end{remark}

\subsection{Local Mean}\label{sec:local-mean}

\begin{remark}
For a continuous output space $\mathcal{Y}$, it suffices for it to be closed under (non-negative) linear operations. Without loss of generality, we assume $\mathcal{Y}=\mathbb{R}$.
\end{remark}

The general form of local regression was presented in the previous section. In this section, we take a different perspective on local regression and introduce the important concept of \term{local mean}.

For any regression model $P(y | \vect{x}), y \in \mathbb{R}$, the following prediction formula holds:
\begin{equation}\label{rf}
  \hat{y}(\vect{x}) = \Exp(y|\vect{x}) = \int y p(y|\vect{x}) \diff y .
\end{equation}
Now, suppose $\{(\vect{x}_i, y_i)\}$ is the sample. We replace the conditional distribution $p(y|\vect{x})$ in \fref{rf} with the local empirical distribution $\sum_i K(x, x_i) \delta_{y_i}$ (see \autoref{rm:local-emp-distr}) to obtain \cite{hastie2009}
\begin{equation}\label{rf-sample}
  \hat{y}(\vect{x}) \approx \frac{\sum_i K(\vect{x}, \vect{x}_i) y_i}{\sum_i K(\vect{x}, \vect{x}_i)} .
\end{equation}


For neighborhood kernels, this substitution has a solid theoretical foundation, as shown by the following theorem.

\begin{theorem}[Lebesgue Differentiation Theorem]\label{th:cond-exp}
  Let $\mathcal{X} \subset \mathbb{R}^p, \mathcal{Y} = \mathbb{R}$, then
  \begin{align}\label{cond-exp}
  \Exp(y|\vect{x}_*) &= \lim_{U(\vect{x}_*) \to \vect{x}_*} \Exp(y|\vect{x} \in U(\vect{x}_*)) \nonumber \\
  &\approx \frac{1}{M}\sum_{\vect{x}_i \in U(\vect{x}_*)} y_i ,
  \end{align}
  where $U(\vect{x}_*)$ is a sufficiently small neighborhood of $\vect{x}_*$, and $M$ is the size of the subsample $\{\vect{x}_i \in U(\vect{x}_*)\}$ that falls within $U(\vect{x}_*)$.
\end{theorem}

\subsubsection{Definition of Local Mean}

\fref{rf-sample} constructs a non-parametric/lazy regression model, which is the local mean.

\begin{definition}[Local Mean/Regression]\label{df:local-ave}
The local mean model is directly determined by the following prediction function:
\begin{equation}\label{local-ave}
  \hat{y}(x_*) := \frac{\sum_i K(x_*, x_i) y_i}{\sum_i K(x_*, x_i)} ,
\end{equation}
where $K$ is the kernel, and $\{(x_i, y_i)\}$ is the sample. This is clearly a weighted form of \fref{cond-exp}. (Here, the input space $\mathcal{X}$ is not explicitly restricted, while the output space $\mathcal{Y}$ allows linear operations.)
\end{definition}

\begin{figure}[H]
  \centering
  \includegraphics[width=0.2\textwidth]{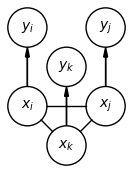}
  \caption{Local mean probability graph (This is not a strict probabilistic graph, in that it combines both directed and undirected graphs, with the bottom being a pairwise \term{Markov network/random field}. For the case of the self-localizing kernel, any two nodes must be connected.)}
  \label{fig:local-mean-pgm}
\end{figure}

To visually illustrate the model components, we provide a probability graph for local mean in \autoref{fig:local-mean-pgm}.

\begin{remark}
  For large data problems, we recommend using the Monte Carlo method mentioned in \autoref{sec:local-monte-carlo} \cite{szummer2001}:
  \begin{equation}\label{local-mean-mc}
  \sum_{x_i \sim p(x_i|x_*)} y_i, \quad p(x|x_*) \sim K(x_*, x) .
  \end{equation}
\end{remark}

\begin{fact}
Local mean is the local decision parameter estimation under the squared loss $(y_i - \theta)^2$ (assuming $\mathcal{Y} = \mathbb{R}$), i.e.,
\begin{equation*}
 \min_\theta \sum_i K(x_*, x_i) (y_i - \theta)^2.
\end{equation*}
\end{fact}

This shows that the local mean is essentially a local constant regression, and any function can be approximated as constant within a local region. Therefore, we have the following conclusion.

\begin{fact}\label{fc:local-ave}
  The localization of regression models is essentially local mean; local mean is universal.

  More precisely, the local regression model based on kernel $K$ can be converted into a local mean based on (empirical) kernel $L$. We refer to $L$ as the \term{equivalent (empirical) kernel}.
\end{fact}

\begin{remark}
  From the perspective of minimizing squared loss, the kernel matrix for local mean should be normalized, i.e., \fref{local-ave}, but as a regression model, local regression does not have this strict requirement.
\end{remark}

\begin{figure}[H]
  \centering
  \includegraphics[width=\midwidth]{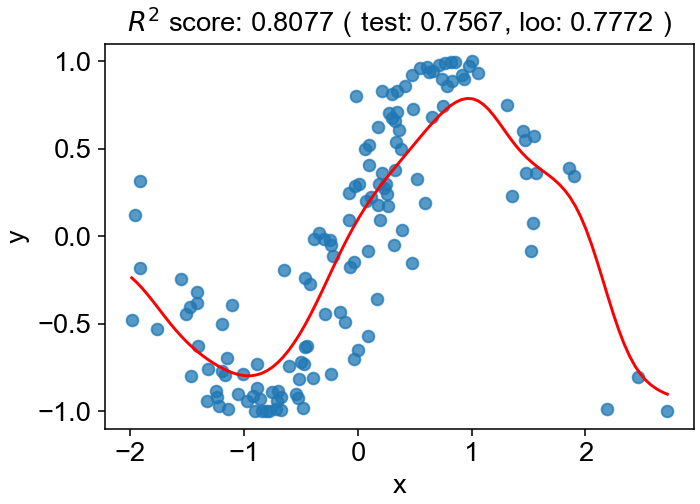}
   \caption{Fitting results of the local mean (with the same data as above)}
  \label{fig:local-mean-regression-1d}
\end{figure}

\autoref{fig:local-mean-regression-1d} shows that, with an appropriate choice of kernel, the local mean can achieve the same fitting effect as local linear regression.

\begin{history}
  \fref{local-ave} is also called the \term{Nadaraya-Watson Kernel Weighted Average} \cite{hastie2009, tosatto2021}. The corresponding regression model/method is called \term{Nadaraya-Watson regression}. This concept was first introduced in 1964 by E. A. Nadaraya and J. S. Watson \cite{nadaraya1964, watson1964}. Similar expressions also appeared in other fields \cite{savitzky1964,franke1980}.
\end{history}

Finally, we find that the local mean is the sample form of the smoothing effect of a normalized kernel, see the following conclusion shows.

\begin{fact}\label{ft:smooth-regression}
Assuming the sample comes from the model $y = f(x)$, when $K(x_*, x)$ is a (relative to $x$) normalized kernel, we have
\begin{equation}\label{op-reg}
\hat{y}(x_*) \approx (Kf)(x_*).
\end{equation}
In particular, when $K(\cdot)$ is a convolution kernel (and $\int K = 1$), the local mean is exactly the convolution of $K$ and $f$, i.e.,
\begin{equation}\label{conv-reg}
\hat{y}(x_*) \approx (K * f)(x_*).
\end{equation}
\end{fact}

\subsubsection{Local Mean Error}

As a lazy predictor, $\hat{y}(\cdot)$ of \fref{local-ave} can be applied to a sample $\matr{X}' = \{x_i'\}$, resulting in a compact matrix form:
\begin{equation}\label{local-ave-matrix}
  \hat{\vect{y}} = \tilde{\matr{K}} \vect{y} ,
\end{equation}
where $\hat{\vect{y}} = \{\hat{y}(x_i)\}, \vect{y} = \{y_i\}$, and the stochastic matrix $\tilde{\matr{K}}$ is the row-normalization of kernel matrix $K(\matr{X}', \matr{X})$.

The training error for local mean in \fref{local-ave-matrix} is
\begin{align}\label{local-ave-loss}
  J(K) = & \sum_i \left| y_i - \frac{\sum_j K(x_i, x_j) y_j}{\sum_j K(x_i, x_j)} \right|^2 \nonumber \\
  = & \|\vect{y} - \tilde{\matr{K}} \vect{y}\|_2^2 = \|\tilde{\matr{L}} \vect{y}\|_2^2 ,
\end{align}
where the symbols have the same meaning as in \fref{local-ave-matrix}, and $\tilde{\matr{L}}$ is the corresponding Laplacian. We can call \fref{local-ave-loss} the \term{local mean error} as a reference formula for evaluating and optimizing the kernel.

However, this error can lead to overfitting. To address this, we construct a loss function based on \term{leave-one-out}:
\begin{align}\label{local-ave-loo}
  J_{\mathrm{loo}}(K) := & ~ \sum_i \left| y_i - \frac{\sum_{j \neq i} K(x_i, x_j) y_j}{\sum_{j \neq i} K(x_i, x_j)} \right|^2 \nonumber \\
  = & ~ \|\vect{y} - \tilde{\matr{K}}^\circ \vect{y}\|_2^2 ,
\end{align}
where $\tilde{\matr{K}}^\circ$ is the normalization of the \term{hollow matrix} $\matr{K}^\circ := \matr{K} - \diag\{\matr{K}_{ii}\}$.

\subsubsection{Local mean dimensionality reduction}

Clearly, the local mean itself is a linear lazy transformation. Suppose $y \in \mathbb{R}^r$. As discussed in \autoref{sec:lazy-pred}, local mean can be derived as a dimensionality reduction method from $\mathcal{X}$ to $\mathbb{R}^r$, or an $r$-dimensional embedding, which can be referred to as \term{local mean dimensionality reduction (embedding)}. This is essentially the \term{Local Linear Embedding (LLE)} \cite{roweis2000}, as seen in \autoref{df:lle} and \autoref{ob:lle}. \autoref{fig:lle} is actually obtained through lazy transformation-iteration, not by the standard LLE algorithm.

\subsubsection{Query model}

Looking at \fref{local-ave}, we observe that the local mean has the same form of prediction function as kernel regression\cite{saunders1998}, but with fewer design requirements for the kernel function. We can also interpret \fref{local-ave} as a linear regression with basis functions $\phi_i(x_*) := K(x_*, x_i)/\sum_i K(x_*, x_i)$:
\begin{equation}\label{local-ave-span}
  \hat{y}(x_*) := \sum_i y_i \phi_i(x_*) .
\end{equation}
Here, the coefficients $y_i$ are exactly the output values.

Compared to the kernel regression or the \term{Support Vector Machines (SVMs)}, the advantage of \fref{local-ave-span} is that it does not require calculating the generalized inverse of the kernel matrix, only normalization. Additionally, \fref{local-ave-span} is easier to interpret: the more similar $x_*$ is to $x_i$, the more likely its predicted value will be $y_i$. This is similar to keyword search in databases or networks, so we colloquially call such models \term{query models} (or \term{matching/retrieval models} \cite{vinyals2016}). \fref{local-ave-span} is undoubtedly the simplest form of query model. More generally, we define
\begin{equation*}
  \hat{y}(x_*) := T(\{K(x_*, x_i)\}, \{y_i\}) .
\end{equation*}
where $T$ is a reasonable query operation.

\subsubsection{Inference model}

Introducing feature mappings, for example $K(x_*, x) = \phi(x_*) \cdot \psi(x)$, we construct the following regression model:
\begin{equation}\label{feature-ave-span}
  \hat{y}(x_*) := \frac{\phi(x_*)^{\mathrm{T}} \bmPsi^{\mathrm{T}} \vect{y}}{\phi(x_*)^{\mathrm{T}} \bmPsi^{\mathrm{T}} \one}, \quad \bmPsi = \psi(\matr{X}) .
\end{equation}
This means that, if a feature mapping is given, we can precompute $\bmPsi^{\mathrm{T}} \vect{y}$ and $\bmPsi^{\mathrm{T}} \one$ as part of the model training process, and then predict the output at the target point using \fref{feature-ave-span}.

As a regression model, the denominator in \fref{feature-ave-span} could be ignored. Then \fref{feature-ave-span} becomes a linear regression based on $\phi(x_*)$, and the estimation of coefficients is $\bmPsi^{\mathrm{T}} \vect{y}$. We call it the \term{local mean linear regression model}, which encourages the use of normalized kernels.

Unlike query models, in \fref{feature-ave-span}, the samples $\matr{X} = \{x_i\}$ and $\vect{y} = \{y_i\}$ are combined into a \term{inference rule system}, symbolically represented as
$$\{x_i \to y_i\} := \sum_i y_i \psi(x_i) / \sum_i \psi(x_i),$$
which is then applied to $x_*$ to yield the prediction. We call this an \term{inference model} (or \term{message-passing model} \cite{velickovic2022}). The general form of an inference model is $y_* \sim T(x_*, \{x_i \to y_i\})$, where $T$ is a reasonable reasoning operation.

\begin{figure}[H]
  \centering
  \includegraphics[width=0.35\textwidth]{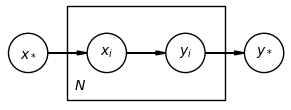}
  \caption{Inference model schematic (This is not a strict probability graph, but intended to show the rough structure of the model.)}
  \label{fig:infer-model}
\end{figure}

We represent an inference model visually in \autoref{fig:infer-model}.

In fuzzy mathematics, the \term{membership function} in mathematics determines the degree of membership of an element to a fuzzy set, quantifying the extent of its belonging. Another fuzzy mathematical concept is the \term{fuzzification} $\mu_{x}(x_*)$, which maps $x_*$ to a membership function. Roughly speaking, $\mu_{x}(x_*)$ represents the degree to which $x$ approaches $x_*$, such as $(1 - |x - x_*|)_+$, and can be represented by a kernel.

\begin{definition}[Fuzzy Inference \cite{jang1993,mendel1995}]\label{df:fuzzy}
\term{Fuzzy inference} can be understood as a mapping from $\mu_x(x_*)$ to $\mu_y(y_*)$:
\begin{subequations}\label{fuzzy-inf}
\begin{align}
  \mu_y(y_*) &:= T'\big(\mu_x(x_*), S(\{\mu_{x \to y}(x_i, y_i)\})\big), \\
  \mu_{x \to y}(x_i, y_i) &:= T(\mu_x(x_i), \mu_y(y_i)),
\end{align}
\end{subequations}
where $S, T, T'$ are reasonable operations. The final output of the system is the \term{defuzzification} of $\mu_y(y_*)$. A simple design scheme is
\begin{equation}\label{fuzzy-inf-simple}
  \mu_{x \to y}(x_i, y_i) := y_i \mu_x(x_i),
\end{equation}
meaning that $y$ is not fuzzified (nor needs to be defuzzified), and $S, T$ simplify to linear combinations.
\end{definition}

By comparison, local mean (in \fref{feature-ave-span}) is a special form of \fref{fuzzy-inf-simple}. The kernel regression also exhibits both query model and inference model forms. This inspires us to generalize the expression of local mean:
\begin{equation}\label{local-mean-general}
  \hat{y}(x_*) := S\left(\{T(K(x_*, x_i), y_i), i = 1, \cdots, N\}\right),
\end{equation}
where $T, S$ are reasonable operations.

If $S(\{\mu_{x \to y}(x_i, y_i)\})$ has a reasonable form, such as \fref{feature-ave-span}, then the inference model will be easier to implement for incremental learning.

\begin{remark}
  It is easy to find the connection with the \term{rough set} approach\cite{pawlak1998,polkowski2002}. Because the kernel function itself can induce a similarity relation: if $K(x,y)<\epsilon$, then $x$ is similar to $y$, where $\epsilon>0$ is a small number. This similarity relation can induce an equivalence relation on the set of sample points (usually not on the sample space), and further construct rough sets and the corresponding decision algorithms.
\end{remark}

\subsubsection{Self-localization kernel local mean}

Based on \autoref{rm:sl-kernel}, we can also apply the self-localization kernel $K(x_*, y_*, x, y)$ to local mean, namely
\begin{equation}\label{self-lazy}
  \hat{y}(x_*) = \frac{\sum_i K(x_*, y_*, x_i, y_i) y_i}{\sum_i K(x_*, y_*, x_i, y_i)}.
\end{equation}
\fref{self-lazy} provides lazy prediction $x_* \mapsto \hat{y}(x_*)$, but it is no longer linear. It is an \term{implicit prediction} that can be done through \term{prediction iteration} (fixed-point iteration concerning $y_*$), where the initial value can be set to the ordinary local mean (also see \autoref{rm:sl-kernel}).

\begin{figure}[H]
  \centering
  \includegraphics[width=\midwidth]{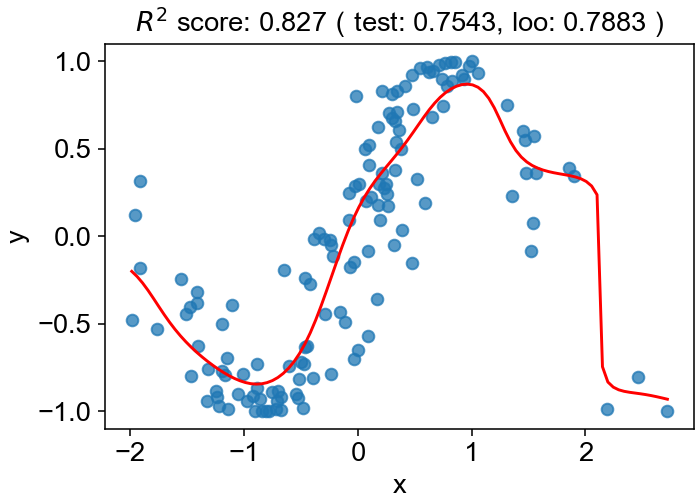}
   \caption{The of the local mean with self-localization kernel(same data as above)}
  \label{fig:self-local-mean-regression-1d}
\end{figure}

Intuitively, the self-localization kernel-local mean can progressively improve the prediction results of the ordinary local mean, \asref{fig:self-local-mean-regression-1d} (compare with \autoref{fig:local-mean-regression-1d}).

\subsection{Local Mode}

The definition of \term{local mode/classification} is entirely analogous to that of local mean/regression.

\begin{definition}[Local Mode/Classification]
  A local mode/classification model is determined by the following decision function:
  \begin{equation}\label{local-mode}
  \delta_k(x_*):=\sum_{y_i=k}K(x_*,x_i), ~k=1,\cdots, K,
  \end{equation}
  where $K$ is the kernel, and $\{(x_i,y_i)\}$ is the set of samples. The final local mode is given by $\argmax_k \delta_k(x_*)$.
\end{definition}

\begin{remark}
  Another definition of local classification is \cite{szummer2001}
  $$
  p(k|x_*) \approx \sum_k p(x_i|x_*) p(k|x_i), ~p(x_*|x_i) \sim K(x_*,x_i).
  $$
  That is, the response probability of $x_*$ is the local mean of all sample response probabilities.
\end{remark}

\begin{fact}
\fref{local-mode} is equivalent to the local mean of $y$ under one-hot encoding (a $K$-dimensional output). Moreover, local mode is a local decision parameter estimation under 0-1 loss $\chr_{y_i\neq \theta}$.
\end{fact}

Specifically, for binary classification with $\pm 1$ encoding, i.e., $\mathcal{Y}=\{1,-1\}$, we have
\begin{equation}\label{local-mode-reg-2}
\hat y(x_*):=\sign(\sum_{i}K(x_*,x_i)y_i).
\end{equation}
We call $y \sim \sum_{i}K(x_*,x_i)y_i$ the \term{local margin model}.

\subsubsection{Local mode linear classifier}

Similar to \fref{feature-ave-span}, we can construct a classifier based on feature mapping.

\begin{definition}[Linear Local Mode/Classification]\label{ex:lin-loc-clf}
Let $K(x_*,x) = \phi(x_*) \cdot \psi(x)$. Construct a linear classifier based on $\phi(x_*)$,
\begin{equation}\label{feature-mode-span}
\delta(x_*) = \phi(x_*) \cdot (\bmPsi^{\mathrm{T}}\vect{y}).
\end{equation}
We call this \term{linear local mode/classification}. When $\phi(x_*)$ is a linear map, this is a conventional linear classifier, while \fref{feature-ave-span} is never linear.
\end{definition}

Specifically, when $\mathcal{Y}=\{1,-1\}$, we have a margin model:
\begin{equation*}
y \sim \phi(x_*) \cdot (\bmPsi^{\mathrm{T}}\vect{y}).
\end{equation*}

\subsubsection{Local Mode Clustering}

\begin{figure}[H]
  \centering
  \includegraphics[width=\midwidth]{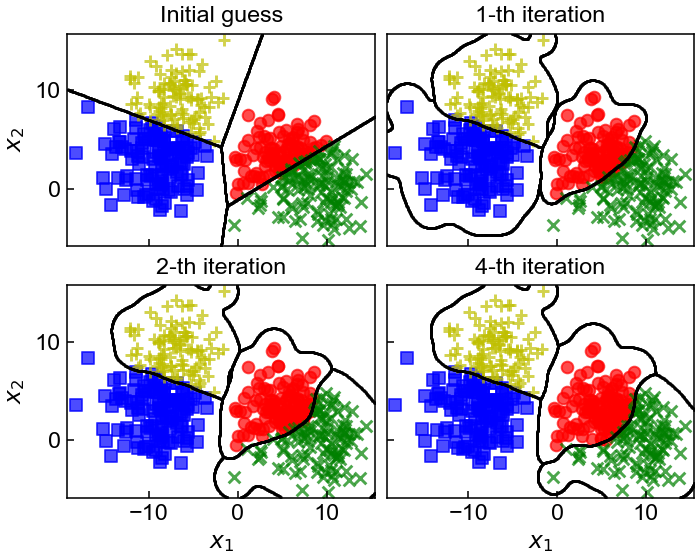}
  \caption{The result of local mode clustering, where the background color represents K-means clustering results for comparison (data generated by \texttt{scikit-learn})}
  \label{fig:local-clf-clu}
\end{figure}

We observe that the local mode classifier also provides a lazy transformation (see \autoref{sec:lazy-pred}):
\begin{equation}\label{local-mode-prob}
y_i\leftarrow\argmax_{y_i}\{\sum_{y_j=k}K(x_i,x_j)\} \quad\text{or}\quad p_i\leftarrow \sum_{j}K(x_i,x_j)p_j ,
\end{equation}
where $p_i$ is the label distribution of data point $x_i$, and the initial value can be set to the one-hot encoding of the K-means clustering result. This leads to a clustering method, which we call \term{local mode clustering}, with clustering effects shown in \asref{fig:local-clf-clu}. This is essentially the basic principle of \term{relaxation labeling} \cite{hummel1983}. Similar methods include \term{linear neighborhood propagation}\cite{wang2006}. \fref{local-mode-prob} shows both the hard and soft forms.

\begin{figure}[H]
  \centering
  \includegraphics[width=\tinywidth]{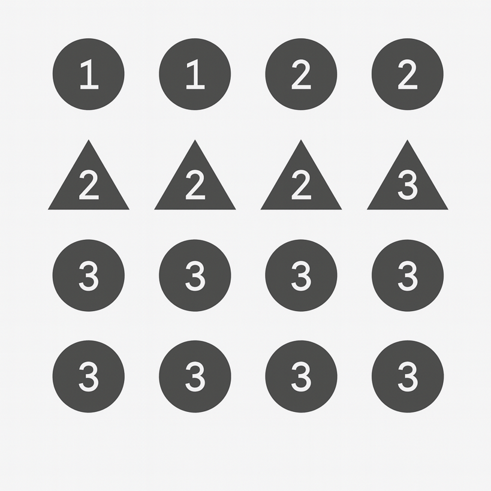}
  \caption{A canonical gestalt law example where the numbers are the labels of clustering}
  \label{fig:gestalt-labeled}
\end{figure}

The following hypothetical clustering in \autoref{fig:gestalt-labeled} (based on a certain self-localization kernel) is used to explain a typical Gestalt law \cite{valerjev2021}.

\subsection{K-Nearest Neighbors Model}

The \term{K-Nearest Neighbors (KNN) model} is a simple and old machine learning model \cite{wettschereck1997,kumar2017}. However, it performs well enough in many problems and is often used as a \term{baseline model} to evaluate the performance of more complex models.

\begin{definition}
  In the paper, the KNN model includes local regression/mean based on nearest neighbor decision,
\begin{equation}\label{knn-model}
  \hat y(x_*):=\frac{1}{K}\sum_{x_i\in \mathcal{N}_K(x_*,\matr{X})}y_i,
\end{equation}
and local classification/mode,
\begin{equation}\label{knn-clf}
  \delta_k(x_*):=|\{y_i=k,x_i\in \mathcal{N}_K(x_*,\matr{X})\}|=\sum_{x_i\in \mathcal{N}_K(x_*,\matr{X})}\chr_{y_i=k},
\end{equation}
where $\{(x_i,y_i)\}$ is the set of samples, and $\mathcal{N}_K(x_*,\matr{X})$ is the $K$-nearest neighbor set.
\end{definition}

For the KNN model, the most important parameter is not the bandwidth of the kernel but the number of nearest neighbors \cite{zhang2018knn}. Compared to neighborhood models (local models using neighborhood kernels), KNN models do not suffer from the lack of training samples near a target point. However, they can result in excessive connections between a target point and samples from different classes. We can combine nearest neighbor kernels with standard local kernels to leverage the advantages of both \cite{mitani2006,gou2011}, for example,
\begin{equation}\label{knn-loc-clf}
\hat y(x_*):=\frac{1}{K}\sum_{x_i\in \mathcal{N}_K(x_*,\matr{X})}K(x_*,x_i)y_i,
\end{equation}
where $K(\cdot)$ is the kernel ($K$ in $\mathcal{N}_K$ refers to the number of nearest neighbors).



\begin{remark}
The KNN model is an ancient non-parametric model. Related studies date back to \cite{fix1951,cover1967}.
\end{remark}

\section{Unsupervised Local Models}

\subsection{Kernel Density Estimation}\label{sec:kde}

Almost all machine learning, especially unsupervised learning, can be summarized as density estimation. While parameter estimation is the most representative form of estimation, we can directly perform density estimation without assuming a specific parametric form. \term{Kernel Density Estimation (KDE)} is one such method \cite{chen2017,gramacki2018}, and is considered as an unsupervised learning technique.

\begin{definition}[Kernel Density Estimation]
Let $K$ be a reasonable normalized (convolution) kernel. Then KDE of the distribution $p$ under kernel $K$ is
\begin{align}\label{kde}
\hat{p}_K(\vect{x}_*) & := \frac{1}{N} \sum_i K_h(\vect{x}_* - \vect{x}_i) = (K_h * \delta_{\matr{X}})(\vect{x}_*) \nonumber\\
& \approx (K_h * p)(\vect{x}_*) \to p(\vect{x}_*), \quad h \to 0,
\end{align}
where $\matr{X} = \{\vect{x}_i, i = 1, \dots, N\}$ are the samples, $\delta_{\matr{X}}$ is the empirical distribution, and the remaining symbols have the same meaning as in \autoref{th:id-approx}.
\end{definition}



\begin{remark}
  Although \fref{kde} and \fref{conv-reg} have similar forms, the normalization conditions are different in the non-convolution case: the former is normalized relative to $\vect{x}_*$, and the latter relative to $\vect{x}_i$.
\end{remark}

\begin{fact}\label{ft:kde-marginal}
  If $K(\vect{x}_*, \vect{x})$ is considered a conditional distribution $p(\vect{x}_*|\vect{x})$, then the KDE of $p$, $\hat{p}$, is exactly the marginal distribution corresponding to the empirical distribution $\delta_X(\vect{x})$ and the conditional distribution $p(\vect{x}_*|\vect{x}) \sim K(\vect{x}_*, \vect{x})$.
\end{fact}

Similarly, the conditional KDE \cite{de2003} (which includes the joint KDE) is defined as:
$$
\hat{p}_K(y|x) := \frac{\hat{p}_K(x,y)}{\hat{p}_K(x)} = \frac{\sum_i K_1(x,x_i) K_2(y,y_i)}{\sum_i K_1(x,x_i)}
$$
This corresponds to smoothing the local empirical distribution $\sum_i K_1(x,x_i)\delta_{y_i}$.

KDE can be understood as the orthogonal decomposition of the density function. Let the two function families $\{\phi_j(x)\}, \{\psi_j(x)\}$ be \term{bi-orthogonal bases}, meaning that $f(x) = \sum_j \inprod{f}{\psi_j} \phi_j(x)$. If we decompose the density function using these bases, we have
\begin{align*}
  p(x) & = \sum_j \inprod{p}{\psi_j} \phi_j(x) \\
  & = \sum_j (\Exp_x \psi_j(x)) \phi_j(x),
\end{align*}
which gives the \term{bi-orthogonal basis density estimate} \cite{principe2010}:
\begin{equation*}
  \hat{p}_{\phi,\psi}(x) := \sum_j \sum_i \phi_j(x) \psi_j(x_i).
\end{equation*}
In particular, when $\phi_j = \psi_j$, this is called the \term{orthogonal basis density estimate}.

It is easy to see that the bi-orthogonal basis density estimate is equivalent to KDE, where the kernel $K(\cdot)$ is determined by the feature mappings $x \mapsto \{\phi_j(x)\}$ and $x \mapsto \{\psi_j(x)\}$.

\subsection{Self-Local Mean}\label{sec:self-local-mean}

The initial goal of this section was to implement clustering using localization methods, but ultimately we focus on the core concept of the paper: \term{Self-Local Mean}.

\subsubsection{Definition of self-local mean}

Assume the sample space $\mathcal{X} \subset \R^p$ (closed under at least linear operations or convex combinations).

\begin{figure}[h]
  \centering
  \begin{subfigure}[t]{0.49\textwidth}
    \centering
    \includegraphics[width=\textwidth]{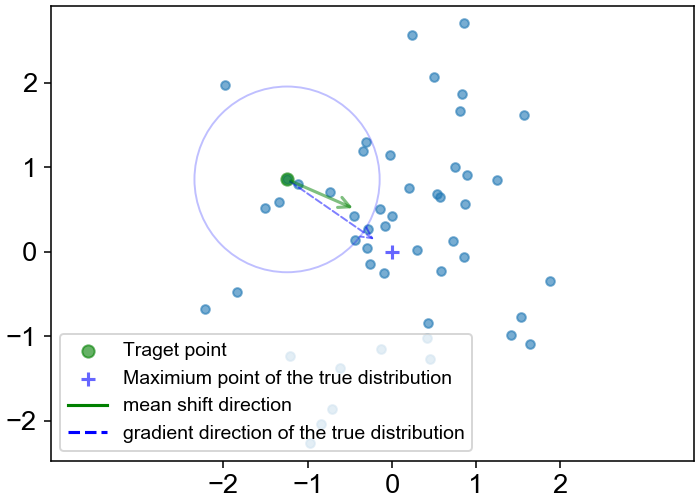}
  \caption{Mean shift of a point}
  \label{fig:meanshift}
  \end{subfigure}
  \hfill
  \begin{subfigure}[t]{0.49\textwidth}
    \centering
    \includegraphics[width=\textwidth]{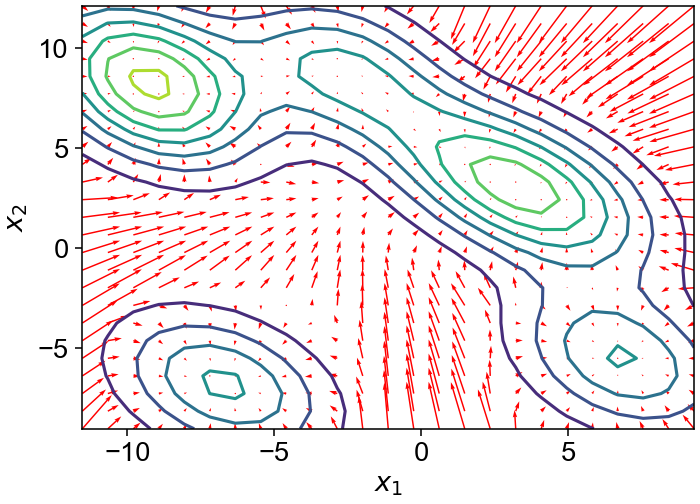}
  \caption{Mean shift vector field}
  \label{fig:meanshift-field}
  \end{subfigure}
  \caption{Illustration of mean shift}\label{fig:meanshift-field-demo}
\end{figure}

\begin{definition}[Self-Local Mean]\label{df:self-local-ave}
  Given a sample $\matr{X} = \{\vect{x}_i, i = 1, \dots, N\}$, the self-local mean is defined as
  \begin{equation}\label{self-local-ave}
  \slm_K(\vect{x}_*; \matr{X}) := \sum_i K(\vect{x}_*, \vect{x}_i) \vect{x}_i / \sum_i K(\vect{x}_*, \vect{x}_i),
  \end{equation}
  which is the local mean of the sample set $\{(\vect{x}_i, \vect{x}_i), \vect{x}_i \in \mathcal{X}\}$. We also call the mapping $\slm_K(\cdot; \matr{X}): \mathcal{X} \to \mathcal{X}$ the self-local mean transformation of the sample set $\matr{X}$, and $\lap_K(\vect{x})$ is referred to as the \term{mean shift (vector field)} \cite{cheng1995,comaniciu2002,aliyari2016}, \asref{fig:meanshift}.
\end{definition}

\begin{remark}
It is evident that the local mean can be viewed as the self-local mean of the sample space $\mathcal{X} \times \mathcal{Y}$ \cite{aliyari2016}, where the kernel does not depend on $y$, and the self-local mean can be considered the local mean of $\mathcal{X} \to \mathcal{X}$. The two are actually mutually inclusive and can both be called local means. Moreover, the probabilistic graph of the local mean is the pairwise Markov network/random field (see \autoref{fig:local-mean-pgm}).
\end{remark}

The self-local mean is formally very similar to \term{kernel Principal Component Analysis (kPCA)}. However, we find a fundamental difference between self-local mean and kPCA: the former reconstructs data using a stochastic matrix, while the latter does so using a projection matrix. The difference between local regression and kernel regression is also similar.

Different from the expectation of KDE, the self-local mean is indeed the expectation of the local empirical distribution (\autoref{rm:local-emp-distr}), while the following conclusion reveals the fundamental connection between self-local mean and KDE.

\begin{fact}\label{fc:meanshift-kde}
  The mean shift $\lap_K(\vect{x})$ and the gradient of KDE $\nabla \hat{p}_L(\vect{x})$ are aligned, where the kernel $K$ of the local mean can be derived from the kernel $L$ (called the \term{shadow kernel} \cite{cheng1995}) of KDE $\hat{p}(\vect{x})$ \cite{bhattacharya1967,fukunaga1975,comaniciu2002,meng2022}:
  \begin{equation}\label{meanshift-kde}
  \lap_K(\vect{x}) \propto \frac{\nabla_{\vect{x}} \hat{p}_L(\vect{x})}{\hat{p}_K(\vect{x})} \approx \nabla_{\vect{x}} \ln p(\vect{x}).
  \end{equation}
\end{fact}

\begin{remark}\label{rm:transition}
  According to \autoref{rm:normalized}, the normalized kernel $\tilde{K}(\vect{x}_*, \vect{x}_i)$ can be interpreted as the transition probability from $\vect{x}_*$ to $\vect{x}_i$. The self-local mean $\tilde{\matr{K}} \matr{X}$ is the mean of the sample set $\matr{X}$ after pointwise transitions.
\end{remark}

\fref{meanshift-kde} says that mean shift $\lap_K(\vect{x})$ approximately aligns with the \term{(Stein) score function}\cite{hyvarinen2005,lyu2012}, $\nabla_{\vect{x}} \ln p(\vect{x})$.

\subsubsection{Local mean lazy autoencoder}

The mapping $\slm_K$ defined in \fref{self-local-ave} is a \term{lazy reconstruction}/\term{lazy autoencoder} (see \autoref{sec:lazy-autoencoder}), referred to as the \term{(self) local mean (lazy) autoencoder}. The total loss for the local mean transformation is
$$J(K) = \|\matr{X} - \tilde{\matr{K}} \matr{X}\|_F^2 = \|\tilde{\matr{L}} \matr{X}\|_F^2,$$
where the symbol meanings are as above.

There are two matrix forms of the self-local mean:
\begin{enumerate}
  \item
  \begin{equation}\label{self-local-ave-matrix-sample}
\slm_K(\matr{X}', \matr{X}) =\tilde{K}(\matr{X}', \matr{X}) \matr{X} : \mathcal{X}^{N'} \times \mathcal{X}^N \to \mathcal{X}^N,
\end{equation}
where $\tilde{K}(\matr{X}', \matr{X})$ is the row-normalization of $K(\matr{X}', \matr{X})$.
  \item
\begin{equation}\label{self-local-ave-matrix}
\slm_K(\matr{X}) = \tilde{\matr{K}} \matr{X} = \{\slm_K(\vect{x}_i; \matr{X})\}_i: \mathcal{X}^N \to \mathcal{X}^N,
\end{equation}
which assumes $\matr{X}' = \matr{X}, \tilde{\matr{K}} := \tilde{K}(\matr{X}, \matr{X})$.
\end{enumerate}

\begin{figure}[h]
  \centering
  \includegraphics[width=\tinywidth]{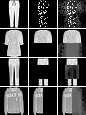}
  \caption{Missing data imputation results based on local mean autoencoder (fashion MNIST dataset)}
  \label{fig:missing-lm}
\end{figure}

\begin{experiment}\label{ex:missing-lm}
  We constructed two layers of local mean autoencoders: pixel-level and image-level, with pixel and image (spatial) as the sample spaces. The experiment steps are as follows. First, random values are used to initialize the filling of all images. Then, the image-level autoencoder fills in the data as any other encoder. Afterward, based on the previous filling, the pixel-level autoencoder performs another independent filling for each image. These two autoencoders alternate filling missing data through their respective reconstruction transformations. The final experimental results are shown in \autoref{fig:missing-lm}.
\end{experiment}

\subsubsection{MeanShift algorithm}

As a lazy reconstruction, $\slm_K$ maps the sample $\matr{X}=\{\vect{x}_i\}$ to $\matr{X}'=\{\slm_K(\vect{x}_i)\}$. The set $\matr{X}'$ should be more representative than $\matr{X}$. By repeatedly applying this, an iterative process is formed that converges to a fixed point approximating $\slm_K(\cdot)$. These fixed points can be considered as the cluster centers of the data set $\{\vect{x}_i\}$. This iterative process is the well-known \term{MeanShift algorithm}\cite{cheng1995,comaniciu2002,comaniciu2003,rao2009}, as shown in \autoref{algo:meanshift}.

\begin{algorithm}[h]
  \caption{MeanShift Algorithm}\label{algo:meanshift}
  \begin{algorithmic}[1]
  \Require Sample $\matr{X}=\{\vect{x}_i\}$
  \State Let the point set $\matr{X}^{(0)}=\matr{X}$;
  \For{starting from $t=1$}
    \State Compute the local mean using \fref{self-local-ave-matrix} (or \fref{self-local-ave-matrix-sample}) $\matr{X}^{(t)}\leftarrow \slm_K(\matr{X}^{(t-1)})$;
    \State Replace $\matr{X}^{(t-1)}$ with $\matr{X}^{(t)}$;
  \EndFor
  \end{algorithmic}
\end{algorithm}

\begin{remark}\label{rm:meanshift}
  In practice, using \fref{self-local-ave-matrix-sample}, where $\matr{X}$ is not overwritten by $\matr{X}^{(t)}$, can achieve the same effect. In fact, the term ``mean shift'' originally referred to this latter situation \cite{fukunaga1975,cheng1995,comaniciu1999}. We can even combine all sample point sets $\matr{X}^{(s)}, s<t$ together to compute $\matr{X}^{(t)}$.
\end{remark}

\begin{remark}\label{rm:modified-meansift}
  We can modify the iteration step of MeanShift as follows:
  \begin{equation}\label{alpha-meanshift}
  \matr{X} \leftarrow\matr{X} + \alpha (\tilde{\matr{K}}\matr{X} -\matr{X}) =  \alpha \tilde{\matr{K}}\matr{X} + (1-\alpha)\matr{X},0\leq\alpha\leq 1 ,
  \end{equation}
  by using the equivalent regularized kernel $\matr{L}=\alpha \tilde{\matr{K}} + (1-\alpha)$.
  which, in addition to controlling the convergence speed, makes the local mean transformation invertible. Moreover, following the idea of \autoref{rm:meanshift}, the iteration can be further modified to
  \begin{equation*}
  \matr{X}^{(t)} \leftarrow \sum_{s<t}\alpha_s \tilde{\matr{K}}^{(s)}\matr{X}^{(s)},
  \end{equation*}
  where $0\leq\alpha_s\leq 1$, and the other symbols have the same meaning as above.
\end{remark}

\begin{figure}[h]
  \centering
  \includegraphics[width=\midwidth]{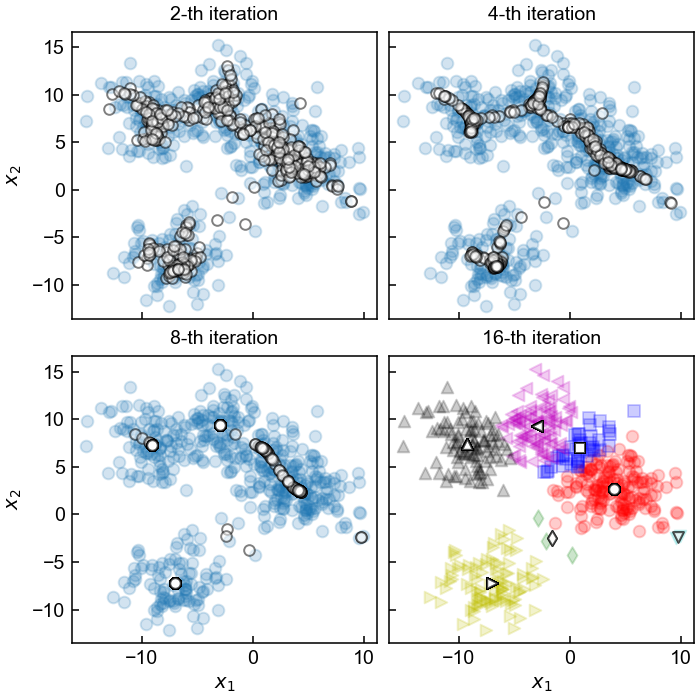}
  \caption{Clustering effect of the MeanShift algorithm (data generated by \texttt{scikit-learn})}
  \label{fig:meanshift-clustering}
\end{figure}

The iteration process in \autoref{algo:meanshift} causes all points to automatically converge to their neighboring regions, resulting in clustering. However, it does not directly provide cluster centers or class labels. Typically, \autoref{algo:meanshift} will separate points from different clusters and group similar points together. Eventually, points in the same cluster will merge into a single point, forming the cluster centers and assigning class labels to each center. \autoref{fig:meanshift-clustering} visually demonstrates the iterative process of the MeanShift algorithm.

\cite{comaniciu2002,li2007} proves the convergence of \autoref{algo:meanshift} under very weak conditions. \autoref{fc:meanshift-kde} shows that the iteration process of \autoref{algo:meanshift} is equivalent to a gradient ascent method that maximizes the value of the density function, and the regions where the points converge are the neighborhoods of local maxima of $\hat{p}(\vect{x})$.

\begin{remark}
  According to \autoref{rm:transition}, the MeanShift iteration process is equivalent to a Markov chain with a transition matrix $\tilde{\matr{K}}$, and the iteration result is the mean under the (multiple) stationary distributions of that Markov chain. In fact, we can fully simulate the Markov chain to implement the MeanShift algorithm, i.e., use Monte Carlo/resampling methods to compute the local mean. This is called the \term{Monte Carlo/Stochastic MeanShift Algorithm}.
\end{remark}

\begin{remark}\label{rm:meanshift-em}
The MeanShift algorithm can be viewed as a (variational) \term{Expectation-Maximization (EM) algorithm}\cite{ding2022}:
\begin{enumerate}
  \item E-step: Compute the local mean $\vect{x}_i^{(t+1)}\leftarrow \slm_K(\vect{x}_i^{(t)};\matr{X})$. This is equivalent to calculating the variational distribution $q(\vect{x}_i^{(t+1)}|\vect{x}_i^{(t)})\sim K(\vect{x}_i^{(t)},\vect{x}_i^{(t+1)})$.
  \item M-step: Each iteration implicitly involves density estimation $\hat{p}$ based on the sample $\{\vect{x}_i^{(t+1)}\}$.
\end{enumerate}
\end{remark}

\subsubsection{Image Segmentation and Edge Detection}

The MeanShift algorithm can also be used for \term{image segmentation} and \term{edge detection}. Unlike VQ, image segmentation requires identifying objects in an image, so the pixel position information must be taken into account, i.e., adopting a self-localization kernel defined on $(p_i, \vect{x}_i)$.

\begin{figure}[h]
  \centering
  \includegraphics[width=0.8\textwidth]{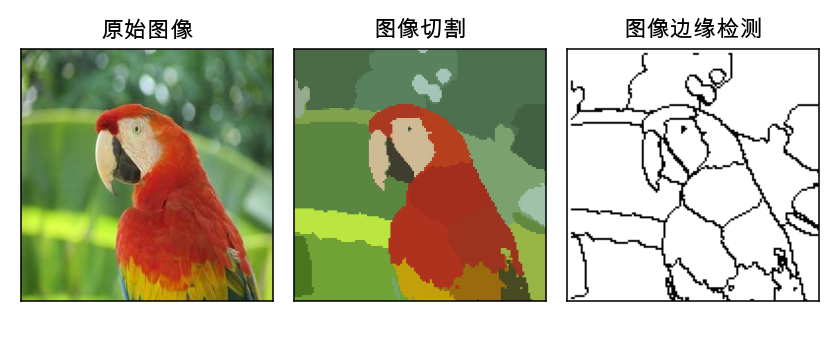}
  \caption{MeanShift image segmentation and edge detection results}
  \label{fig:meanshift-segment}
\end{figure}

By applying a classical edge detection operator to the results of the MeanShift algorithm, or simply comparing the pixel differences of neighboring points, we detect the image edges, \asref{fig:meanshift-segment}.

\begin{figure}[h]
  \centering
  \includegraphics[width=0.6\textwidth]{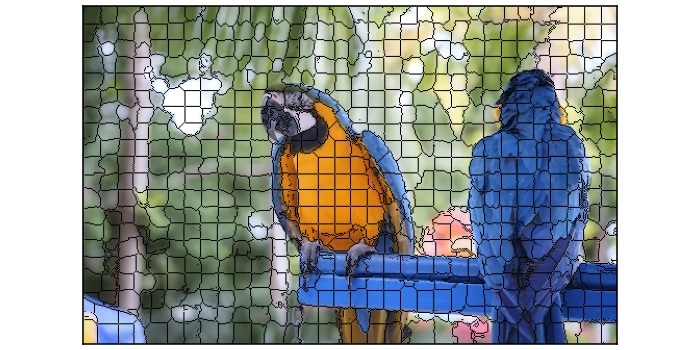}
  \caption{MeanShift superpixel segmentation results}
  \label{fig:meanshift-superpixel}
\end{figure}

Using the prototype method (where both the prototypes and the data points are updated) and setting reasonable parameters, the MeanShift algorithm can produce the effect of \term{superpixel segmentation}, \asref{fig:meanshift-superpixel}.

\subsubsection{Relation with self-organizing mapping}

The iteration process of MeanShift is very similar to that of the \term{Self-Organizing Mapping (SOM)}. In fact, we can consider MeanShift as a centerless form of SOM (see \autoref{sec:centerless-clf}). The difference is that SOM iterates over \term{prototypes}, while MeanShift iterates over the data. This difference is not insurmountable, as the local mean transformation can also act on points outside the sample (as the prototypes \cite{snell2017}), and SOM can select prototypes from the samples. Specifically, we select a set of prototypes $\mathcal{Q}\subset\mathcal{X}$ (or $\mathcal{Q}\subset\matr{X}$) from the sample space (or the samples), and then update $\mathcal{Q}$ according to the following formula:
$$
\vect{x} \leftarrow \slm_K(\vect{x}; \matr{X}), \vect{x}\in \mathcal{Q}.
$$

\begin{figure}[h]
  \centering
  \includegraphics[width=0.75\textwidth]{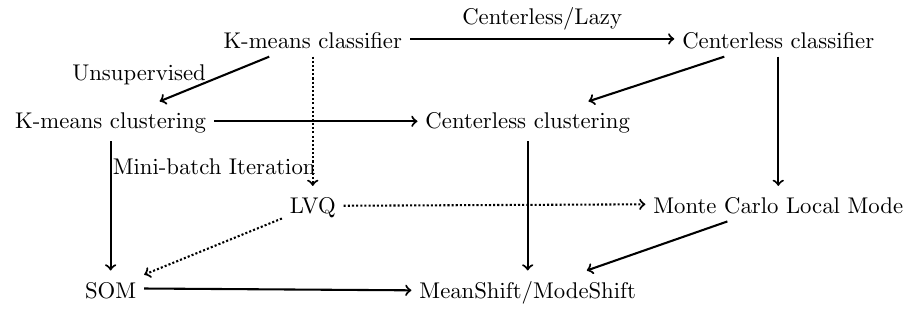}
  \caption{Model construction path based on the K-means center classifier (this diagram is not strict or absolute, but aims to provide heuristic reference for readers; LVQ is short for Linear Vector Quantization)}
  \label{fig:kmeans-meanshift-cube}
\end{figure}

Starting with the K-means center classifier, \autoref{fig:kmeans-meanshift-cube} outlines the general path for constructing related models.

\subsection{Self-Local Modes}\label{sec:self-loc-mode}

\subsubsection{Self-local modes}

Similarly, we define \term{self-local modes} for the discrete cases.

\begin{definition}[Self-local Modes]
The self-local mode of a sample $\{x_i\}$ is the local mode of $\{(x_i, x_i)\}$. The self-local mode transformation $M_k$ can be constructed in a similar manner (see the special form in \fref{lin-loc-mode}).
\end{definition}

The corresponding iterative algorithm can be called the \term{ModeShift algorithm}, which refers to the iterative process of the self-local mode transformation $M_k$. The self-local mode can, of course, also be converted into a self-local mean.

\begin{figure}[h]
  \centering
  \includegraphics[width=\midwidth]{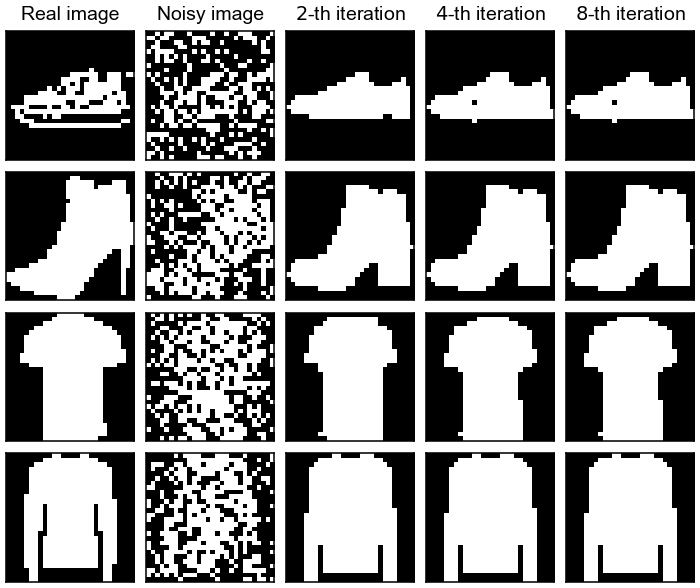}
  \caption{Self-local mode binary image reconstruction effect (only 5\% of binarized Fashion MNIST dataset as the sample $\matr{X}$)}
  \label{fig:modeshift}
\end{figure}

\begin{example}
  Let the sample space $\mathcal{X}=\{1,-1\}^p$, that is, binary encoding for discrete variables. The self-local mode can be defined as
\begin{equation}\label{lin-loc-mode}
  M_K(\vect{x}_*;\matr{X}) := \sign(\slm_K(\vect{x}_*;\matr{X})),
\end{equation}
where $\slm_K$ is the local mean. \fref{lin-loc-mode} can easily reconstruct binary images, \asref{fig:modeshift}.
\end{example}

\begin{example}[Nearest-Neighbor Shift/Label Propagation]
By using different kernels, different types of MeanShift/ModeShift algorithms can be obtained. If we choose a 1-nearest-neighbor kernel, we get the \term{Nearest-Neighbor-Shift algorithm}. The simple nearest-neighbor shift may not be very meaningful; it just moves each point to its closest neighbor. We provide a reasonable restriction in \autoref{algo:nnshift}. This algorithm is equivalent to traversing the data based on the neighborhood relations of data points, eventually forming several trees/forests with $\matr{X}^{(0)}$ as the root node. When applied to images, this is called the \term{Image-Forest Transformation (IFT)}\cite{vargas2019}.
\end{example}

\begin{algorithm}[h]
  \caption{Nearest-Neighbor Shift Algorithm}\label{algo:nnshift}
  \begin{algorithmic}[1]
  \Require Sample $\matr{X}=\{\vect{x}_i\}$
  \State Let the point set $\matr{X}^{(0)}\subset\matr{X}$;
  \For{starting from $t=1$}
    \State For all $x \in \samp{X}\setminus \matr{X}^{(t-1)}$, perform the following update:
    $$
    x' \leftarrow x,~ \text{if}~ x\neq x',~ \text{and $x'$ is the closest point to $x$ within the range $d(x', x) < \delta$}
    $$
    \Comment{No need to actually perform the assignment, just mark $x'$ with the same label as $x$}
    \State Let $\matr{X}'$ be the set of points $x'$ found in the current step that have neighbors satisfying the condition (before assignment). Note that not all $x'$ will have a corresponding $x$. Some may have multiple satisfying neighbors, in which case only one is selected.
    \State Update $\matr{X}^{(t)} \leftarrow \matr{X}^{(t-1)} \cup \matr{X}'$;
  \EndFor
  \end{algorithmic}
\end{algorithm}

Similar to \autoref{ex:lin-loc-clf}, if $K(\vect{x}_*,\vect{x})=\phi(\vect{x}_*) \cdot \psi(\vect{x})$, this is called the \term{linear (self-)local mode}.

\begin{example}[Hopfield Network]
Specifically, when $K(\vect{x}_*,\vect{x})=\vect{x}_* \cdot \vect{x}$, we obtain the classic \term{Hopfield (neural) network}\cite{hopfield1982,krotov2016,albanese2024}:
\begin{equation*}
  M_K(\vect{x}_*;\matr{X})=\sign(\vect{x}_*^{\mathrm{T}}\matr{G}), \matr{G}\propto \matr{X}^{\mathrm{T}}\matr{X}.
\end{equation*}
The Hopfield network algorithm is actually a special case of ModeShift/MeanShift. The energy function of the Hopfield network corresponds to the self-local mean loss. Formally, $\matr{G}$ is close to the covariance of $\matr{X}$, and it is the original form of the \term{Hebb learning rule}.
\end{example}

\subsubsection{Embedding methods}

We commonly use \term{embedding methods} to transform discrete data (such as text) into continuous data.

\begin{definition}[Embedding Methods]\label{df:loc-embedding}
  Let $v(\cdot): \mathcal{X} \to \R^q$ represent the \term{real embedding} of a point $\absv{x} \in \mathcal{X}$, and the corresponding empirical matrix is $\matr{V}=\{v(\absv{x}_i)\}: N \times q$. Define the local mean based on the embedding as
  \begin{equation}\label{loc-emb}
  \hat{v}(\absv{x}_i) = \sum_{j} K(\absv{x}_i,\absv{x}_j) v(\absv{x}_j) / \sum_j K(\absv{x}_i,\absv{x}_j)~
\text{or}~ \hat{\matr{V}} = \tilde{\matr{K}} \matr{V},
  \end{equation}
  where the kernel $K$ and $v(\absv{x}_i)$ are independently designed and do not need to be based on $v(\absv{x}_i)$.
\end{definition}


\begin{remark}
  In \autoref{df:loc-embedding}, $\mathcal{X}$ is typically a discrete space, but it can also be continuous. An embedding is just any mapping from $\mathcal{X}$ to $\R^q$ where $q$ is relatively small.
\end{remark}

We cannot directly apply the MeanShift algorithm to iterate over \fref{loc-emb}, because it only gives $\hat{v}(\absv{x}_i)$ and does not provide a method to update $\absv{x}_i$. In order to proceed with the next iteration, we need to decode the result to obtain the updated sample. Therefore, we introduce a decoder $w$, constructing an autoencoder:
\begin{equation}\label{emb-ae}
  (\hat{v}: \mathcal{X} \to \R^q, w: \R^q \to \mathcal{X}).
\end{equation}
Now, we can compute the reconstruction $\absv{x}_i' = w(\hat{v}(\absv{x}_i))$ at each iteration.

Assume $\mathcal{X} \subset \R^p$ (for discrete variables, a one-hot encoding of $|\mathcal{X}| = p$ can be performed first). We recommend using a linear decoder, $\hat{\matr{V}} \mapsto \hat{\matr{V}} \matr{W}^{\mathrm{T}}, \matr{W} \in \R^{q \times p}$. The iteration now becomes the following sample transformation:
\begin{equation}\label{loc-ave-ae}
\matr{X} \mapsto \hat{\matr{V}} \matr{W}^{\mathrm{T}} = \tilde{\matr{K}} \matr{V W}^{\mathrm{T}}: \R^{N \times p} \to \R^{N \times p} .
\end{equation}

\begin{definition}[Query-Key-Value Representation]\label{df:qkv}
In \fref{loc-emb}, $K$ and $v$ form a general quantitative representation of data, which can be called the \term{kernel-value representation}. Further, if we have $K(x, y) = \phi(x) \cdot \psi(y)$, where $\phi, \psi$ are feature mappings, then $\phi, \psi, v$ together form a complete real embedding that describes discrete variables, called the \term{query-key-value representation}\cite{vaswani2017, su2024}. Together with the decoder $w$, this representation forms a complete autoencoder as in \fref{emb-ae}, which can be referred to as the \term{query-key-value autoencoder}. (If the feature mapping is not explicitly given, then we jus call it as the \term{kernel-value representation/autoencoder}.)
\end{definition}

\section{Local Time Series Models}\label{sec:local-ts}

Time series on $\mathcal{X}$ can be transformed into a time-independent $\mathcal{T} \to \mathcal{X}$ supervised learning model. Thus, applying local methods to time series is entirely possible.

We assume the time series can be represented in the following form of a discrete dynamical system/autoregressive model:
\begin{equation}\label{dis-ds}
  x_{t+1} \sim f(x_t,t), x_t \in \mathcal{X},
\end{equation}
where $\mathcal{X} = \mathbb{R}^p$ (or a reasonable subset). It is easy to see that the localized form of \fref{dis-ds} is
\begin{equation}\label{local-ts-1}
  \hat{\vect{x}}_{t+1} = \sum_s K(\vect{x}_t,t,\vect{x}_s,s)\vect{x}_{s+1}/\sum_s K(\vect{x}_t,t,\vect{x}_s,s).
\end{equation}

\subsection{Local Sequence Models/Temporal (Self-)Local Mean}\label{sec:loc-ar}

Now, by slightly modifying \fref{local-ts-1}, we obtain the following form.

\begin{definition}[Local Sequence Model]\label{df:local-ts}
  A \term{local sequence model} can be represented as the following local mean, called the \term{temporal/sequential (self-)local mean}:
  \begin{equation}\label{local-ts}
  \hat{\vect{x}}_{t} = \slm_K(\vect{x}_t,t) := \sum_s K(\vect{x}_t,t,\vect{x}_s,s)\vect{x}_{s}/\sum_s K(\vect{x}_t,t,\vect{x}_s,s),
  \end{equation}
  where no restrictions is placed on the type of the time variable. In matrix form, this is $\hat{\matr{X}} = \tilde{\matr{K}} \matr{X}$, and the corresponding total loss is
  \begin{equation}
    J(K) = \|\matr{X} - \tilde{\matr{K}} \matr{X}\|^2_{F},
  \end{equation}
  where the stochastic matrix $\tilde{\matr{K}} = \{K(\vect{x}_t,t,\vect{x}_s,s)/\sum_s K(\vect{x}_t,t,\vect{x}_s,s)\}_{ts}$, and the design matrix $\matr{X} = \{\vect{x}_s\}$.
\end{definition}

\begin{remark}
Here, the kernel $K$ is defined on $\mathcal{X} \times \mathcal{T}$, where $\mathcal{T}$ is the time set. In contrast to the ``\term{static kernel}'' defined on $\mathcal{X}$, it can be called a \term{temporal kernel} or \term{dynamic kernel} (on $\mathcal{X}$).
\end{remark}

\begin{remark}
  \fref{local-ts} can be written as a lazy transformation on the sequences:
\begin{equation}\label{local-ave-seq}
  \{\hat{\vect{x}}_{t}\} = \slm_K(\{\vect{x}_t\}): \mathcal{X}^* \to \mathcal{X}^*.
\end{equation}
\end{remark}

The temporal local mean is essentially the local mean for the $\mathcal{T} \to \mathcal{X}$ transformation based on the self-localization kernel; the temporal kernel is simply an alias for the self-localization kernel in sequence models. Moreover, it is not difficult to construct a temporal regression model from $\mathcal{X}^*$ to $\mathcal{Y}^*$, which is actually a self-local mean on $\mathcal{X}\times \mathcal{Y}\times \mathcal{T}$.

\begin{fact}\label{fc:local}
All local models are equivalent in some sense, and essentially they are all local means.
\end{fact}

From the form of \fref{local-ts}, we can see that the localization of the time series transforms the Markov-like linear dependency of the sample into ``graph dependency''. This implies that the local sequence model can detect potential dependencies between any two variables in the sequence. In this sense, we assert that \autoref{df:local-ts} provides a universal sequence model. In the context of deep learning, this universal model is named the \term{self-attention mechanism}\cite{vaswani2017}. When the kernel in \fref{local-ts} does not depend on the time parameters $t$ and $s$, it degenerates into the ordinary local mean \cite{lee2018}.

\begin{example}
A simple design for the temporal kernel is (refer to \fref{kernel-sep})
\begin{equation}\label{tsai-kernel}
  K(\vect{x}_t,t,\vect{x}_s,s) = K_1(\vect{x}_t,\vect{x}_s) K_2(t,s) ~\text{or}~ K_1(\vect{x}_t,\vect{x}_s) K_2(t-s) ,
\end{equation}
where the first term reflects the graph dependency of variables that are independent of time, essentially being a kernel function on the sample space, and the second term represents what is known as \term{position encoding}\cite{shaw2018,tsai2019}. $K(\vect{x}_t,t,\vect{x}_s,s)$ represents the dynamic dependency, while $K_1(\vect{x}_t,\vect{x}_s)$ represents the static dependency, i.e., the static kernel. A more specific form is to let $K_2(t-s) = \chr_{|t-s| < \delta},s,t\in \mathbb{Z}$. We have
  \begin{equation*}
  \hat{\vect{x}}_{t} = \sum_{|t-s| < \delta} K(\vect{x}_t,\vect{x}_s)\vect{x}_{s}/\sum_{|t-s| < \delta} K(\vect{x}_t,\vect{x}_s) .
  \end{equation*}
  In this case, the temporal local mean is essentially the classic \term{weighted moving average}.
\end{example}

\begin{example}[Classic Self-Attention Mechanism]\label{rm:self-att}
  The classic self-attention mechanism uses the kernel,
  \begin{equation*}
    K(\vect{x}_t + p(t), \vect{x}_s + p(s)),
  \end{equation*}
  where $p(\cdot): \mathcal{T} \to \mathcal{X}$ is the so-called \term{absolute position encoding}. The famous language model, \term{Transformer}, makes full use of this mechanism \cite{vaswani2017}.
\end{example}

\begin{example}[Relative Position Encoding Self-Attention Mechanism]
  \cite{dai2019} provides the temporal kernel,
  \begin{equation*}
  K(\vect{x}_t,t,\vect{x}_s,s) = K_1(\vect{x}_t,\vect{x}_s) K_2(\vect{x}_t,t,s) ~\text{or}~ K_1(\vect{x}_t,\vect{x}_s) K_2(\vect{x}_t,t-s)
  \end{equation*}
  where $K_2(\vect{x}_t,t,s)$ or $K_2(\vect{x}_t,t-s)$ is the so-called \term{relative position encoding} (compare with \fref{tsai-kernel}).
\end{example}

\begin{example}
Let $K(\vect{x}_t,t,\vect{x}_s,s) = 0$ when $s > t$. Then we have
$$
\hat{\vect{x}}_{t} = \sum_{s \leq t} K(\vect{x}_t,t,\vect{x}_s,s)\vect{x}_{s}/\sum_{s \leq t} K(\vect{x}_t,t,\vect{x}_s,s).
$$
This formula implements a general autoregressive model, which is also a component of the trendy GPT model \cite{radford2018}.
\end{example}

\begin{figure}[h]
  \centering
  \includegraphics[width=0.9\textwidth]{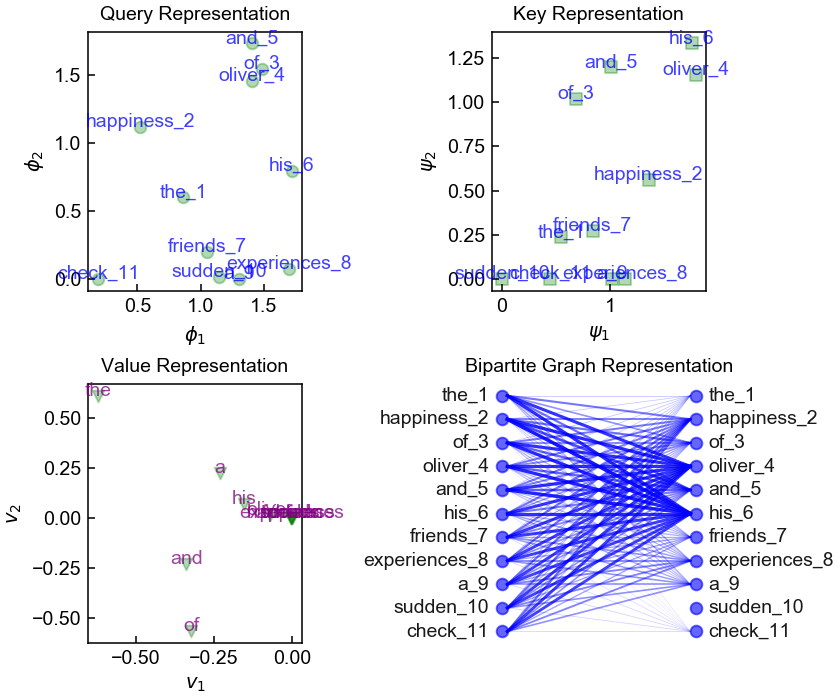}
  \caption{Demonstration of temporal query-key-value representations of words, where the value is fixed and position-independent, obtained through word embeddings(The darker the color of the edge and the thicker the line, the larger the corresponding element of the kernel matrix).}
  \label{fig:qkv}
\end{figure}

\begin{remark}
  By plotting the $(T, T)$ bipartite graph (where $T$ is the sequence length), we can visualize the temporal kernel \cite{vaswani2017}: vertices $\{\vect{x}_t\}$ are plotted on both sides, and the edge $(\vect{x}_t,\vect{x}_s)$ has a weight of $K_{t,s} = K(\vect{x}_t,t,\vect{x}_s,s)$, \asref{fig:qkv}.
\end{remark}

\begin{remark}
  Under normal circumstances, the self-attention mechanism/temporal local mean with discrete kernel only processes sequences of data points with a fixed length, because the kernel matrix requires a fixed size. To handle sequences of arbitrary lengths, the structure of \term{Recurrent Neural Network (RNN)} can be incorporated. Of course, if there is an explicit kernel function, then there is no such concern.
\end{remark}

\subsection{Temporal Self-Local Mode}

Similarly, we can define the \term{temporal (self-)local mode} to handle discrete value sequence problems.

\begin{definition}[Temporal Self-Local Mode]\label{df:loc-embedding-ts}
  A discrete value local sequence model can be represented as the following local mode:
  \begin{equation*}
  \hat{x}_{t} = \argmax_x \sum_{x_s = x} K(x_t,t,x_s,s).
  \end{equation*}
  We can also use embedding methods. Specifically, introducing a real embedding $v: \mathcal{X} \to \mathbb{R}^d$, a discrete value local sequence model can be represented as the following local mean:
  \begin{equation}\label{loc-embedding-ts}
  \hat{v}(x_t) = \sum_{s} K(x_t,t,x_s,s)v(x_s)/\sum_{s} K(x_t,t,x_s,s),
  \end{equation}
  where the kernel $K$ and $v(x_t)$ are generally designed independently (we do not recommend defining $v$ in relation to $t$).
\end{definition}

If we decompose the kernel $K$ in \fref{loc-embedding-ts} as the inner product of feature mappings, $K(x_t,t,x_s,s) = \phi(x_t,t) \cdot \psi(x_s,s)$, then $\phi, \psi, v$ together form a \term{temporal query-key-value representation}. If the kernel takes the form in \fref{local-feature-exp}, we construct the \term{query-key-value model} proposed in \cite{vaswani2017} for language models.

\subsection{Temporal Prediction}

A unique task of the time-series model is temporal prediction, that is, predicting $x_t$, and even $x_{t + 1},x_{t+2},\cdots$ based on $x_{s}, s = 1,\cdots, t - 1$. As an autoencoder, the temporal local model can accomplish such a task as a constrained reconstruction task (see \autoref{rm:supervised-ae}). There may be many technical details to consider, but we will not discuss them here.

\section{Local Model Examples}

Similar to kernel methods, localization methods can turn simple models into very powerful ones \cite{fan1996,markus2000,cheng2009,zhang2009,fan2011,meier2014,sugiyama2015}. Let us take a look at several specific cases.


\subsection{Local Linear Regression}

Local linear regression, with its simple form and the ability to approximate any machine learning model $y \sim f(\vect{x}), \vect{x} \in \mathbb{R}^n$, has become one of the most widely studied topics \cite{cleveland1979, cleveland1988}. We now derive the prediction formula for local linear regression.

Returning to \autoref{ex:local-linear-regression}, solving for \fref{local-linear-est} gives
\begin{gather}\label{local-linear-regression}
  \hat{y}(\vect{x}_*) = \hat{\beta}(\vect{x}_*) \cdot \vect{y} = L(\vect{x}_*, \matr{X}) \vect{y} , \\
  L(\vect{x}_*, \matr{X}) := \vect{x}_*^{\mathrm{T}} (\matr{X}^{\mathrm{T}} D(\vect{x}_*) \matr{X})^{-1} \matr{X}^{\mathrm{T}} D(\vect{x}_*), D(\vect{x}_*) := \diag\{ K(\vect{x}_*, \vect{x}_i) \} , \nonumber
\end{gather}
where $\{ (\vect{x}_i, y_i) \}$ is the sample. This confirms \autoref{fc:local}: local linear regression is equivalent to local means (based on empirical kernels).

We can also obtain the corresponding \term{local Ridge regression}:
\begin{gather*}
  \hat{y}(\vect{x}_*) = L_\lambda(\vect{x}_*) \cdot \vect{y}, \\
  L(\vect{x}_*) := \vect{x}_*^{\mathrm{T}} (\matr{X}^{\mathrm{T}} D(\vect{x}_*) \matr{X} + \lambda)^{-1} \matr{X}^{\mathrm{T}} D(\vect{x}_*), \lambda > 0. \nonumber
\end{gather*}

\begin{remark}
  To ensure that $L$ is a valid kernel, the input $\vect{x}$ should include an intercept term, i.e., $\matr{X}$ includes the column vector $\one_N$. Otherwise, if $\vect{x}_* = 0$, then $L(\vect{x}_*, \matr{X}) = 0$. Once the intercept term is included, $L(\vect{x}_*, \matr{X})$ is automatically normalized.
\end{remark}

\subsection{Local Classifiers}\label{sec:local-clf}

Classifiers can roughly be divided into two major types:
\begin{enumerate}
  \item Linear classifiers: The decision boundary for classification is a hyperplane, such as \term{logistic regression} and \term{Support Vector Machine (SVM)}.
  \item Center classifiers: The classification result is determined by the distance from the point to the center, such as \term{Linear Discriminant Analysis (LDA)} and \term{Quadratic Discriminant Analysis (QDA)}. See \autoref{df:center-classifier}.
\end{enumerate}

Since a local linear classifier can clearly be transformed into a local linear regression, here we will only construct the \term{local center classifiers}, as the results of applying the localization trick to center classifiers.

The center $\mu$ of a set $\{x_i\}$ can be defined as:
\begin{equation*}
\hat{\mu} = \argmin_{\mu} \sum_{i} d(x_i, \mu),
\end{equation*}
where $d$ is a distance. Then the \term{local center} (relative to $x_*$) is the solution to the following optimization problem:
\begin{equation}\label{local-center}
  \min_{\mu} \sum_i K(x_*, x_i) d(x_i, \mu),
\end{equation}
The local center is a generalization of the local mean. When $d$ is the Euclidean distance, the local center is the local mean, i.e., \fref{local-ave}.

\begin{definition}
  A local center classifier is determined by the following decision function:
  \begin{equation}\label{local-center-clf}
  \delta_k(x_*) = -d(x_*, \hat{\mu}_k(x_*)),
  \end{equation}
  where $\hat{\mu}_k(x_*)$ is the (relative to $x_*$) local center of the $k$-th class data $\{x_i: k\}$.
\end{definition}

A typical example of a local center classifier is the \term{local LDA}\cite{fan2011, sugiyama2015}. To avoid the time-consuming calculation of the local centers, we consider the centerless form (refer to \autoref{sec:centerless-clf}).

\begin{definition}\label{df:local-centerless-clf}
  A local centerless classifier is determined by the following decision function:
  \begin{align}\label{local-centerless}
  \delta_k(x_*) = & -\sum_{x_i: k} K(x_*, x_i) d(x_*, x_i) / \sum_{x_i : k} K(x_*, x_i) \nonumber \\
  & [+ \sum_{x_i, x_j : k} K(x_*, x_i) K(x_*, x_j) d(x_i, x_j) / (\sum_{x_i : k} K(x_*, x_i))^2],
  \end{align}
  where the regularization term in $[\cdot]$ could be ignored (refer to \autoref{df:centerless-clf}).
\end{definition}

\fref{local-centerless} derives a lazy transformation:
  \begin{equation}\label{local-centerless-lazy}
  \matr{R} \mapsto \softmax\left(- \left( (\matr{K} \circ \matr{D}) \matr{R} \right) \oslash \matr{KR} \right),
  \end{equation}
where $\matr{K}$ is the kernel matrix, $\matr{D} = \{ d(x_i, x_j) \}$ is the distance matrix, and $\matr{R}$ is the (probability) response matrix.

\subsection{Local Center Clustering}

Assume that within a small neighborhood, there is only one clustering center. Therefore, when constructing a local model for clustering, we only need to consider a single clustering point.

\begin{definition}[Local Center Clustering/Center Shift]\label{df:local-center-clu}
\term{Local center clustering} is essentially fully determined by the local center of the entire dataset:
\begin{equation}
  \mu^* = \argmin_{\mu} \sum_i K(x_*, x_i) d(x_i, \mu), x_*, x_i\in\R^p,
\end{equation}
We refer to the mapping $x_* \mapsto \mu^*$ given by \fref{local-center} as \term{center shift}.
\end{definition}

As a special case of local center clustering, \term{local K-means clustering} is determined by the solution to the following optimization problem:
\begin{equation}\label{local-kmeans}
\min_{\vect{\mu}} \sum_i K(\vect{x}_*, \vect{x}_i) \|\vect{x}_i - \vect{\mu}\|^2_2.
\end{equation}
The solution to \fref{local-kmeans} is the local mean, so the local mean can be viewed as a localization of K-means clustering. This also verifies \autoref{fc:local}.


\subsection{Local Centerless Clustering}

When the local center $\mu$ is restricted to the sample points, this becomes \term{local K-medoids clustering}.

According to the definition of medoids, \fref{local-center} can be expressed in the following centerless form:
\begin{gather}\label{local-medoid}
x_i \mapsto x_{j^*}, \quad j^* = \argmin_j (\sum_k\matr{K}_{ik} \matr{D}_{kj}/\sum_k\matr{K}_{ik}),  \nonumber\\
 \text{or}\quad x_i \mapsto x_{j^*}, \quad j^* = \argmin_j (\tilde{\matr{K}} \matr{D})_{ij},
\end{gather}
where $\tilde{\matr{K}}$ is the normalized kernel matrix, and $\matr{D} = \{ d(x_i, x_j) \}$ is the distance matrix.

\begin{remark}
  The local centerless model ingeniously combines the kernel (matrix) and the distance (matrix). Since the distance can also serve as a kernel, the local centerless model is actually a hierarchical local model (see \autoref{sec:hie-loc}).
\end{remark}

\begin{figure}[h]
  \centering
  \includegraphics[width=\midwidth]{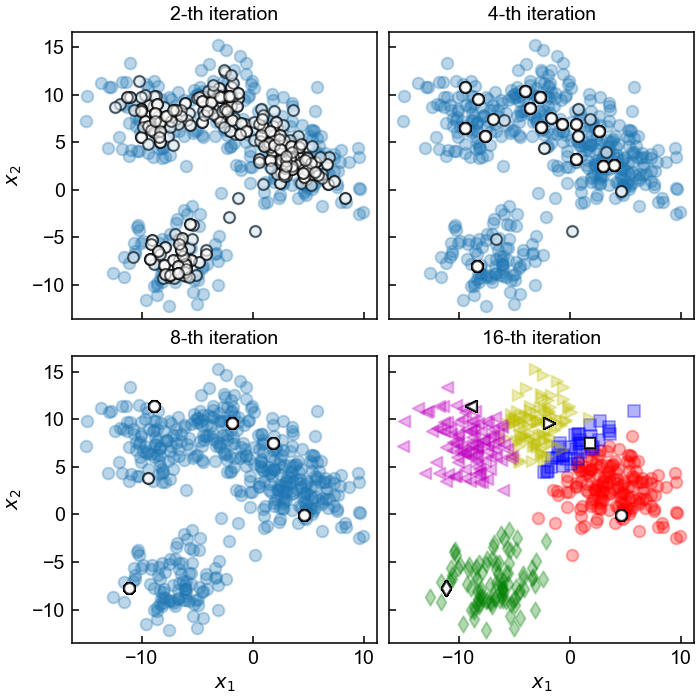}
  \caption{Clustering effect of the MedoidShift algorithm (data is the same as that in \autoref{fig:meanshift-clustering} and no post-processing is required in this case)}
  \label{fig:medoidshift-clustering}
\end{figure}

\fref{local-medoid} provides a lazy transformation of $\matr{X} \mapsto \matr{X}$, and combines the kernel matrix with the distance matrix. Therefore, we can construct the \term{MedoidShift algorithm}\cite{sheikh2007}. Unlike MeanShift, MedoidShift can be directly applied to clustering. However, to fix the number of clusters, some necessary post-processing is still required. See \autoref{fig:medoidshift-clustering} for the demonstration of the MedoidShift clustering algorithm.

\begin{remark}
Please note the difference between \fref{local-medoid} and \fref{local-centerless-lazy}, as well as the clustering models derived from both.
\end{remark}

\begin{figure}[H]
  \centering
  \includegraphics[width=\midwidth]{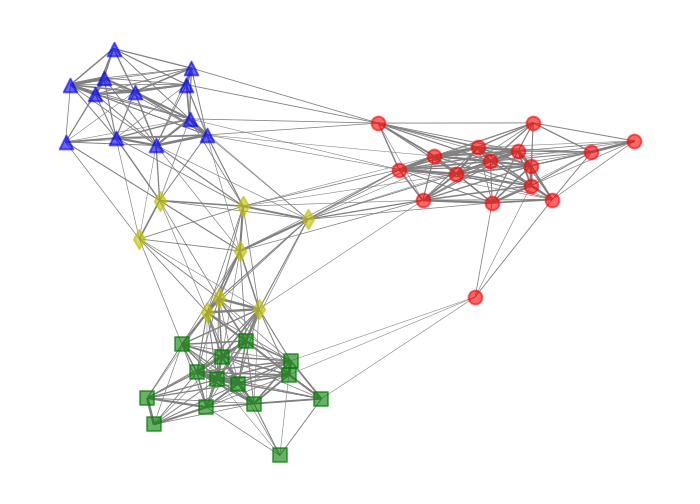}
   \caption{Network analysis by MedoidShift (not substantially depend on the coordinates of nodes, but rather on the pointwise distances.)}
  \label{fig:medoidshift-network}
\end{figure}

The centerlsss form or MedoidShift is not dependent on the Euclidean distance and thus can be applied to network analysis, \asref{fig:medoidshift-network}.

\subsection{Local PCA}\label{sec:local-pca}

Peaple solved the latent variable model $P(x, z)$ by optimizing the \term{variational lower bound(ELBO)}. Therefore, one strategy for constructing local latent variable models is to establish the concept of ``local ELBO''. However, we recognize that within a local neighborhood, the latent variable model should also be linear, like PCA. Therefore, we directly construct \term{local PCA} \cite{kambhatla1997, weingessel2000}.

\begin{definition}[Local PCA]
  Local PCA is determined by the solution to the following optimization problem:
  \begin{equation}\label{local-pca}
    \min_{\matr{V} \in O_{p \times r}} \sum_i K(\vect{x}_*, \vect{x}_i) \|\vect{x}_i - \matr{V} \matr{V}^{\mathrm{T}} \vect{x}_i \|_2^2,
  \end{equation}
  where $\{\vect{x}_i\}$ are the samples (in \fref{local-pca}, the constant term should not be omitted; readers should add it themselves). In particular, when $K(\vect{x}_*, \vect{x})$ is a neighborhood kernel (\autoref{df:neighbor-kernel}), it is referred to as \term{neighborhood PCA}. A similar definition applies to \term{nearest-neighbor PCA}.
\end{definition}

Solving \fref{local-pca} is equivalent to performing weighted PCA/SVD on the design matrix $\matr{X}$, where the weights are $\matr{W}_{ij} = K(\vect{x}_*, \vect{x}_i)$. The corresponding solution is denoted as $\matr{V}_{\vect{x}_*}$.

\begin{figure}[H]
  \centering
  \includegraphics[width=0.55\textwidth]{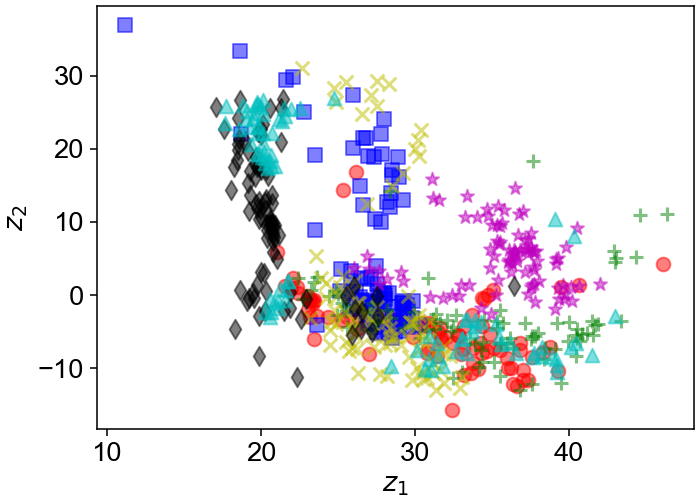}
  \caption{Dimensionality reduction effect of Local PCA}
  \label{fig:lpca}
\end{figure}

Local PCA gives the following nonlinear encoder:
\begin{equation}\label{lpca-encode}
  \phi(\vect{x}) = \matr{V}_{\vect{x}}^{\mathrm{T}} (\vect{x} - \vect{\mu}_{\vect{x}}) + \matr{V}^{\mathrm{T}} (\vect{\mu}_{\vect{x}} - \vect{\mu}), \quad \vect{x} \in \R^p,
\end{equation}
where $\vect{\mu}_{\vect{x}}, \matr{V}_{\vect{x}}$ are produced by local PCA, while $\vect{\mu}, \matr{V}$ are produced by global PCA. If the global part is not included in \fref{lpca-encode}, then $\phi(\vect{x})$ will be concentrated near the origin and fail to reflect the true position of $\vect{x}$ in the sample space. The dimensionality reduction effect of local PCA is shown in \autoref{fig:lpca}.

However, when constructing autoencoders, we encounter difficulties similar to those in kPCA, namely the absence of an explicit decoder. One approach is to solve the nonlinear equation $\phi(\vect{x}) \approx \vect{z}$ and compute $\vect{x}^*$ as the decoding of $\vect{z}$. Another approach is \term{inverse mapping learning}\cite{mika1998}.


\begin{definition}[Local Reconstruction]
  Local PCA has a reconstruction transformation for each target point $\vect{x}$, given by:
  \begin{equation}\label{pc-shift}
  \vect{x} \mapsto \matr{V}_{\vect{x}} \matr{V}_{\vect{x}}^{\mathrm{T}} (\vect{x} - \vect{\mu}_{\vect{x}}) + \vect{\mu}_{\vect{x}} = (1 - \matr{V}_{\vect{x}} \matr{V}_{\vect{x}}^{\mathrm{T}}) \vect{\mu}_{\vect{x}} + \matr{V}_{\vect{x}} \matr{V}_{\vect{x}}^{\mathrm{T}} \vect{x}.
  \end{equation}
  We refer to the transformation as the \term{local reconstruction} based on PCA.
\end{definition}

Formally, the local reconstruction presented in \fref{pc-shift} is a kind of local mean with a particular regularized kernel.

\begin{figure}[h]
  \centering
  \includegraphics[width=0.65\textwidth]{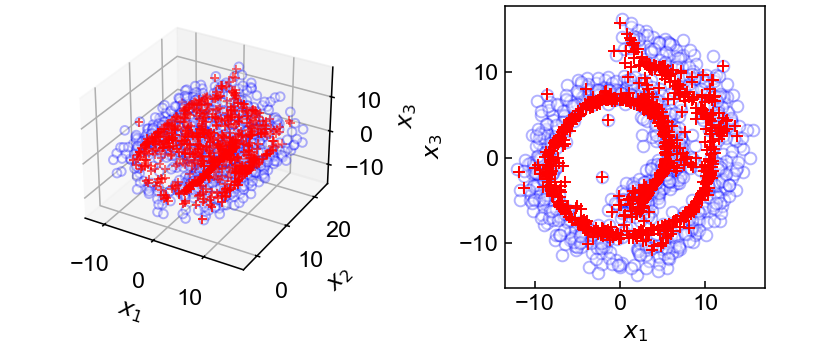}
  \caption{Local reconstruction effect based on nearest-neighbor PCA (the left image is a swiss roll dataset generated by \texttt{scikit-learn}, and the right image shows its projection onto the plane for easier observation by the reader)}
  \label{fig:local-reconstruct}
\end{figure}

As a kind of mean shift, $(1 - \matr{V}_{\vect{x}} \matr{V}_{\vect{x}}^{\mathrm{T}})(\vect{\mu}_{\vect{x}} - \vect{x})$ can be called the \term{principal component shift (PC-Shift)}. The corresponding iterative process/algorithm can be called the PC-Shift algorithm, and its effect is shown in \autoref{fig:local-reconstruct}. For similar idea see \cite{yao2016}.

Obviously, the regular local mean could be regarded as ``local constant encoder'', i.e. the solution to the following optimization problem:
\begin{equation}\label{local-constant-encoding}
  \min_{\vect{z}} \sum_i K(\vect{x}_*, \vect{x}_i) \|\vect{x}_i - \vect{z}\|_2^2.
\end{equation}
Therefore, we draw the following conclusions.

\begin{fact}
  Local dimensionality reduction (reconstruction), local clustering, and self-local mean are equivalent concepts in a certain sense.
\end{fact}

\subsubsection{Local Autoencoder}

Constructing a general local autoencoder is not difficult. We directly apply the localization trick to the reconstruction loss of a general autoencoder:
\begin{equation}\label{local-ae}
  \min_{\phi, \psi} \sum_i K(\vect{x}_*, \vect{x}_i) l(\vect{x}_i, \psi(\phi(\vect{x}_i))) .
\end{equation}

Corresponding concepts such as local reconstruction can be further defined. However, applying the localization method to complex models may have limited effect. More in-depth discussions are left to the reader.

\subsection{Kernel Density Classifier}

We introduce a simple application of KDE. Consider the generative classification model $p(x, y)$. We use KDE to estimate the generative distribution $p_k(x) = p(x \mid y = k)$ of each class:
\begin{equation}\label{gen-distr-kde}
  \hat{p}_k(x) \sim \sum_{x_i:k} K_h(x, x_i),
\end{equation}
where $\sum_{x_i:k}$ denotes the summation over the samples of class $y = k$. This expression is exactly the same as \fref{kde}. We then give the discriminant function,
\begin{equation}\label{kd-clf}
  \delta_k(x) = \ln \hat{p}_k(x) + \ln \hat{\pi}_k,
\end{equation}
where $\hat{\pi}_k$ is directly estimated by the empirical distribution of $y$. The classification model determined by \fref{kd-clf} is called the \term{kernel density (Bayes) classifier}. A typical example is the \term{kernel density naive Bayes classifier} \cite{hastie2009}.

\begin{fact}
  Kernel density classification is local mode/classification with a prior term.
\end{fact}

\section{Adaptive Localization Methods}\label{sec:adaptive-local}

Similar to kernel methods, localization methods also have adaptive forms, specifically minimizing $J(K)$ in \fref{local-kernel-loss} mentioned earlier.

\begin{definition}[Adaptive Localization Method]
We define the general form of the adaptive localization method as
\begin{equation}\label{min-local-kernel}
  \min_K J(K; \matr{X}) ,
\end{equation}
where $\matr{X}$ represents the samples. If feature mappings are introduced, \fref{min-local-kernel} can be rewritten as
\begin{equation}\label{min-local-feature}
  \min_{\phi, \psi} J(K(\phi, \psi); \matr{X}),
\end{equation}
where $K(\phi, \psi)$ is the kernel determined by the feature mappings $\phi, \psi$. The kernel obtained in this way is referred to as the \term{adaptive (localization) kernel}.
\end{definition}

The optimization problem \eqref{min-local-kernel} directly defines a statistical model on the sample space $\mathcal{X}^N$, and it cannot be reduced to an additive form (\autoref{df:additive-stat}). We also call this \term{(localization) kernel learning}. \fref{min-local-feature} can be referred to as \term{feature mapping learning}.

\begin{example}
  A common example is the adaptive self-local mean, which is given by the following optimization problem:
  \begin{align}\label{adaptive-self-local-ave}
   & \min_{K} \sum_i \left( \vect{x}_i - \sum_{j} K(\vect{x}_i, \vect{x}_j) \vect{x}_j / \sum_{j} K(\vect{x}_i, \vect{x}_j) \right)^2 \nonumber\\
  \iff & \min_{K} \|\matr{X} - \tilde{\matr{K}} \matr{X}\|_2^2,
  \end{align}
  where $\tilde{\matr{K}}$ is the normalized kernel matrix corresponding to $K$ (it is recommended to use its hollow form).
\end{example}

We introduce three forms of adaptive kernels: \term{parametric kernel}, \term{multi-kernel}, and \term{discrete kernel}.

\subsection{Parametric Kernel}\label{sec:param-local-kernel}

\begin{definition}[Parametric Kernel]
  Let $l(x, \theta)$ be a given loss function. Introduce a parameter $w$ to the kernel $K$ (as a model hyperparameter), denoted as $K(x, y; w)$, and define the parametric local loss as
  \begin{equation}
  J(x_*, \theta, w) := \sum_j K(x_*, x_j; w) l(x_j, \theta),
  \end{equation}
  where $w$ is tunable. The local parameter estimate corresponding to the sample point $x_i$ is denoted as $\hat{\theta}(x_i; w)$. The optimal parameter $w$ is the solution to the solution to the following optimization problem (as the parametric form of \fref{local-kernel-loss}):
  \begin{equation}\label{local-loss-param}
  \min_{w} J(w) := \sum_i l(x_i, \hat{\theta}(x_i; w)).
  \end{equation}
\end{definition}

\begin{remark}
  To avoid overfitting, $K$ is recommended to be in hollow form when calculating errors, or testing data $\{x_i'\}$ should be used, such as
    \begin{equation*}
  \min_{w} J(w) := \sum_i l(x_i', \hat{\theta}(x_i; w)).
  \end{equation*}
\end{remark}

\begin{example}
  For the local mean in \fref{local-ave}, the optimal kernel parameter $w$ is the solution to the solution to the following optimization problem:
  \begin{align}\label{adaptive-local-ave}
   & \min_{w} \sum_i \left( y_i - \sum_{j} K(x_i, x_j; w) y_j / \sum_{j} K(x_i, x_j; w) \right)^2 \nonumber\\
  \iff & \min_{w} \|\vect{y} - \tilde{\matr{K}}(w) \vect{y}\|_2^2.
  \end{align}
The most typical case is to optimize the bandwidth, i.e., $\tilde{\matr{K}}(w)$ is the normalized kernel matrix $\{ K\left( \frac{\vect{x}_j - \vect{x}_i}{w} \right) \}_{ij}, \vect{x}_i \in \R^p$.
\end{example}

\begin{figure}[h]
  \centering
  \begin{subfigure}[t]{0.49\textwidth}
    \centering
    \includegraphics[width=\textwidth]{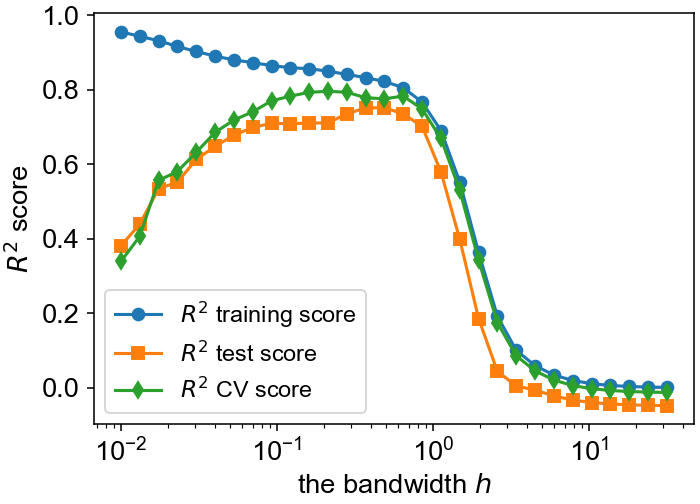}
    \caption{Relationship between the bandwidth $h$ of the local linear regression kernel and the $R^2$ score}
    \label{fig:local-regression-window}
  \end{subfigure}
  \hfill
  \begin{subfigure}[t]{0.49\textwidth}
    \centering
    \includegraphics[width=\textwidth]{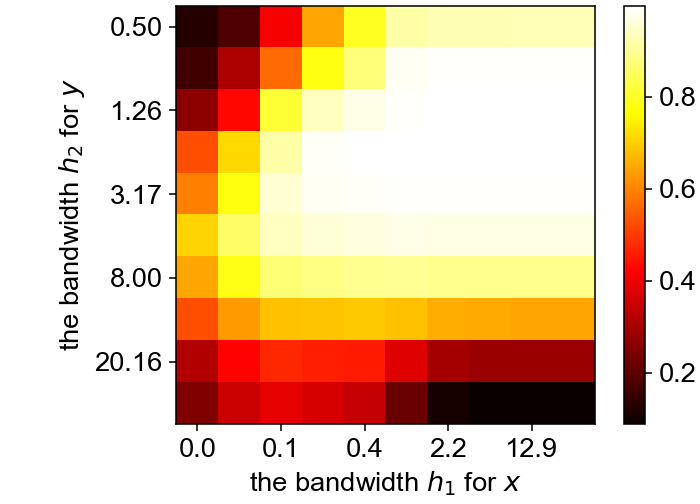}
    \caption{Heatmap of the bandwidth of the self-localization kernel input $x$ and output $y$ with $R^2$ scores}
    \label{fig:self-local-mean-window}
  \end{subfigure}
  \caption{Relationship between the bandwidth $h$ of the adaptive kernel and $R^2$ score}\label{fig:local-window}
\end{figure}

\autoref{fig:local-regression-window} plots the relationship between the bandwidth $h$ of the local linear regression kernel and the $R^2$ score of the model. In this example, the leave-one-out $R^2$ score accurately reflects model overfitting and provides an optimal bandwidth close to the test $R^2$ score. In fact, we used the optimized bandwidth (approximately 0.005) when plotting \autoref{fig:local-linear-regression-1d}. Image processing also refers to the results of \autoref{fig:self-local-mean-window}.

\subsection{Multi-Kernel}

\begin{definition}[Multi-Kernel]\label{df:multi-kernel}
  A multi-kernel can be understood as a special case of a parameterized kernel, defined as
  \begin{equation*}
K(\absv{x}_*,\absv{x};\{w_m\}):=\sum_m w_m K_m(\absv{x}_*,\absv{x}), w_m\geq 0,
  \end{equation*}
  where $K_m$ is a given kernel, and we can require that $\sum_mw_m=1$. The corresponding local loss is
  \begin{equation*}
J(\absv{x}_*,\theta,\{w_m\}):=\sum_j\sum_mw_mK_m(\absv{x}_*,\absv{x}_j)l(\absv{x}_j,\theta).
  \end{equation*}
\end{definition}

\begin{example}
  The Filtered KDE constructed by \cite{marchette1994} is a KDE based on multi-kernels.
\end{example}

A multi-kernel can be viewed as a simple combination of a local model and the original model (which can be called the \term{global model}):
\begin{align}\label{global-local-kernel}
J(\absv{x}_*,\theta,w) := & \sum_j (wK(\absv{x}_*,\absv{x}_j)+(1-w)\delta_{\absv{x}_*\absv{x}_j})l(\absv{x}_j,\theta) \nonumber\\
= & w\sum_j K(\absv{x}_*,\absv{x}_j)l(\absv{x}_j,\theta)+(1-w)l(\absv{x}_*,\theta),  0\leq w\leq 1 .
\end{align}

Since \fref{global-local-kernel} contains only two kernels, it can be referred to as a \term{bi-kernel}.
The kernel matrix corresponding to the bi-kernel $wK(\absv{x}_*,\absv{x})+v\delta_{\absv{x}_*\absv{x}}$ is
\begin{equation}\label{double-kernel}
w\matr{K} + v\eye, w, v\geq 0.
\end{equation}

The bi-kernel form \fref{double-kernel} has appeared in the kernel ridge regression \cite{saunders1998} and \fref{alpha-meanshift}, that is, regularized kernels.

\begin{remark}
  Another formulation of the ``local + global'' bi-kernel is $wK(\absv{x}_*,\absv{x})+v$, with the corresponding kernel matrix being $w\matr{K} + v\one_{N\times N}$.
\end{remark}

It is not difficult to derive the local mean for a multi-kernel:
\begin{equation*}
\hat{y}(\absv{x}_*) := \sum_{m,i}w_{m}K_m(\absv{x}_*,\absv{x}_i)\absv{y}_i/\sum_{m,i}w_{m}K_m(\absv{x}_*,\absv{x}_i), w_m\geq 0.
\end{equation*}

\begin{remark}
Reasonably assume that $\sum_{j}K_m(\absv{x}_i,\absv{x}_j)\approx d(x_i)$, and $\sum_{m}w_{m}=1$, then
\begin{equation}\label{mk-approx}
J(\{w_m\}) \approx \|\sum_{m}w_{m}\matr{L}_m\vect{y}\|_F^2 .
\end{equation}
where $\matr{D}=\mathrm{diag}(\{d(\absv{x}_i)\})$, and $\matr{L}_m$ is the normalized Laplacian of $K_m$. Minimizing \fref{mk-approx} will be a constrained quadratic programming problem.
\end{remark}

\subsection{Multi-Head}
The multi-kernel \cite{vaswani2017} concept is close to the multi-head mechanism in attention models, but not exactly the same. We provide a strict definition of the multi-head, based on the autoencoder defined in \fref{loc-ave-ae}. Regarding its variants (as a form of parameter binding), refer to \cite{shazeer2019,ainslie2023,meng2025}.

\begin{definition}[Multi-Head]\label{df:multihead}
  Let $(\hat{\matr{V}}_m,\matr{W}_m), m=1,\cdots, M$ (with the corresponding kernels denoted as $K_m$) be $M$ autoencoders of the form defined in \fref{loc-ave-ae}, then the multi-head model is a local mean of the following form,
  \begin{equation}\label{multihead}
  \hat{\matr{X}} = \frac{1}{M} \sum_m \hat{\matr{V}}_m\matr{W}_m^{\mathrm{T}}= \frac{1}{M} \begin{pmatrix}\hat{\matr{V}}_1& \cdots & \hat{\matr{V}}_M\end{pmatrix} \begin{pmatrix}\matr{W}_1& \cdots & \matr{W}_M\end{pmatrix}^{\mathrm{T}}.
  \end{equation}
\end{definition}

Formally, when $\matr{W}_m$ is a scalar, \fref{multihead} degenerates into the multi-kernel case. Therefore, we view the multi-head model as a generalization of the multi-kernel. Multi-head can also be seen as an ensemble of (local mean) autoencoders, or multi-task learning for autoencoders.

\subsection{Discrete Kernels}

Let $\mathcal{X}$ be a discrete sample space, and the matrix $\matr{K}:\mathcal{X}\times \mathcal{X}$ represent the kernel on this space. In this paper, the concept of a discrete kernel always means that the elements of $\matr{K}$ are learnable unknown parameters. A typical discrete kernel is
\begin{equation*}
\matr{K}=\bmPhi\bmPsi^{\mathrm{T}} ~\text{or}~ \mathrm{softmax}({\bmPhi\bmPsi^{\mathrm{T}}}),
\end{equation*}
where $\bmPhi=\phi(\matr{X}), \bmPsi=\psi(\matr{X}): N\times d$ are feature matrices, and there are at most $2d|\mathcal{X}|$ real-valued parameters (which can be constrained to be non-negative). Similar to kernel methods, it should be ensured that $d<|\mathcal{X}|$.

Here, we primarily construct a local model based on discrete kernels, which is the self-local mean/mode adaptive form and can serve as the foundation for large language models.

\subsubsection{Discrete kernel self-local mean}

First, introduce the embedding $v:\mathcal{X}\to \R^q$, and the feature mappings $\phi, \psi:\mathcal{X}\to \R^d$ to form query-key-value representations (see \autoref{df:qkv}). Let $\matr{V}=v(\matr{X})$. If $\matr{V}$ is also learnable, the model will have at most $|\mathcal{X}|(2d+q)$ parameters.

\begin{definition}[Discrete Kernel Self-Local Mean]\label{df:discrete-local}
A \term{discrete kernel (self) local mean} model refers to the solution to the following optimization problem about self-local mean:
  \begin{gather}\label{discrete-local}
  \min_{\phi,\psi}\sum_i\|v(x_i) - \hat v(x_i)\|_2^2 \nonumber \\
  \hat v(x_i):=(\phi(x_i)\cdot \sum_j\psi(x_j)v(x_j)) / (\phi(x_i)\cdot \sum_j\psi(x_j)) , \nonumber\\
  \text{that is} ~ \min_{\bmPhi,\bmPsi}\|\matr{V} - \hat{\matr{V}}\|_F^2, \hat{\matr{V}} := \bmPhi\bmPsi^{\mathrm{T}}\matr{V} \oslash (\bmPhi\bmPsi^{\mathrm{T}}\one_{N\times q}), \\
  \bmPhi,\bmPsi: N\times d,\matr{V}: N\times q, \nonumber
  \end{gather}
where $\{x_i\}$ are the samples (and $\matr{V}$ can either be pre-defined or optimized simultaneously).
\end{definition}

If the feature mappings take the form of \fref{local-feature-exp}, \fref{discrete-local} becomes the classic form of the query-key-value model \cite{vaswani2017}:
\begin{equation}\label{softmax-dis-local}
  \min_{\bmPhi,\bmPsi}\|\matr{V} - \softmax(\bmPhi\bmPsi^{\mathrm{T}})\matr{V}\|_F^2.
\end{equation}

Similar to kernel methods, the concept of a discrete kernel can also be applied to continuous spaces (typically subsets of $\R^p$). We directly provide the following definition.

\begin{definition}[Discrete Kernel Local Decision Model]
  \begin{equation}
  \min_{\matr{K}}\sum_{i}l(x_i,\hat{\theta}(x_i;\matr{K})),
  \end{equation}
  where the kernel matrix $\matr{K}=\{K(x_i,x_j)\}$ is an $N$-dimensional matrix, which should have reasonable constraints, and $\hat{\theta}(x_i;\matr{K})$ is the solution to the solution to the following optimization problem
  \begin{equation}
  \min_\theta \big\{J(x_i,\theta; \matr{K}):=\sum_j\matr{K}_{ij}l(x_j,\theta)\big\}.
  \end{equation}
\end{definition}

\begin{remark}
If the kernel function $K(\cdot)$ is not explicitly given, the discrete kernel local model will no longer be responsible for predicting points outside the samples. This is the same as the discrete kernel model in kernel methods.
\end{remark}

\subsubsection{Discrete kernel temporal self-Local mean}

If the functions $\phi,\psi$ in \fref{discrete-local} depend on the sample index $t$, i.e., defined on $\mathcal{X} \times \mathcal{T}$ where $\mathcal{T}$ is the index set, we naturally obtain the term \term{Discrete Kernel Temporal (Self) Local Mean}.

\begin{definition}[Discrete Kernel Temporal Self-Local Mean]\label{df:discrete-local-ts}
A discrete kernel temporal self-local mean model refers to the solution to the following optimization problem concerning self-local mean:
  \begin{gather}\label{discrete-local-ts}
  \min_{\phi,\psi} \sum_t\|v(x_t) - \hat v(x_t)\|_2^2 \nonumber \\
  \hat v(x_t):=(\phi(x_t,t)\cdot \sum_s\psi(x_s,s)v(x_s)) / (\phi(x_t,t)\cdot \sum_s\psi(x_s,s)) , \nonumber\\
  \text{that is} ~ \min_{\bmPhi,\bmPsi}\|\matr{V} - \hat{\matr{V}}\|_F^2, \hat{\matr{V}} := \bmPhi\bmPsi^{\mathrm{T}}\matr{V} \oslash (\bmPhi\bmPsi^{\mathrm{T}}\one_{N\times q}), \\
  \bmPhi,\bmPsi: N\times d,\matr{V}: N\times q, \nonumber
  \end{gather}
  where $\{x_t,t=1,\cdots, T\}$ is a time series.
\end{definition}

The parameter size of this model generally will not exceed $O(T)$, and may even be limited to $O(|\mathcal{X}|d)$. If a single-layer neural network is directly constructed on the sequence space $\mathcal{X}^T$, the parameter size could reach $O(T^2)$.

\subsection{Discrete Kernel Local Mean Autoencoder}\label{sec:discrete-local-ae}

The matrix parameters in the autoencoder \eqref{loc-ave-ae} should be adaptively determined by the samples, that is,
\begin{equation}\label{discrete-local-ae}
  \min_{\bmPhi,\bmPsi,\matr{W}} d(\matr{X}, \hat{\matr{V}}\matr{W}^{\mathrm{T}}),
\end{equation}
where $\matr{X}$ is the design matrix, and $d$ is a suitable distance (divergence or Euclidean distance). (The matrix $\matr{V}$ can be predefined or optimized simultaneously.) The multi-head form of \fref{discrete-local-ae} is
\begin{equation}\label{discrete-local-mh}
\min_{\substack{\bmPhi_m, \bmPsi_m, \matr{W}_m \\ m=1,\cdots,M}} d(\matr{X}, \frac{1}{M}\sum_m\hat{\matr{V}}_m\matr{W}_m^{\mathrm{T}}).
\end{equation}

If the kernels or feature mappings in \fref{discrete-local-ae} depend on time, it is called a discrete kernel temporal autoencoder. This is exactly the \term{self-attention layer} used in modern language models.

\begin{experiment}\label{ex:qkv}
  Load the document \textit{Oliver Twist} from the project Gutenberg, accessible at \href{https://www.gutenberg.org/ebooks/730}{https://www.gutenberg.org/ebooks/730}, and convert it into several independent word sequences. The total loss is the sum of the autoencoder reconstruction losses for all word sequences. We calculate the optimal temporal query-key-value representations $(\bmPhi,\bmPsi,\matr{V})$ based on \fref{discrete-local-ae}, where $\matr{V}$ is predefined using the word embeddings from CBOW.
\end{experiment}

\begin{figure}[h]
  \centering
  \begin{subfigure}[t]{0.49\textwidth}
    \centering
    \includegraphics[width=\textwidth]{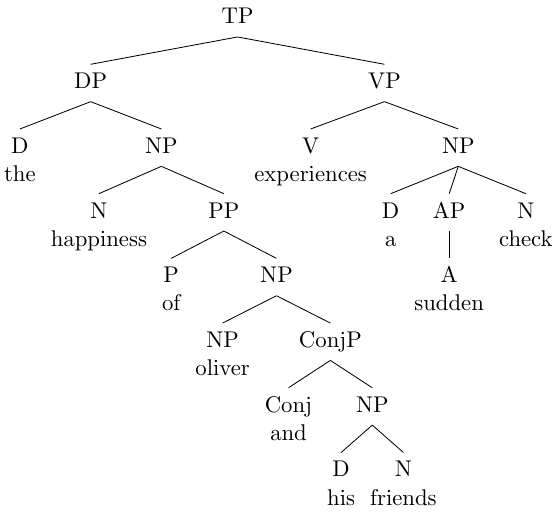}
    \caption{Illustration of Generative Grammar}
    \label{fig:gen-syntax}
  \end{subfigure}
  \begin{subfigure}[t]{0.49\textwidth}
    \centering
    \includegraphics[width=\textwidth]{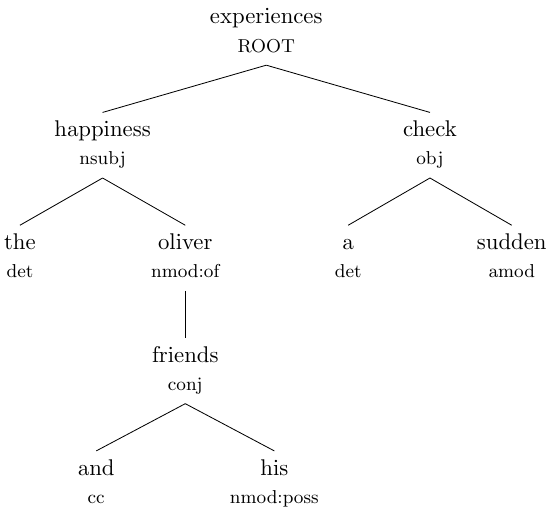}
    \caption{Illustration of Dependency Grammar}
    \label{fig:dep-syntax}
  \end{subfigure}
  \caption{Generative grammar and dependency grammar (based on the sentence in \autoref{ex:qkv})}
  \label{fig:syntax}
\end{figure}

Generative grammar, proposed and developed by the famous linguist N. Chomsky \cite{chomsky1965,chomsky1995,cohen2022}, can be viewed as a generative model with latent variables. It actually provides a hierarchical (and recursive) query–key–value representation for word sequences: leaf nodes (\term{terminal symbols}) correspond to observable variables, while internal nodes (\term{non-terminal symbols}) correspond to latent variables, and edges correspond to feature mappings/generation probabilities (\asref{fig:gen-syntax}). In contrast, \term{dependency grammar} \cite{osborne2020,shen2021structformer,zhao2024dependency} is more appropriately described as a tree-structured (at least a sparse graph, rather than a complete graph) kernel–value representation, which need not explicitly specify latent variables or feature mappings (\asref{fig:dep-syntax}). From a formal perspective, the hierarchical structure induced by generative grammar is essentially recursive; a detailed investigation of this property is deferred to future work.

\subsection{Asymmetric Multi-Dimentional Scaling}

Similar to kernel methods, we introduce the problem of approximate low-dimensional feature mapping (low-rank decomposition of kernel) for localization methods:
\begin{equation}\label{approx-ld-feature}
K(x,x') \approx \phi(x)\cdot \psi(x'),
\end{equation}
or more generally, $K(x,x') \approx F(\phi(x), \psi(x'))$.

\begin{definition}[Asymmetric MDS]\label{df:mds-local}
Given the kernel $K$, the strain loss function is constructed as follows,
\begin{equation}\label{strain-local}
  \mathrm{strain}(\{\vect{\phi}_i\},\{\vect{\psi}_i\}):=\sum_{ij}d(\vect{\phi}_i\cdot\vect{\psi}_j,K(x_i,x_j)), \vect{\phi}_i,\vect{\psi}_i\in \R^q,
\end{equation}
where $\vect{\phi}_i,\vect{\psi}_i\in \R^q$ represent the discrete feature mappings corresponding to the kernel $K$, and $d$ is a suitable distance. The \text{Multi-Dimentional Scaling (MDS)} is the optimization problem of minimizing \fref{strain-local}. It provides two low-dimensional embeddings for $x_i$, namely $x_i \mapsto \vect{\phi}_i$ and $x_i \mapsto \vect{\psi}_i$.
\end{definition}

Under the squared loss, if the strain function has no further constraints, the $\{\vect{\phi}_i\}$ and $\{\vect{\psi}_i\}$ in \fref{strain-local} can be estimated using SVD. If there are non-negative constraints, \term{Non-Negative Matrix Factorization (NMF)} can be used to solve it.

Since the localization kernel $K$ need not be symmetric and is decomposed into two feature mappings, we refer to the MDS based on localization kernels as \term{Asymmetric MDS} to distinguish it from the MDS in classical settings \cite{saeed2018}.

\begin{remark}
The concept of \term{local MDS} was proposed by \cite{chen2009}. However, this model is a modification of the strain function, applying the localization trick to the strain function.
\end{remark}

\begin{figure}[h]
  \centering
  \includegraphics[width=0.65\textwidth]{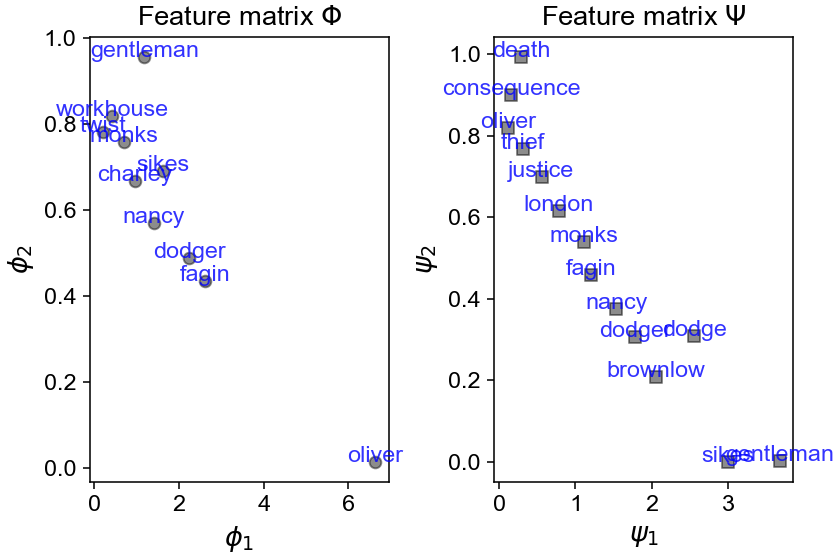}
  \caption{Demonstration of Asymmetric MDS plane embedding}
  \label{fig:amds}
\end{figure}

\begin{experiment}\label{ex:amds}
Utilize the documents from \autoref{ex:qkv}, and set $K(x,y):=\ln(1+\mathrm{e}^{N(x,y)})$, where $N(x,y)$ is the count of how often words $x$ and $y$ appear in the same paragraph in the document. This results in an asymmetric kernel. We introduce non-negative constraints and thus perform NMF decomposition on the kernel matrix. The plane embedding result is shown in \autoref{fig:amds}.
\end{experiment}

\begin{remark}
  Isometric Mapping (ISOMAP)\cite{tenenbaum2000,ghojogh2023}, Laplacian Eigenmap (LE)\cite{belkin2001,luo2009},
Stochastic Neighbor Embedding (SNE)\cite{hinton2002,vandermaaten2008}, Locality Preserving Projection (LPP)\cite{he2003}, Neighborhood Component Analysis (NCA)\cite{goldberger2004} and Maximum Variance Unfolding (MVU)\cite{sun2006} are several classical low-dimensional embedding methods. They can all be regarded as special forms of MDS with different distances and other special settings. For more embedding methods, please refer to \cite{mcinnes2018,nanga2021,wang2021}.
\end{remark}

\subsection{Adaptive Feature Mapping}

The adaptive feature mapping here refers to directly adaptive localization models based on feature mappings. Following is the moste well-known example.

\begin{definition}[CBOW \cite{mikolov2013a}]\label{ex:cbow}
Review \autoref{ex:wv}. Construct the adaptive form of \fref{cbow}:
  \begin{equation}\label{cbow-adaptive}
  \max_{\matr{V}_I,\matr{V}_O\in \R^{|\mathcal{W}|\times d}}\sum_i \big\{\vect{\delta}_{c_i}^{\mathrm{T}}\matr{V}_I\matr{V}_O^{\mathrm{T}} \vect{\delta}_{w_i} - \ln \sum_{w\in \mathcal{W}}\mathrm{e}^{\vect{\delta}_{c_i}^{\mathrm{T}}\matr{V}_I\matr{V}_O^{\mathrm{T}} \vect{\delta}_{w}}\big\},
  \end{equation}
  where $\matr{V}_I := \{v_I(w),w\in\mathcal{W}\}: N\times d$ is \term{input representation} of the word $w$ (query feature matrix), $\matr{V}_O := \{v_O(w),w\in\mathcal{W}\}: N\times d$ is \term{output representation} of the word $w$ (key feature matrix).
\end{definition}

\begin{figure}[h]
  \centering
  \includegraphics[width=0.7\textwidth]{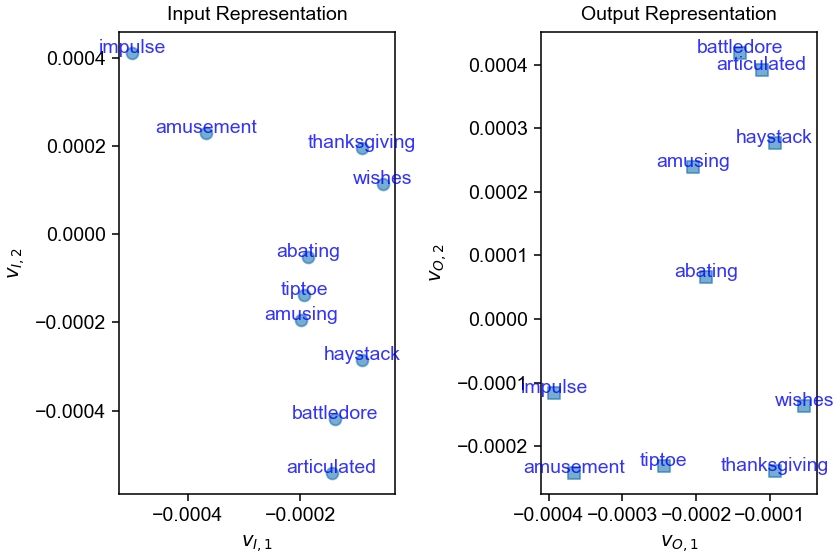}
  \caption{CBOW word embedding effect}
  \label{fig:cbow}
\end{figure}

Direct optimization of \fref{cbow-adaptive} is not easy. We demonstrate the word embedding effects of CBOW based on the following fact, \asref{fig:cbow}.

\begin{fact}
    If we minimize only the first term of \fref{cbow-adaptive} and enforce that $\matr{V}_I$ and $\matr{V}_O$ are orthogonal matrices, the word embedding reduces to the SVD of the co-occurrence matrix $\bmDelta_{w}^{\mathrm{T}}\bmDelta_{c}: |\mathcal{W}|\times |\mathcal{W}|$, where $\bmDelta_w := (\vect{\delta}{w_1}, \cdots, \vect{\delta}{w_N})^{\mathrm{T}}$, which is the matrix of size $N \times |\mathcal{W}|$ formed by concatenating the row vectors $\vect{\delta}_{w_i}$ (the feature matrix under one-hot encoding), and the definition of $\bmDelta_c$ is similar.
\end{fact}

\section{Other Topics}

Before concluding the paper, we provide some extensions of local models/localization methods, including the construction of Transformer.

\subsection{Denoising Autoencoder}\label{sec:dae}

The following fact is a direct corollary of \autoref{th:id-approx}.

\begin{fact}\label{fc:sample-noising}
  Let $\{\vect{x}_i\}$ be samples from an unknown distribution $p$. Sampling from $p$, $\tilde{\vect{x}}_i\sim p$, can be approximated by simulating $\hat p$ as:
 \begin{equation}\label{sample-kernel}
  \tilde{\vect{x}}_i|\vect{x}_i\sim K_h(\tilde{\vect{x}}_i,\vect{x}_i),\vect{x}_i\sim p, h\to 0.
  \end{equation}
  Specifically, when the density function $K$ (e.g., Gaussian distribution) acts as a convolution kernel, the approximation is:
  \begin{equation}\label{sample-noising}
  \tilde{\vect{x}}_i= \vect{x}_i+\vect{\epsilon}_i,\vect{\epsilon}_i\sim K_h, h\to 0,
  \end{equation}
  where $\vect{\epsilon}_i$ is the noise/disturbance, independent of $\vect{x}_i$.
\end{fact}

\autoref{fc:sample-noising} provides a \term{data augmentation} method. Suppose we are training a classifier using the sample set $\{(\vect{x}_i,y_i),i=1,\cdots,N\}$. If the sample size is too small and the training performance is poor, we can now generate an unlimited number of augmented samples $\{(\tilde{\vect{x}}_{il},y_i),i=1,\cdots,N,l=1,\cdots,L\}$ using \fref{sample-noising}, thereby improving training performance.

\begin{definition}\label{df:dae}
For the sample $\{\vect{x}_i\}$, noise can be added to generate self-supervised samples $\{(\tilde{\vect{x}}_{il},\vect{x}_i)\}$ (without adding noise to the output). A \term{Denoising Autoencoder (DAE)}\cite{vincent2008,alain2016} is an autoencoder that learns from these augmented samples.
\end{definition}

Let $T(\cdot)$ be the reconstruction of the DAE.
For the noise addition method in \fref{sample-noising}, training the DAE is equivalent to training the model $T(\vect{x}) - \vect{x}$ using self-supervised samples $\{(\tilde{\vect{x}}_{il},\vect{\epsilon}_{il})\}$.

\begin{figure}[h]
  \centering
  \includegraphics[width=0.5\textwidth]{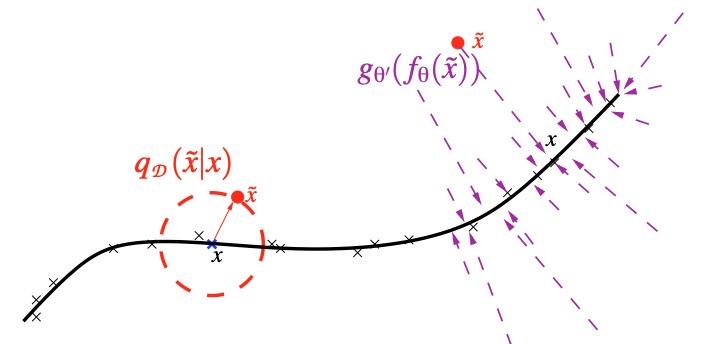}
  \caption{Diagram of DAE Data Operation (quoted from \cite{vincent2010}; original notation differs slightly from this paper)}
  \label{fig:dae}
\end{figure}

Formally, the DAE can be viewed as a \term{variational autoencoder (VAE)} where the encoder is $T(\cdot)$ and the decoder is $p(\tilde{\vect{x}}|\vect{x})$. Noisy data means that only $\tilde{\vect{x}}$ is observed, while $\vect{x}$ is treated as a latent variable, though it can also be utilized. If $T(\cdot)$ itself is composed of an encoder $\phi:\mathcal{X}\to\mathcal{Z}$ and a decoder $\psi:\mathcal{Z}\to\mathcal{X}$, where $\mathcal{Z}$ is the latent space, then $\phi$ should be the encoder, and $p(\tilde{\vect{x}}|\vect{x})$ should be combined with $\psi$ to form the (probabilistic) decoder. See \autoref{fig:dae} for reference.

\begin{remark}\label{ob:dae}
Simply put, the autoencoder/reconstruction $T(\cdot)$ can be viewed as the inverse operation of noise addition, that is, approximating the conditional/variational/transition distribution $p(\absv{x}|\tilde{\absv{x}})\sim K(\tilde{\absv{x}},\absv{x})$. Intuitively, it can be expressed as:
\begin{equation*}
\ce{$x$ <=>[$p(\tilde{x}|x)\sim K(\tilde{x},x)$][$x=T(\tilde{x})$]{$\tilde{x}$}}
\end{equation*}
or (to emphasize the use of the latent variable $\absv{z}$):
\begin{center}
\begin{tikzpicture}
    \node (x) at (0,1) {$x$};
    \node (x') at (2,1) {$\tilde{x}$};
    \node (z) at (1,0) {$z$};
    \draw[->] (x) -- node[above, scale=0.6] {$p(\tilde{x}|x)\sim K(\tilde{x},x)$} (x');
    \draw[->] (x') -- node[right] {$\phi$} (z);
    \draw[->] (z) -- node[left] {$\psi$} (x);
\end{tikzpicture}
\end{center}
\end{remark}

According to \autoref{ob:dae}, we add noise to the point $\absv{x}_0$ (or sample from the distribution $p(\absv{x}|\tilde{\absv{x}})\sim K(\tilde{\absv{x}},\absv{x})$), obtaining $\tilde{\absv{x}}_0$, and then reconstruct $\tilde{\absv{x}}_0$ to obtain $\absv{x}_1$. $\absv{x}_1$ should be closer to the true sample than $\tilde{\absv{x}}_0$, so it can replace (or mix into) the original sample. Then $\tilde{\absv{x}}_1$ can be generated by similar manners. This iterative process forms the \term{DAE-Markov Chain}\cite{alain2016}:
\begin{equation*}
  \absv{x}_0\to\tilde{\absv{x}}_0\to \absv{x}_1\to\tilde{\absv{x}}_1\to\cdots\to \absv{x}_t\to\tilde{\absv{x}}_t\to\cdots,
\end{equation*}
or $\absv{x}_0\to\tilde{\absv{x}}_0\to z_0 \to \absv{x}_1\to\cdots$ to emphasize the application of the latent variable $\absv{z}$.

We would like to call the process \term{Noising-Denoising (Alternating) Iterative Algorithm}. During the iteration process, the newly generated samples can be used to retrain the DAE, and the trained DAE can be used to generate subsequent samples, forming a noising-estimation-denoising alternating iterative process.

\begin{fact}\label{fc:gsn}
  If the reconstruction $T(\cdot)$ can arbitrarily approximate the conditional distribution $p(\absv{x}|\tilde{\absv{x}})$, then the stationary distribution of the above DAE-Markov chain is the true distribution $p(\absv{x})$. In fact, this is equivalent to performing Gibbs sampling on the joint distribution $p(\tilde{\absv{x}},\absv{x})\sim K(\tilde{\absv{x}},\absv{x})$.
\end{fact}

\autoref{fc:gsn} shows that the DAE-Markov chain provides a distribution simulation method that can improve upon the method in \autoref{fc:sample-noising}. This is exactly the basic principle of the \term{Generative Stochastic Network (GSN)}\cite{alain2016}.

\subsection{Score Models}\label{sec:score}

From \fref{meanshift-kde}, we know that mean shift $\lap_K(\vect{x})$ approximates score function $\nabla_{\vect{x}} \ln p(\vect{x})$. Specifically, we have the following result in which the (expectation of) mean shift is proportional to the score function.

\begin{theorem}[Tweedie's Formula \cite{efron2011}]\label{th:tweedie}
  Let $\vect{z} \sim p_{\vect{z}}$ and $\vect{x}|\vect{z} \sim N(\vect{z}, \sigma^2)$, where $\sigma^2>0$ is known. Clearly, the marginal distribution of $\vect{x}$ is $p(\vect{x})=p_{\vect{z}}*\phi_\sigma$, where $\phi_\sigma$ is the Gaussian density function/kernel. Then we have the famous \term{Tweedie's formula}:
\begin{equation}\label{tweedie}
\Exp(\vect{z}|\vect{x}) = \vect{x} + \sigma^2\nabla_{\vect{x}} \ln p(\vect{x}),
\end{equation}
\end{theorem}

\begin{definition}\label{df:score}
  A model is called a \term{score-based model} \cite{hyvarinen2005,vincent2011} if it is fundamentally based on scores or its primary task is to estimate the score of the true distribution.
\end{definition}

According to this definition and \autoref{fc:meanshift-kde}, the MeanShift is a score-based model. We also have the following useful formula for scores \cite{bao2022}.

\begin{lemma}\label{ft:score}
  \begin{align}
  \nabla_{\vect{x}} \ln p(\vect{x}) &= \Exp_{p(\vect{z}|\vect{x})}\nabla_{\vect{x}} \ln p(\vect{x}|\vect{z})\label{score}\\
  \Exp_{\vect{x}}\|F(\vect{x})-\nabla_{\vect{x}} \ln p(\vect{x})\|_2^2 &\sim \Exp_{p(\vect{x},\vect{z})}\|F(\vect{x})-\nabla_{\vect{x}} \ln p(\vect{x}|\vect{z})\|_2^2 \label{score-match}
  \end{align}
\end{lemma}

\autoref{ft:score} shows that score estimation can be transformed into estimating the conditional score $\nabla_{\vect{x}} \ln p(\vect{x}|\vect{z})$, where the conditional distribution $p(\vect{x}|\vect{z})$ can be artificially designed and does not depend on the true distribution $p(\vect{x})$. Specifically, when $p(\vect{x}|\vect{z})\sim N(\vect{z},\sigma^2)$, \fref{score} derives Tweedie's formula. In this case, approximating the score function using the model $F(\vect{x}):\R^p\to\R^p$ is equivalent to fitting the self-supervised data $\{(\tilde{\vect{x}}_i,-\vect{\epsilon}_i)\}$, where $\tilde{\vect{x}}_i=\vect{x}_i+\sigma\vect{\epsilon}_i, \vect{\epsilon}_i\sim N(0,1)$. This is equivalent to using DAE, $T(\vect{x})=\vect{x}+\sigma^2 F(\vect{x})$, for self-supervised learning (see \autoref{df:dae}).

\begin{fact}\label{fc:score-dea}
  Score models and DAE can be considered equivalent, and Tweedie's formula shows that this DAE is essentially a local mean autoencoder. In other words, sample noising and local mean can be regarded as two operations that are inverse transformations of each other\cite{botev2007}, expressed as:
\begin{equation*}
\ce{x <=>[$p(\tilde{x}|x)\sim K(\tilde{x},x)$][$p(x|x_*)\sim K(x_*,x)$]{$\tilde{x}~\text{or}~x_*$}}
\end{equation*}
\end{fact}

\begin{remark}\label{rm:meanshift-sem}
In \autoref{rm:meanshift-em}, we have pointed out the connection between the MeanShift algorithm and the EM algorithm. Now DAE based on the local mean autoencoder derives the MeanShift algorithm with sample noising. Therefore, roughly speaking, it can be viewed as a stochastic EM algorithm:
\begin{enumerate}
  \item E-step: The same as that in \autoref{rm:meanshift-em};
  \item M-step: Involves KDE $\hat{p}_K$ that requires sample noising.
\end{enumerate}
Considering the relationship between the EM algorithm and \term{active inference}, when an intelligent system (such as the human brain) adapts to or interacts with the environment, it is equivalent to adding noise to the beliefs in the consciousness. Meanwhile, the perceptual inference in the consciousness is akin to computing the local mean of the perceptual results to form new beliefs.
\end{remark}

\subsection{Local Mean Diffusion Model}\label{sec:local-mean-diffusion}

\autoref{fc:score-dea} explains the principle of the DAE based on the local mean. We can use \autoref{fc:gsn} to construct a distribution simulation method. In this section, we design an iterative simulation method with a different approach.

We first gradually add noise to the original sample $\matr{X}_0 = \{\vect{x}_i\} \sim p_0$, obtaining $\matr{X}_t, t=1, \cdots, T$. This process forms a Markov chain and ultimately converges to the \term{stationary distribution} $p_\infty \approx p_T$. Then, for each step, we perform the inverse operation, which is the local mean:
\begin{equation}\label{dm-inv}
  \matr{X}_{t-1}' \leftarrow \slm_K(\matr{X}_{t}'; \matr{X}_{t-1}),
\end{equation}
where $\matr{X}_{T}'$ is sampled from $p_T$ (the regularized form \eqref{alpha-meanshift} is recommended). As the result of the iteration, $\matr{X}_{0}'$ approximately simulates $p_0$. This gives a generative model, which can be referred to as the \term{local mean diffusion model}\cite{botev2007}. It essentially replaces the neural network in the \term{diffusion model}\cite{sohl2015,ho2020,song2021score} with a local mean autoencoder.

Different noise addition procedures can be used to generate various local mean diffusion models. Here, we only consider the Gaussian setting.

\begin{definition}[Gaussian Local Mean Diffusion Model]\label{df:gaussian-localmean-diffusion}
  Gradually add Gaussian noise to the given sample $\matr{X}_0$:
  \begin{gather*}
  \vect{x}_t = a_t\vect{x}_{t-1} + \vect{\epsilon}_{t} ~\text{or}~ b_t\vect{x}_{0} + \vect{\eta}_{t}\\
  \vect{x}_0 \in \matr{X}_0, \vect{\epsilon}_{t} \sim N(0, \sigma_t^2), \vect{\eta}_{t} \sim N(0, s_t^2), t = 1, \cdots, T
  \end{gather*}
  where the scaling factor $0 < a_t \leq 1$ controls the variance of the data, $0 < \sigma_t^2 < 1$ is the noise variance (increasing as $t$ increases), and $b_t, s_t^2$ are functions of $a_t, \sigma^2_t$. This noise addition process forms a \term{Brownian motion}, with a stationary distribution that is Gaussian. We assume that $\matr{X}_T$ follows an easy-to-sample distribution $N(0, \sigma^2)$, where $\sigma^2 > 0$ is known. Now, generate $\matr{X}_T' \sim N(0, \sigma^2)$, and then generate $\matr{X}_{t}'$ in reverse using \fref{dm-inv} and the noise addition process. The final $\matr{X}_{0}'$ is the model-generated sample.
\end{definition}



\begin{remark}\label{rm:sde}
  \cite{song2021score} points out the connection between diffusion models and \term{Stochastic Differential Equations (SDEs)} \cite{song2021score}, namely that diffusion models are numerical simulations of diffusion processes. According to this fact, the noise addition is the numerical simulation of a SDE, while the MeanShift iteration with noised samples in \fref{dm-inv} is the numerical simulation of the relative reverse-time process, \asref{tab:sde-lm}. Also see \autoref{fc:score-dea}.
\end{remark}

\begin{table}[h]
    \centering
    \caption{The correspondence between SDEs and local mean}\label{tab:sde-lm}
    \begin{tabular}{c|c|c}
        \toprule
        \thbf{Models} & \thbf{the process} & \thbf{inverse process} \\
        \midrule
        SDE & simulation of SDE &  simulation of reverse-time SDE\\
        local mean & noise addition to samples & MeanShift \\
        DAE & noise addition to samples & reconstruction \\
        \bottomrule
    \end{tabular}
\end{table}

\subsection{Local Linear Embedding}\label{sec:lle}

MDS provides a method for low-dimensional/plane embedding. We now introduce another classic model with similar functionality, as an application of discrete kernels: \term{Local Linear Embedding (LLE)}\cite{roweis2000}. The original idea of LLE is to introduce an ``invariant'' of the embedding operation, namely the discrete kernel.

\begin{definition}[Local Linear Embedding]\label{df:lle}
  LLE first constructs a discrete kernel local mean optimization problem:
  \begin{equation}\label{lle-ker}
  \min_{\tilde{\matr{K}}} \|\matr{X} - \tilde{\matr{K}}\matr{X}\|_F^2,
  \end{equation}
  where $\matr{X} \in \R^{N \times p}$ is the design matrix, $\tilde{\matr{K}}$ is a stochastic matrix, and if the sample point $\vect{x}_j$ is far from $\vect{x}_i$, then $\tilde{\matr{K}}_{ij} = 0$ (i.e., a normalized neighborhood kernel matrix). Then, solve the following approximate fixed-point problem:
  \begin{equation}\label{lle-z}
  \min_{\matr{Z} \in \R^{N \times r}, r < p} \|\matr{Z} - \tilde{\matr{K}}\matr{Z}\|_F^2.
  \end{equation}
  Finally, the solution $\matr{Z}$ to \fref{lle-z} is the low-dimensional embedding of $\matr{X}$.
\end{definition}

LLE provides an equivalent form of MDS, but it is typically limited to adaptive neighborhood kernels, and calculates low-dimensional embeddings by minimizing local mean reconstruction errors, rather than through low-rank decomposition. In relation to \autoref{rm:transition}, LLE treats the kernel as an approximate invariant for low-dimensional embedding, essentially assuming that the low-dimensional data $\matr{Z}$ preserves the state transition relationships within the original data $\matr{X}$.

It is not difficult to see that for data $\matr{X}$, PCA gives a suboptimal solution to \fref{lle-z}. It is anticipated that the dimensionality reduction effect of LLE will surpass that of PCA. Since the largest eigenvalue of $\tilde{\matr{K}}$ is 1, the solution to \fref{lle-z} is close to the SVD of $\tilde{\matr{K}}$. It is hypothesized that LLE's result is close to kPCA based on $\tilde{\matr{K}}$. Finally we can also combine the two optimization problems in \autoref{df:lle}:
\begin{equation}\label{lle-joint}
  \min_{\tilde{\matr{K}}, \matr{Z}} \|\matr{X} - \tilde{\matr{K}}\matr{X}\|_F^2 + \|\matr{Z} - \tilde{\matr{K}}\matr{Z}\|_F^2.
\end{equation}

\begin{figure}[h]
  \centering
  \begin{subfigure}[t]{0.49\textwidth}
    \centering
  \includegraphics[width=\textwidth]{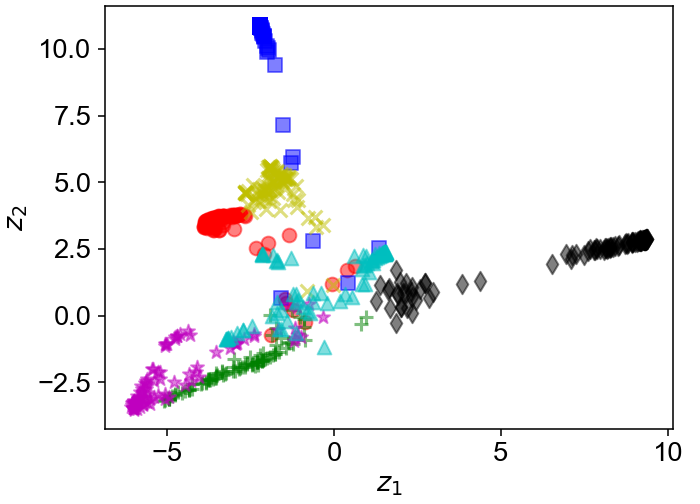}
  \caption{Without using label information}\label{fig:lle-2d}
  \end{subfigure}
  \begin{subfigure}[t]{0.49\textwidth}
    \centering
    \includegraphics[width=\textwidth]{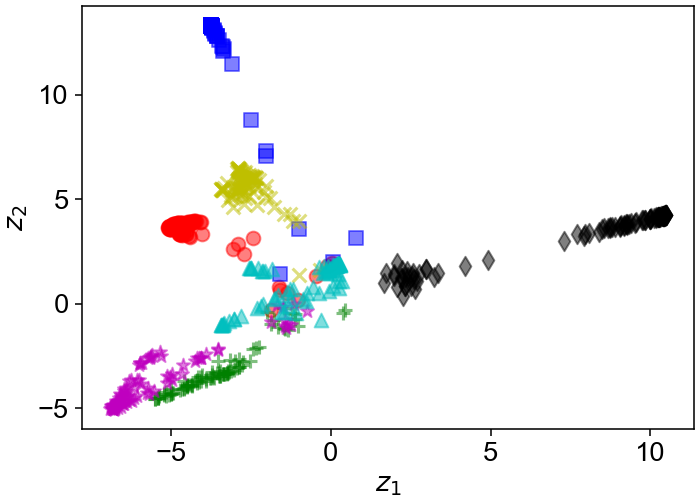}
    \caption{Using label information}\label{fig:lle-labeled-2d}
  \end{subfigure}
  \caption{LLE Dimensionality Reduction / Plane Embedding Results (fashion mnist image dataset)}
  \label{fig:lle}
\end{figure}

\begin{remark}\label{ob:lle}
\autoref{df:lle} is the original definition of LLE \cite{roweis2000}. We can make simple modifications, such as replacing the discrete kernel with a general parameterized kernel or multiple kernels.
Compared to MDS, LLE must first learn the discrete kernel, but it can provide a fixed kernel, obviating the need to solve the optimization problem in \fref{lle-ker}. The core of LLE is the optimization problem in \fref{lle-z}, while \fref{lle-ker} is just an adaptive kernel scheme. \autoref{fig:lle} shows a plane embedding using a fixed Epanechnikov kernel, which achieves good results.
\end{remark}

\subsection{Hierarchical Local Models}\label{sec:hie-loc}

A local model is determined by a local loss function $J(x_*,\theta;K)$, which itself is a loss function at a single sample point. Based on this, another local loss can be constructed as follows:
\begin{align}\label{local-local}
J'(x_*,\theta;K,K') &= \sum_k K'(x_*, x_k) J(x_k, \theta; K) \nonumber \\
&= \sum_{kj} K'(x_*, x_k) K(x_k, x_j) l(x_j, \theta).
\end{align}
The matrix form of this is $J'(\matr{X}, \theta) = K'(\matr{X}, \matr{X}) K(\matr{X}, \matr{X}) l(\matr{X}, \theta)$. These two losses are not the same. We call \fref{local-local} the local-local loss, and the corresponding decision model is the local-local model.

It is easy to define the ``local-local mean'' as:
\begin{equation}\label{local-local-ave}
\hat{\vect{x}}_i = \sum_{kj} \tilde{\matr{K}}'_{ik} \tilde{\matr{K}}_{kj} \vect{x}_j ~ \text{or} ~ \hat{\matr{X}} = \tilde{\matr{K}}' \tilde{\matr{K}} \matr{X},
\end{equation}
where $\tilde{\matr{K}}$ and $\tilde{\matr{K}}'$ are the normalized kernel matrices of $K(\matr{X}, \matr{X})$ and $K'(\matr{X}, \matr{X})$, respectively.

Obviously, the local-local model/mean can still be seen as a local model/mean with the composite kernel $\int K'(x,z) K(z,y) \, dz$ or $\sum_k K'(x, x_k) K(x_k, y)$.

\begin{definition}[Local-Local Mean] \label{df:local-local}
To avoid the degeneration of \fref{local-local-ave} into the standard local model, we introduce a nonlinear function $f: \R^p \to \R^p$ between the two kernels, and consider $\tilde{\matr{K}}'$ as the kernel matrix corresponding to $f(\tilde{\matr{K}} \matr{X})$, instead of $\matr{X}$. Now the local-local mean is defined as:
\begin{equation}\label{local-local-ave-f}
\hat{\vect{x}}_i = \sum_k \tilde{\matr{K}}'_{ik} f\left( \sum_j \tilde{\matr{K}}_{kj} \vect{x}_j \right) ~ \text{or} ~ \hat{\matr{X}} = \tilde{\matr{K}}' f(\matr{KX}),
\end{equation}
where $\tilde{\matr{K}}'$ is the normalized kernel matrix of $K'(\matr{X}', \matr{X}')$, and $\matr{X}' = f\left( \sum_j \tilde{\matr{K}}_{kj} \vect{x}_j \right) \neq \matr{X}$ (through pointwise operations $f(\{\vect{x}_i\}) = \{ f(\vect{x}_i) \}$, with $f$ automatically applied to $\R^{N \times p}$).
\end{definition}

\begin{remark}
  One should not confuse \fref{local-local-ave-f} with the traditional $pN$ input-$pN$ output fully connected feedforward neural network, because the network weights $\matr{K}$ have a special form and do not traverse the space $\R^{pN \times pN}$.
\end{remark}

\begin{definition}
Clearly, we can further construct ``local-local-local models'' and so on. These models are collectively referred to as \term{hierarchical local models}.
\end{definition}

\begin{example}
  The missing data imputation model in \autoref{ex:missing-lm} can be regareded as a hierarchical local model composed of pixel-level and image-level local means. These hierarchical local models at different levels can be called the \term{cascaded local models}. For the task of text processing, we can construct cascaded local models based on the word-level and the document-level local means.
\end{example}

\begin{remark}
A hierarchical local model can be regarded as directly conducting numerical simulation on the reverse process of the SDE, using different kernel functions at each step. Therefore, a deep hierarchical local model usually does not need to perform prediction/reconstruction iteration like a common local mean when making predictions/reconstructions (see \autoref{sec:autoencoder}).
\end{remark}

The famous Transformer model is a type of hierarchical (sequence) local model, where a two-layer neural network is embedded between the two local means. See the definition below.

\begin{definition}\label{df:transformer}
The encoder of a Transformer can be simply represented by the following sequence mapping:
\begin{equation}\label{transformer}
\{\hat{\vect{x}}_t\} = f_L \left( m_{K_L}(\cdots f_1(m_{K_1}(\{\vect{x}_t\})) \cdots) \right),
\end{equation}
where $m_{K_l}$ is the (sequence) local mean transformation of kernel $K_l$, and $f_l: \R^{\mathrm{T}} \to \R^{\mathrm{T}}$ is a two-layer feedforward neural network. (In \cite{vaswani2017}, $L$ is set to 6.)
\end{definition}

\autoref{df:transformer} describes only the encoder part of the Transformer, and the decoder part has a similar structure \cite{vaswani2017}. This encoder itself is an autoencoder, although the encoding-decoding mapping is not explicitly stated.

\begin{remark}
  Transformers also use multi-head structures and cross-layer connections, which can be represented using multi-head and multi-kernel methods discussed in \autoref{sec:param-local-kernel}.
\end{remark}

\subsection{Non-Local Models}\label{sec:non-local}

The so-called \term{non-local model} follows the same form as the local model in \autoref{df:local-decision}, but it is non-local relative to regular topological/neighbor structures.

\begin{definition}[Non-Local Model]\label{df:non-local}
Let $J(x_*, \theta; K) = \sum_i K(x_*, x_i) l(x_i, \theta)$ be the local loss function at target point $x_*$. Introduce the \term{non-local similarity} $\mathcal{P}(x)$, representing the index set of points $x_i$ that are similar to $x$. The non-local model is determined by the following loss function:
  \begin{equation}\label{non-local-loss}
  J_{NL}(x_i, \theta; K) := \sum_{j \in \mathcal{P}^\circ(x_i)} w_{ij} J(x_j, \theta; K),
  \end{equation}
  where $j \in \mathcal{P}^\circ$ means $j \in \mathcal{P}(x_i)$ and $j \neq i$ (the hollow form of $\mathcal{P}$, which is not necessary), and $w_{ij}$ is a reasonable weight. We call \fref{non-local-loss} the non-local loss, and the corresponding statistical decision is called the \term{non-local decision}.
\end{definition}

\begin{remark}
  Clearly, $w_{ij}$ and the kernel $K$ (and $\mathcal{P}(\cdot)$) can be fused together. Thus, we can also define the non-local model directly on an ordinary decision model:
  \begin{equation}\label{non-local-d-loss}
  J_{NL}(x_i, \theta) := \sum_{j \in \mathcal{P}(x_i), j \neq i} w_{ij} l(x_j, \theta).
  \end{equation}
\end{remark}

Introduce a kernel $W$ such that $W(x_i, x_j) = w_{ij}$, which brings us closer to the form in \fref{local-loss}:
\begin{equation}\label{local-non-local-loss}
\sum_{j \in \mathcal{P}(x_*)} W(x_*, x_j) J(x_j, \theta; K) ,
\end{equation}
The matrix form is:
\begin{equation}\label{local-non-local-loss-matrix}
J_{NL}(\matr{X}, \theta) = W(\matr{X}, \matr{X}) K(\matr{X}, \matr{X}) l(\matr{X}, \theta),
\end{equation}
We call $W$ (or the fusion of $\mathcal{P}(\cdot)$ and kernel $K$) the \term{non-local kernel}.


\autoref{df:non-local} does not specify a concrete setting for $w_{ij}$, but this is the key to non-local methods. Below is a common form of non-local machine learning models \cite{zhang2010}, with a conventional setting for $w_{ij}$.

\begin{definition}[Non-Local Machine Learning Model]\label{df:nl-ml}
Let $y \sim f(\vect{x}, \theta), \vect{x} \in \R^p$ be a machine learning model, then its non-local model is defined as:
  \begin{equation}\label{min-non-local-reg}
  \min_\theta \sum_{j \in \mathcal{P}(\vect{x}_i), j \neq i} w_{ij} \sum_k K(\vect{x}_j, \vect{x}_k) l(y_k, f(\vect{x}_k, \theta)) ,
  \end{equation}
  where $w_{ij} \approx \mathrm{e}^{-\frac{1}{2} \|\vect{z}_i - \vect{z}_j\|^2_{\bmSigma}}$, $\vect{z}_i := \{ y_l : \|\vect{x}_l - \vect{x}_i\| < \delta \}, \delta > 0$, and $\bmSigma$ is a suitable positive definite matrix. Other symbols have the same meaning as above.
\end{definition}

\begin{remark}
In \autoref{df:nl-ml}, $w_{ij}$ represents the sum of the differences of all points in the neighborhood between $\vect{x}_i$ and $\vect{x}_j$. Another simplified setting is $\vect{z}_i := \sum_{\|\vect{x}_l - \vect{x}_i\| < \delta} y_l, \delta > 0$. In this case, $w_{ij}$ is the difference of the mean of the neighboring points between $\vect{x}_i$ and $\vect{x}_j$.
\end{remark}

Based on the setting in \autoref{df:nl-ml}, it is easy to see that the kernel of the non-local machine learning model is determined by the similarity of the output $y_i$. This is an empirical kernel, rather than a regular self-localization kernel, or more specifically, a self-localization kernel on $(x_i, \vect{z}_i)$, where $\vect{z}_i$ is the set of all outputs corresponding to the neighborhood of $x_i$, including $y_i$.

Non-local methods naturally lead to concepts like \term{non-local means (NLM)} \cite{buades2005, buades2006, protter2009}.

\begin{definition}[Non-Local Means] NLM is defined as:
  \begin{equation*}
  \hat{y}_t \approx \sum_{s \in \mathcal{P}(x_t)} w_{ts} y_s,
  \end{equation*}
  where the symbols have the same meaning as above.
\end{definition}

\begin{figure}[h]
  \centering
  \includegraphics[width=0.27\textwidth]{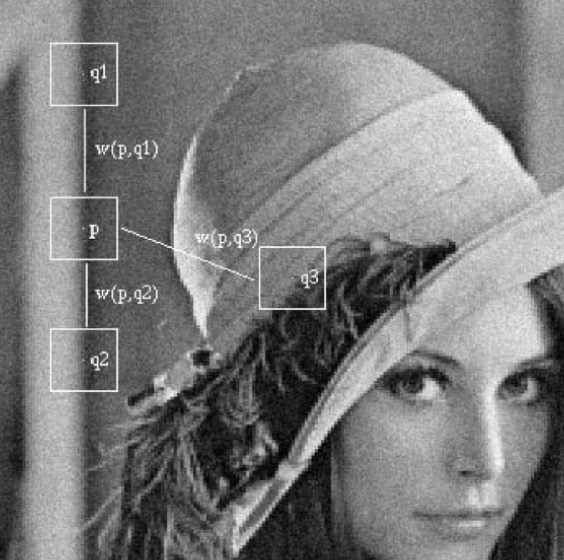}
  \caption{Illustration of NLM, where the block $(p, \vect{z}_p)$ is the neighborhood of the data points closest to the target point $(p, \vect{x}_p)$, based on the non-local kernel (quoted from \cite{buades2005}).}
  \label{fig:nl-kernel}
\end{figure}

NLM is commonly used in image processing \cite{buades2005, aubert2006}, where $\vect{x}_i$ represents pixel positions, and $y_i$ represents pixel values, \asref{fig:nl-kernel}. This is naturally a local sequence model (\autoref{df:local-ts}).

Although the non-local model is structurally the same as the local-local model, it is still a local model. However, \cite{zhang2010} points out that the non-local model is more robust than regular local models. \cite{wang2018} combined this approach with neural networks, creating a novel deep learning model.

\subsection{Graph Theory Inspired-Kernels}

According to \autoref{rm:kernel-graph}, we construct several variants of kernels inspired by graph theory.

\subsubsection{Heterogeneous kernels}

Localization kernels do not necessarily have to be defined on $\mathcal{X}\times \mathcal{X}$. They can also be defined on $\mathcal{X}\times \mathcal{Y}$, called \term{heterogeneous (localization) kernels}, where $\mathcal{X}\neq \mathcal{Y}$. The concept of heterogeneous kernels comes entirely from heterogeneous graph models \cite{shi2022}. Heterogeneous kernels abandon all restrictions on regular kernels, meaning any binary function can potentially become a heterogeneous kernel. Models constructed from these kernels are called \term{heterogeneous localization kernel models}. This paper will not delve into these models in detail but will provide a few examples for reference.

\begin{definition}[Heterogeneous Local Regression/Mean]
Define a heterogeneous kernel $K$ on $\mathcal{X}\times \mathcal{Z}$. We modify the local regression/mean (\autoref{df:local-ave}) as follows:
\begin{equation}\label{heter-local-ave}
  \hat y(x_*):=\sum_i K(x_*,z_i)y_i/\sum_i K(x_*,z_i),
  \end{equation}
  where $\{(z_i,y_i)\}$ is the sample. The corresponding loss is:
  \begin{equation*}
  J(K) = \sum_i \big|y_i-\sum_j K(x_i,z_j)y_j/\sum_j K(x_i,z_j)\big|^2.
  \end{equation*}
\end{definition}

Take an example to explain the meaning of model \eqref{heter-local-ave}. Our task is to predict $\hat{y}$ given $x_*\in\mathcal{X}$, such as in image classification. However, we only have a dataset on $\mathcal{Z}\times \mathcal{Y}$, $\{(z_i,y_i)\}$, rather than on $\mathcal{X}\times \mathcal{Y}$, such as in text classification data. The trick of this model is to construct a heterogeneous kernel $K(x, z)$ on $\mathcal{X}\times \mathcal{Z}$ to complete the prediction task, such as measuring how well a text describes an image.

\begin{definition}[Heterogeneous Self-Local Mean]
Define a heterogeneous kernel $K$ on $\mathcal{X}\times \mathcal{Y}$. We construct the heterogeneous self-local mean as follows:
\begin{equation}\label{heter-self-local-ave}
  \hat{x}_*:=\sum_i K(x_*,y_i)x_i/\sum_i K(x_*,y_i),
  \end{equation}
  where $\{(x_i,y_i)\}$ is the sample.
\end{definition}

The meaning of model \eqref{heter-self-local-ave} is that if we already have reliable knowledge about the heterogeneous kernel $K(x,y)$ on $\mathcal{X}\times \mathcal{Y}$, then we do not need to construct a kernel $K(x,x')$ on $\mathcal{X}$ to compute the local mean of the samples $\{x_i\}$.

\begin{remark}
  A heterogeneous kernel can be used to construct a homogeneous kernel: $K(f_1(x),f_2(x'))$, where $K$ is a heterogeneous kernel on $\mathcal{Z}_1\times \mathcal{Z}_2$, and $f_1:\mathcal{X}\to \mathcal{Z}_1,f_2:\mathcal{X}\to \mathcal{Z}_2$ are reasonable mappings.
\end{remark}


Following \autoref{sec:local-feature}, we construct feature mappings $\phi(x)$ and $\psi(y)$ for the heterogeneous kernel $K(x,y)$. We further introduce a low-rank approximation/low-dimensional embedding for the heterogeneous kernel:
\begin{equation*}
  K(x,y)\approx \phi(x)\cdot \psi(y).
\end{equation*}

\subsubsection{Hyperkernels}

Inspired by the concept of hypergraphs, we can also construct \term{($m$-ary) hyperkernels}, which are kernels defined on multiple sample spaces, $K:\mathcal{X}_1\times \cdots \times \mathcal{X}_m\to\mathbb{R}$. In particular, we define a multivariate kernel on $\mathcal{X}$ as $K:\mathcal{X}^m\to\mathbb{R}$. If $\mathcal{X}_1=\cdots =\mathcal{X}_m$, then the hyperkernel is homogeneous; otherwise, it is heterogeneous.

Let $\samp{X}_l=\{x^{l}_i\}$ be samples on $\mathcal{X}_l$ for $l=1,\cdots,m$. The tensor
\begin{equation*}
K(\samp{X}_1,\cdots,\samp{X}_m)=\{K(x^{1}_{i_1},\cdots,x^{m}_{i_m})\}_{i_1\cdots i_m}
\end{equation*}
is called the \term{(empirical) hyperkernel tensor}.

An $m$-ary (hyper)kernel can have $m$ feature mappings $\phi_j, j=1,\cdots, m$, e.g.,
$$K(x_1,\cdots,x_m)=F(\phi_1(x_1),\cdots,\phi_m(x_m)).$$

\begin{remark}
Hyperkernels are multivariate/listwise; ordinary localization kernels are bivariate/pairwise; standard loss functions (such as likelihood) are univariate/pointwise.
\end{remark}

\begin{example}[Mixup \cite{zhang2017mixup}]
\cite{zhang2017mixup} proposed a data augmentation method that generates new samples by mixing multiple data points. This can be generally represented using a hyperkernel as
\begin{equation*}
p(x_3|x_1,x_2) \sim K(x_1,x_2,x_3).
\end{equation*}
\end{example}

Hyperkernels can capture higher-order similarities in data, offering unique benefits in low-dimensional embeddings beyond those of standard kernels. Consider the following example.

\begin{figure}[h]
\centering
\begin{subfigure}[t]{0.4\textwidth}
\centering
\includegraphics[width=\textwidth]{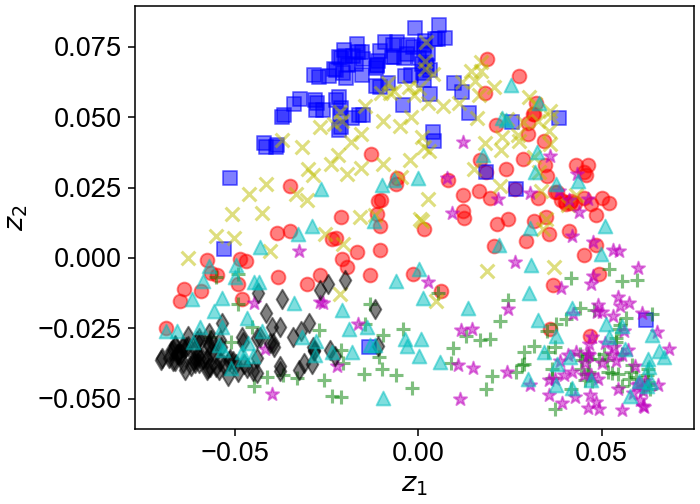}
\caption{Without label information}\label{fig:trimap-2d}
\end{subfigure}
\begin{subfigure}[t]{0.4\textwidth}
\centering
\includegraphics[width=\textwidth]{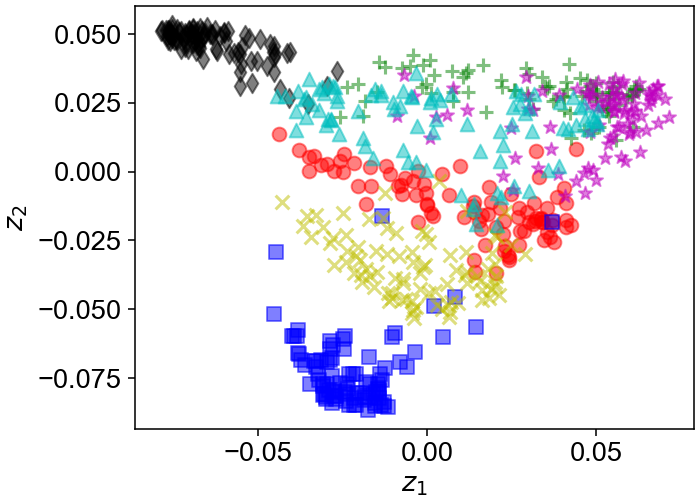}
\caption{With label information}\label{fig:trimap-labeled-2d}
\end{subfigure}
\caption{TriMap's planar embedding effect (on a subset of the Fashion MNIST dataset)}\label{fig:trimap}
\end{figure}

\begin{example}[TriMap \cite{amid2019}]
Let $s$ denote a similarity function on the sample space $\mathcal{X}$. We define a ternary kernel on $\mathcal{X}$ as
\begin{equation*}
K(x_1,x_2,x_3) = \frac{s(x_1,x_2)}{s(x_1,x_2)+s(x_1,x_3)}, \quad x_1,x_2,x_3\in\mathcal{X},
\end{equation*}
which expresses how much closer $x_2$ is to $x_1$ compared with $x_3$. This can be termed a \term{contrast kernel}. The corresponding \term{ternary MDS} is formulated as
\begin{gather}\label{trimap-mds}
\min_{{\vect{z}_i}}\sum_{i_1i_2i_3} K(x_{i_1},x_{i_2},x_{i_3}) h(d_{i_1i_2}^2- d_{i_1i_3}^2)
d_{ij} := \|\vect{z}_{i} - \vect{z}_{j}\|_2,
\end{gather}
where $h(\cdot)$ is a well-designed increasing function. When $h(x)=x$, \fref{trimap-mds} is equivalent to MDS based on the kernel matrix ${\sum_{i_3}K(x_{i_1},x_{i_2},x_{i_3})}$. Its planar embedding effect is illustrated in \autoref{fig:trimap}.
\end{example}

\begin{example}[Ordinal Embedding \cite{terada2014}]
Let $d$ be a dissimilarity measure on $\mathcal{X}$. We define a quaternary kernel on $\mathcal{X}$ as
\begin{equation*}
K(x_1,x_2,x_3,x_4) = \begin{cases}
1, & d(x_1,x_2) < d(x_3,x_4)\\
0, & \text{otherwise}.
\end{cases}
\end{equation*}
A feasible \term{quaternary MDS} is given by
\begin{equation*}
\min_{{\vect{z}_i}}\sum_{i_1i_2i_3i_4}K(x_{i_1},x_{i_2},x_{i_3},x_{i_4})\chr_{(\|\vect{z}_{i_1}- \vect{z}_{i_2}\|_2>\|\vect{z}_{i_3} - \vect{z}_{i_4}\|_2)}.
\end{equation*}
\end{example}

\subsubsection{Random kernels}

A \term{random kernel} is a kernel whose values are random variables, or equivalently, a kernel with stochastic values. This concept is evidently inspired by \term{random graphs}\cite{drobyshevskiy2019}.

\begin{example}\label{ex:random-kernel}
Let a random kernel (matrix) be defined as $\matr{K}_{ij}\sim B(p_{ij})$, where $B$ represents the Bernoulli distribution. This kernel is fully characterized by the matrix ${p_{ij}}$, where $p_{ij}$ may be given or unknown.
\end{example}

The kernel in \autoref{ex:random-kernel} is a parameter with prior distributions. Hence, random kernel models can be regarded as applying Bayesian methods to local models.

\section{Summary}

This paper systematically explores the localization method, a topic that is not frequently addressed in standard machine learning monographs. The localization method is a general-purpose technique, which is manifested in two aspects: based on any statistical model, we can construct a corresponding local model; even if the corresponding global model is extremely simple, a local model can approximate any model. In fact, our true objective is to prove that any model (at least those models that have been constructed by people) is essentially a local model/local mean.

We define local decision-making as a general form of local models. The key to localization methods is defining a localization kernel. The construction conditions for localization kernels are much more relaxed than for positive definite kernels. Localization kernels can also have feature mappings, but there is no specific relationship between the two. We then introduce several useful concepts, such as local mean/mode, self-local mean/mode, and their sequential models. We find that local regression has a form similar to kernel regression.

Self-local mean is the core concept of the method, which essentially refers to the local mean based on self-localization kernels. We pay special attention to the principle of the MeanShift algorithm and present its applications in clustering and image processing. We highlight the relationship between self-local mean and KDE. In fact, self-local mean and sample noise addition are inverse operations. We further discuss temporal self-local mean, which is essentially the local mean of self-localization kernels. Most notably, the self-attention mechanism is a form of temporal self-local mean.

We argue that any local model is essentially equivalent and can be represented using local means. When converting a local model into a local mean representation, the corresponding kernel transforms into an empirical kernel form (\fref{emp-kernel}). Local linear regression illustrates this point well.

During the process of introducing the basic principle, we mentioned numerous classic models that have a connection with the localization method, such as relaxation labeling, Hopfield networks, local linear embedding and fuzzy inference.

After introducing several specific local models, we discuss DAE, score-based models, and diffusion models, all of which have deep connections with local means. Meanwhie, it is pointed out that noise addition and local averaging correspond to the forward and backward processes of the SDE.

We briefly introduce hierarchical local models and construct the Transformer using the concept of local means. Formally, non-local means are also local means, but they provide a special construction of self-localization kernels.

Finally, inspired by graph theory, we construct some novel kernel functions, which may become important research objects in the future. Future work also includes the integration of local models with incremental/transfer/lifelong learning (as well as other machine learning paradigms), and the specific implementation of adaptive kernel algorithms.


\bibliographystyle{unsrtnat}
\bibliography{reference}

\appendix


\section{Comparison with kernel methods}

From the perspective of kernel construction, kernel methods can be seen as a special form of localization methods. However, we are not yet ready to draw this conclusion. Comparing localization methods with kernel methods is highly meaningful \cite{hastie2009}. Please refer to \autoref{tab:comparison-ker-loc} and \autoref{tab:analogy-ker-loc}.

\begin{table}[h]
\centering
\caption{Comparison of Localization Methods and Kernel Methods}
\label{tab:comparison-ker-loc}
\resizebox{\textwidth}{!}{%
\begin{tabular}{c|l|l}
\toprule
\thbf{Aspect} & \thbf{Localization Methods} & \thbf{Kernel Methods} \\
\midrule
Theoretical Basis & Graph theory, topology, Markov networks & Real analysis, with a well-developed theory \\
Kernel Construction Conditions & More relaxed & Must satisfy positive definiteness \\
Neighborhood/Topological Concepts & Usually depends on the concepts & Not significant \\
Computational Complexity & Higher; each data point requires calculation. & relatively efficient \\
Generality & Highly general, almost unlimited & Kernel tricks have limited applicability \\
Applicable Problems & More suitable for large datasets & Very suitable for high-dimensional cases \\
Data Format & Almost no restrictions & Almost no restrictions \\
Kernel (Matrix) & Analogous to joint distributions & Analogous to covariance \\
Normalized Kernel (Matrix) & Analogous to transition distributions & Analogous to correlation coefficients \\
\bottomrule
\end{tabular}
}
\end{table}

\begin{table}[h]
\centering
\caption{Analogy Between Localization Methods and Kernel Methods}\label{tab:analogy-ker-loc}
\resizebox{\textwidth}{!}{%
\begin{tabular}{c|c}
\toprule
\thbf{Localization Methods} & \thbf{Kernel Methods} \\
\midrule
Localization Kernel & Positive Definite Kernel \\
Feature Mapping (Unrelated to Kernel) & Feature Mapping \\
Localization Techniques/Models & Kernel Techniques/Models \\
Localization Kernel Matrix & Kernel Matrix \\
Kernel Matri Related To Stochastic Matrix & Kernel Matrix Related To Projection Matrix\\
Weighted Directed Graph & Weighted Undirected Graph \\
Empirical/Adaptive (Localization) Kernel & Empirical/Adaptive Kernel \\
Approximate Feature Mapping & Approximate Feature Mapping \\
(Asymmetric) Normalization & (Symmetric) Normalization \\
Regularization & Regularization \\
Smoothing Space & RKHS \\
Smoothing Distance & MMD \\
(Self) Local Mean, Local Regression & Mean Embedding, Kernel Regression \\
(Self) Local Mode & Kernel SVM \\
Local Mean (Lazy) Autoencoder & Kernel Autoencoder \\
Asymmetric MDS, LEE, Local PCA & MDS, Kernel PCA \\
\bottomrule
\end{tabular}
}
\end{table}



\section{Center-Based Classifiers}\label{sec:center-classifier}

We first define a special class of classifiers and then highlight their relationship with linear classifiers.

\begin{definition}[Center-Based Classifier \cite{gao2007}]\label{df:center-classifier}
  On the point set $\mathcal{X}$, construct the decision function for the \term{center-based classifier}:
  \begin{equation}\label{center-clf}
  \delta_k(\vect{x})= - d(\vect{x},\vect{\mu}_k), k=1,\cdots,K,
  \end{equation}
  where the unknown parameter $\vect{\mu}_k$ is called the \term{center} of the classifier, and $d$ is the \term{distance function} on $\mathcal{X}$ (which may include unknown parameters besides $\vect{\mu}_k$). The estimate of the center is equivalent to the solution to the following optimization problem:
  \begin{equation*}
  \min_{\{\vect{\mu}_k\}} \big\{\sum_i d(\vect{x}_i,\vect{\mu}_{y_i}) = \sum_k\sum_{\vect{x}_i:k}d(\vect{x}_i,\vect{\mu}_k) \big\}.
  \end{equation*}
\end{definition}

It is not difficult to prove that the estimate of the center $\vect{\mu}_k$ is:
\begin{equation*}
\hat{\vect{\mu}}_k = \argmin_{\vect{\mu}_k}\sum_{\vect{x}_i:k}d(\vect{x}_i,\vect{\mu}_k).
\end{equation*}

\begin{remark}
Here, the distance function is not a strict distance in the formal sense, but rather a measure of dissimilarity between two points \cite{sullivant2018}. Moreover, the distance can be replaced with the similarity, like $s(x,y)\propto \exp^{-d(x,y)}$.
\end{remark}


\begin{example}
LDA/QDA are center-based classifiers (with priors), whose center is the sample mean of each class, and the distance is the \term{Mahalanobis distance}.
\end{example}

\begin{example}
  The simplest center-based classifier is the \term{K-means classifier} on $\mathcal{X} \subset \R^p$:
\begin{equation}\label{kmeans-clf}
\delta_k(\vect{x})= -\|\vect{x}-\vect{\mu}_k\|_2^2,
\end{equation}
  i.e. the distance is derived from the 2-norm (or \term{Euclidean distance}).
\end{example}

Clearly, the estimate of the parameter $\vect{\mu}_k$ for the K-means classifier is the mean of the sub-sample ${\vect{x}_i:k}$ of the $k$-th class, where $N_k$ is the size of the sub-sample.

\section{Centerless Classifiers}\label{sec:centerless-clf}

By substituting the center with the sample mean, it can be shown that \fref{kmeans-clf} is equivalent to the following centerless form:
\begin{equation}\label{kmeans-clf-centreless-d}
  \delta_k(\vect{x}) \sim -\frac{1}{N_k}\sum_{\vect{x}_i:k} \|\vect{x}-\vect{x}_i\|_2^2 + \Var(\{\vect{x}_i:k\}) .
\end{equation}

\begin{definition}[Centerless Classifier]\label{df:centerless-clf}
Generally, we construct the following \term{centerless classifier}:
\begin{equation}\label{centerless-clf}
  \delta_k(\vect{x})=-\frac{1}{N_k}\sum_{\vect{x}_i:k} d(\vect{x},\vect{x}_i) + \frac{1}{2N_k^2}\sum_{\vect{x}_i,\vect{x}_j:k}  d(\vect{x}_i,\vect{x}_j),
\end{equation}
or more generally,
\begin{equation*}
  \delta_k(\vect{x})=-d(\vect{x},\{\vect{x}_i:k\}) + \Omega_d(\{\vect{x}_i:k\}),
\end{equation*}
where $d(\cdot,\{\cdot\})$ and $\Omega_d(\cdot)$ are the distance from a point to a set of points derived from the distance $d$, and a measure reflecting the dispersion of the point set, respectively.
\end{definition}

\begin{remark}\label{rm:clf-cen-eq}
  The K-means classifier and the centerless form given by \fref{kmeans-clf-centreless-d} are completely equivalent, while a general center-based classifier and the centerless classifier derived from the same distance are not strictly equivalent, but their predictions will not differ significantly.
\end{remark}


\section{Lazy Prediction and Lazy Models}\label{sec:lazy-pred}

For parametric models, prediction is an intermediate task that depends on parameter estimation: for a parametric model $P(x|\theta)$, we first estimate $\hat{\theta}(S)$ based on the sample $S$, and then predict according to the determined distribution $P(x|{\hat{\theta}})$. As a prediction of $x$, $\hat{x}$ depends on $\hat{\theta}$ and ultimately on the sample $S$, denoted as $\hat{x}(S)$.

\begin{definition}[Lazy Model]\label{df:lazy}
If the prediction $\hat{x}(S)$ has a relatively simple closed-form expression, then parameter estimation is not required before making predictions. We refer to such a prediction as a \term{lazy prediction}. The function $\hat{x}(S)$ itself is regarded as a model that only performs predictions, known as a \term{lazy model} or \term{delayed model}. Specifically, if $\hat{x}(S)$ is linear in $S$, it is called a linear lazy prediction/model.
\end{definition}


From a lazy prediction $\hat{x}(S)$, we can derive a mapping $S\to S'$, where $S'$ consists of multiple (random) prediction results of $\hat{x}(S)$, simply denoted as $S'=TS$, which is referred to as a \term{lazy transformation/mapping}. Such transformation can continuously update samples to reduce prediction errors.


However, we are more accustomed to discussing lazy models in the context of machine learning where \term{lazy models} or lazy prediction is based on conditional prediction\cite{maron1997,wettschereck1997,bontempi1999,xie2002,garcia2009}.

\begin{definition}[Lazy Model]\label{df:cond-lazy}
Similar to \autoref{df:lazy}, if the conditional prediction $\hat{y}(x;S)$ can be expressed in some closed form, then there is no need for substantial parameter estimation before making predictions. Such a prediction is referred to as \term{lazy conditional prediction}, or simply lazy prediction, and the corresponding model is still called a lazy model.
\end{definition}

\begin{remark}
  A lazy model may not involve any learning process, yet the term \term{lazy learning} (sometimes also referred to as \term{instance-based learning} \cite{aha1991}) is still used to describe its overall execution process.
\end{remark}


\begin{definition}[Lazy Transformation]\label{df:lazymap}
Let $\samp{X}=\{x_i\},\samp{y}=\{y_i\}$, and $S=(\samp{X},\samp{y})$.
Applying a lazy prediction to $\samp{X}'=\{x_i'\}$ results in the corresponding predictions $\samp{y}'=\{\hat{y}(x_i',\samp{X},\samp{y})\}$. Fixing $\samp{X}'$ and $\samp{X}$, we regard this as the \term{lazy transformation/mapping} from $\samp{y}$ to $\samp{y}'$, denoted as 
$$\samp{y}'=T(\samp{X}',\samp{X})\samp{y}.$$
A lazy transformation typically refers to the case where $\samp{X}'=\samp{X}$, which is simply written as $\samp{y}'=T(\samp{X})\samp{y}$. Specifically, if $T(\samp{X}',\samp{X})$ or $T(\samp{X})$ is a linear transformation, it is called a \term{linear lazy transformation}.
\end{definition}

\begin{remark}\label{rm:lazymap}
A lazy transformation is essentially a lazy prediction, but it emphasizes the transformation in the output space $\mathcal{Y}$. Such lazy prediction lead to modifications of the predictions for the same input samples. These modifications may positively impact the final prediction results.
\end{remark}

\begin{remark}
  A linear lazy transformation is indeed a local mean based on the emperical kernel.
\end{remark}

If we repeatedly apply a lazy transformation, it will converge to a special output $\vect{y}^*$ related to the data $\samp{X}$:
\begin{equation}\label{fp}
T(\samp{X})^n\vect{y}\to \vect{y}^*, ~n\to\infty.
\end{equation}
Since $\vect{y}^*$ does not fundamentally depend on the true outputs $\vect{y}$, this provides a special approach to unsupervised learning (dimensionality reduction if $\vect{y}$ is continuous, or clustering if $\vect{y}$ is discrete). To distinguish it from the true output (labels), we refer to the elements of $\vect{y}^*$ or any $T(\samp{X})^n\vect{y}$ in \fref{fp} as the pseudo-labels.

\begin{definition}\label{df:lazymap-model}
We refer to the above iterative process as \term{lazy transformation iteration}. Since it simply repeats sample predictions, it is often called a \term{prediction iterative} algorithm. The unsupervised model derived from this process is called a \term{lazy transformation model}. If the lazy prediction/transformation is stochastic, then the iterative process forms a Markov chain of samples, called a \term{lazy transformation Markov chain}.
\end{definition}


\begin{remark}
  The transformation sequence $T(\samp{X})^n$ in \fref{fp} may have a limit, denoted as $T(\samp{X})^\infty$, which is called the \term{equivalent lazy transformation} of the lazy transformation model.
\end{remark}

\begin{remark}\label{rm:lazy-new-sample}
  For a new sample set $\samp{X}'$, the lazy prediction/mapping is given by $\hat{\vect{y}}'=T(\samp{X}',\samp{X})\vect{y}$. This mapping can also be iterated multiple times. We believe this improves prediction results and is equivalent to mixing new and old samples, learning from them together, and then making new predictions.
\end{remark}

\begin{example}
  The prediction formula for linear regression is 
  $$\vect{y}^* = \matr{X}(\matr{X}^{\mathrm{T}}\matr{X})^{-1}\matr{X}^{\mathrm{T}}\vect{y},$$
  so it can be regarded as a linear lazy transformation, while prediction iteration has no effect on it. For ridge regression, $\vect{y}^* = \matr{X}(\matr{X}^{\mathrm{T}}\matr{X}+\lambda)^{-1}\matr{X}^{\mathrm{T}}\vect{y}$, prediction iteration does not play a significant role either.
\end{example}

\begin{example}
  K-means classifer in centerless form is a lazy model and K-means algorithm is its predition iteration.
\end{example}

There are various strategies for implementing lazy transformation iteration, and it does not have to be performed exactly as in \fref{fp}. For example, we can use the \term{mini-batch method} for iteration: first, initialize part of $\vect{y}_C$, then perform the following iteration:
\begin{equation}\label{fp-batch}
\vect{y}_B\leftarrow T(\samp{X}_B,\samp{X}_C)\vect{y}_C
\end{equation}
where $B,C\subset\{1,\cdots, N\}$ are subsets of sample indices, and typically, let $B\cap C=\emptyset$. After each update of \fref{fp-batch}, $C$ can be replaced by $B\cup C$. An intuitive description of \fref{fp-batch} is that the output of some samples propagates to all samples through the iterative rule. A lazy transformation model/algorithm that primarily improves using the mini-batch method can be called a \term{propagation model/algorithm}.


\section{Autoencoder}\label{sec:autoencoder}

\begin{definition}
  An autoencoder consists of a learnable encoder $\phi: \mathcal{X} \to \mathcal{Z}$ and a decoder $\psi: \mathcal{Z} \to \mathcal{X}$ (also learnable), and is solved by the solution to the following optimization problem:
$$
\min_{\phi,\psi}\sum_id(x_i,\psi(\phi(x_i))),
$$
where $d$ denotes a reasonable distance and $\{x_i\}$ is a sample. The reconstruction of the autoencoder is $T := \psi\phi: \mathcal{X} \to \mathcal{X}$.
\end{definition}

An autoencoder may not have an explicit prediction formula, instead, it needs to solve an approximate fixed-point problem:
$$
\min_{x \in \mathcal{X}} d(x, Tx),
$$
where $T$ is known or estimated. The prediction/generation of $x$ determined in this way is regarded as an implicit prediction, called the \term{reconstruction iteration}, i.e., $\hat{x}\approx T^n x_0$ with $n \to \infty$ and $x_0$ being a reasonable initial value. If the reconstruction is stochastic, then the iterative process forms a Markov chain on $\mathcal{X}$, called a \term{reconstruction Markov chain}. The limiting distribution of this Markov chain can be expected to be the distribution of $x$, which gives rise to a generative model.

\begin{remark}\label{rm:supervised-ae}
  The supervised model of $\mathcal{X}\to \mathcal{Y}$ can be implemented using an autoencoder on $\mathcal{X}\times \mathcal{Y}$. When making predictions based on $x_*$, we need to solve the following approximate fixed-point problem:
  $$
\min_{y \in \mathcal{Y}} d((x_*,y), T(x_*,y)),
$$
which is also the principle of missing data imputation.
\end{remark}

Based on the above discussion, we propose the following hierarchical model.
\begin{definition}
The hierarchical auto-encoder is composed of a set of autoencoders $(\phi_l,\psi_l), l = 1,\cdots, L$. During learning, it is usually only required that  $\psi_L\phi_L\cdots\psi_1\phi_1$ approximates the identity mapping, rather than requiring each  $\psi_l\phi_l$ to approximate the identity mapping.
\end{definition}

The hierarchical autoencoder is regarded as learning a reconstruction Markov chain directly. Based on the autoencoder, to generate $x$, one only needs to calculate the reconstruction $\psi_L\phi_L\cdots\psi_1\phi_1(x)$ once, without (or with few) reconstruction iterations based on this reconstruction.

\begin{remark}
  It is worth noting that the reconstruction (and its iteration) of the hierarchical autoencoder is the principle of the trendy Chain of Thought (CoT) \cite{wei2022}.
\end{remark}

\section{Lazy autoencoder}\label{sec:lazy-autoencoder}

\begin{definition}
If an autoencoder does not explicitly provide an encoder or a decoder (especially in the parametric form), but only has a reconstruction (called lazy reconstruction), it is called an \term{lazy autoencoder}, denoted as $T(x; \samp{X})$. The lazy autoencoder yields implicit lazy prediction or reconstruction iteration.
\end{definition}

\begin{remark}
  Strictly speaking, (lazy) reconstruction iteration is a form of (lazy and implicit) prediction rather than predictive iteration. Lazy transformation is not lazy reconstruction because it is directly defined on samples. When lazy reconstruction acts on the entire sample, it results in a lazy transformation, i.e. $\samp{X}\mapsto T(\samp{X}; \samp{X})$.
\end{remark}

\begin{example}
The self-local mean is a type of lazy autoencoder, and the MeanShift iteration can be regarded as its reconstruction iteration.
\end{example}

\begin{example}
  As a discrete autoencoder, K-means clustering can lead to lazy reconstruction:
  \begin{equation}\label{centerless-recon}
  T\vect{x}=\frac{1}{|\{\phi(\vect{x}_i)=\phi(\vect{x})\}|}\sum_{\phi(\vect{x}_i)=\phi(\vect{x})}\vect{x}_i, \vect{x}\in \R^p,
  \end{equation}
  where $\phi$ is the K-means classifier(centerless form), as its encoder.
\end{example}

\section{Categorical-Style Notation}

Assume that all machine learning models form a category. Let $\mathrm{loc}(M)$ denote the application of the localization trick to a given model $M$ that is an object of the category. We refer to the ``functor'' $\mathrm{loc}$ as a constructor transforming one model to another.

The local model $\mathrm{loc}(M)$ has the following important properties:
\begin{enumerate}
    \item $\mathrm{loc}(M) \simeq \mathrm{loc}(\mathrm{loc}(M))$ (up to isomorphism in the categorical sense, though not strictly).
    \item $\mathrm{loc}(M)$ is more complex than $M$, particularly when $M$ is linear.
    \item $\mathrm{loc}(M)$ is universal in nature, akin to the power of a neural network.
\end{enumerate}



\end{document}